\documentclass[acmsmall,natbib=true]{acmart}

\renewcommand\footnotetextcopyrightpermission[1]{} 

\usepackage{natbib}
\usepackage{subcaption}

\usepackage{amsmath}

\DeclareMathOperator*{\argmin}{arg\,min}

\newcommand{\w}{\mathbf{w}}
\newcommand{\e}{\mathbf{e}}
\newcommand{\delw}{\delta \mathbf{w}}

\newcommand{\gradw}{\nabla_\w L}

\newcommand{\hess}{\mathbf{H}}
\newcommand{\hessinv}{\mathbf{H}^{-1}}

\renewcommand{\vec}[1]{\mathbf{#1}}

\def\va{{\mathbf{a}}}

\def\vg{{\mathbf{g}}}

\def\vs{{\mathbf{s}}}

\def\vu{{\mathbf{u}}}
\def\vv{{\mathbf{v}}}
\def\vw{{\mathbf{w}}}
\def\vx{{\mathbf{x}}}
\def\vy{{\mathbf{y}}}

\newcommand{\R}{\mathbb{R}}

\AtBeginDocument{%
  \providecommand\BibTeX{{%
    \normalfont B\kern-0.5em{\scshape i\kern-0.25em b}\kern-0.8em\TeX}}}

\acmVolume{}
\acmNumber{}
\acmArticle{}
\acmMonth{}

\begin{document}

\title{Sparsity in Deep Learning: Pruning and growth for efficient inference and training in neural networks}

\author{Torsten Hoefler}
\email{htor@inf.ethz.ch}
\affiliation{%
  \institution{ETH Z\"urich}
  \city{Z\"urich}
  \country{Switzerland}
  \postcode{8092}
}
\author{Dan Alistarh}
\email{dan.alistarh@ist.ac.at}
\affiliation{%
	\institution{IST Austria}
	\city{Klosterneuburg}
	\country{Austria}
	\postcode{3400}
}
\author{Tal Ben-Nun}
\email{talbn@inf.ethz.ch}
\affiliation{%
	\institution{ETH Z\"urich}
	\city{Z\"urich}
	\country{Switzerland}
	\postcode{8092}
}
\author{Nikoli Dryden}
\email{ndryden@ethz.ch}
\affiliation{%
  \institution{ETH Z\"urich}
  \city{Z\"urich}
  \country{Switzerland}
  \postcode{8092}
}
\author{Alexandra Peste}
\email{alexandra.peste@ist.ac.at}
\affiliation{%
	\institution{IST Austria}
	\city{Klosterneuburg}
	\country{Austria}
	\postcode{3400}
}

\renewcommand{\shortauthors}{Torsten Hoefler et al.}
\renewcommand{\shorttitle}{Sparsity in Deep Learning}

%

\begin{abstract}
The growing energy and performance costs of deep learning have driven the community to reduce the size of neural networks by selectively pruning components. 
Similarly to their biological counterparts, sparse networks generalize just as well, if not better than, the original dense networks. 
Sparsity can reduce the memory footprint of regular networks to fit mobile devices, as well as shorten training time for ever growing networks.
In this paper, we survey prior work on sparsity in deep learning and provide an extensive tutorial of sparsification for both inference and training.
We describe approaches to remove and add elements of neural networks, different training strategies to achieve model sparsity, and mechanisms to exploit sparsity in practice.
Our work distills ideas from more than 300 research papers and provides guidance to practitioners who wish to utilize sparsity today, as well as to researchers whose goal is to push the frontier forward.
We include the necessary background on mathematical methods in sparsification, describe phenomena such as early structure adaptation, the intricate relations between sparsity and the training process, and show techniques for achieving acceleration on real hardware. 
We also define a metric of pruned parameter efficiency that could serve as a baseline for comparison of different sparse networks.  
%
We close by speculating on how sparsity can improve future workloads and outline major open problems in the field.
\end{abstract}

%


\maketitle

\newcommand{\para}[1]{\noindent\textcolor{green}{[para: #1]}\newline}
\newcommand{\parad}[1]{\noindent\textcolor{green}{[para: #1]}\newline}
\renewcommand{\parad}[1]{}
\newcommand{\htor}[1]{\textcolor{blue}{[htor: #1]}}
\newcommand{\dan}[1]{\textcolor{orange}{[dan: #1]}}
\newcommand{\tal}[1]{\textcolor{magenta}{[tal: #1]}}
\newcommand{\ndryden}[1]{\textcolor{teal}{[ndryden: #1]}}
\newcommand{\alex}[1]{\textcolor{green}{[alex: #1]}}

\ \newline \textcolor{white}{.}
\hspace{1pt} \emph{The supreme goal of all theory is to make the irreducible basic elements as simple and as few as possible without having to surrender the adequate representation of a single datum of experience - } 
\begin{flushright} Albert Einstein, 1933 \end{flushright}
\ \newline

\section{Introduction}

\parad{intro}
Deep learning shows unparalleled promise for solving very complex real-world problems in areas such as computer vision, natural language processing, knowledge representation, recommendation systems, drug discovery, and many more.
With this development, the field of machine learning is moving from traditional feature engineering to neural architecture engineering. However, still little is known about how to pick the right architecture to solve a specific task.
Several methods such as translational equivariance in convolutional layers, recurrence, structured weight sharing, pooling, or locality are used to introduce strong inductive biases in the model design. 
Yet, the exact model size and capacity required for a task remain unknown and a common strategy is to train overparameterized models and compress them into smaller representations. 

\parad{background - popular science bio motivation}
Biological brains, especially the human brain, are hierarchical, sparse, and recurrent structures~\cite{karl_hierarchical_models} and one can draw some similarities with the inductive biases in today's artificial neural networks. 
Sparsity plays an important role in scaling biological brains---the more neurons a brain has, the sparser it gets~\cite{Herculano-Houzel19008}. 
Furthermore, research has shown that a human brain starts sparse, has an early phase of densification followed by massive pruning, and then remains at a relatively stable sparsity level. Yet, even fully-grown brains change up to 40\% of their synapses each day~\cite{2017-hawkins}. 
Many of today's engineered pruning techniques have intuitive biological analogies, which we will mention throughout the text and discuss in Section~\ref{sec:discussion}. Yet, the computational substrates (biological tissue vs. CMOS) result in very different constraints.

\parad{DL models are overparameterized - and that's needed!}
Artificial deep learning models are traditionally dense and  over-parameterized, sometimes to the extent that they can memorize random patterns in data~\cite{zhang2017understanding} or that 95\% of the parameters can be predicted from the remaining 5\%~\cite{denil2014predicting}. 
This may be linked to empirical evidence suggesting that over-parameterized models are easier to train with stochastic gradient descent (SGD) than more compact representations~\cite{glorot2011deep,mhaskar2016deep,2020-li,kaplan2020scaling}. \citet{brutzkus2017sgd} and \citet{du2019gradient} show that such gradient descent techniques provably train (shallow) over-parameterized networks optimally with good generalization. Specifically, they show that over-parameterization leads to a strong ``convexity-like property'' that benefits the convergence of gradient descent. Recent theoretical results~\cite{allenzhu2019convergence,neyshabur2018understanding} seem to support these findings and indicate that training dynamics and generalization rely on  overparameterization. 

\parad{why would we go sparse?}
This over-parameterization comes at the cost of additional memory and computation effort during model training and inference. In particular, for inference on mobile and battery-driven devices and in cost-conscious settings, sparse model representations promise huge savings. Concretely, sparse models are easier to store, and often lead to computational savings. 
Furthermore, overparameterized models tend to overfit to the data and degrade generalization to unseen examples. Following Occam's razor, sparsification can also be seen as some form of regularization, and may improve model quality by effectively reducing noise in the model. Specifically, the framework of Minimum Description Length provides an attractive formulation with a Bayesian interpretation and a clear interpretation as data compression~\cite{2007-grunwald}, as we discuss later. 

Many, especially older, works centered on improved generalization through sparsification. 
Early research~\cite{1988-mozer} focused on models with tens to hundreds of parameters also describe better interpretability of their sparsified versions. However, with today's models using millions or billions of parameters, it is to be seen if sparsity improves explainability and interpretability significantly.
The recent work of \citet{2019-bartoldson} models pruning as ``noise'' similar to dropout or data augmentation to explain generalization. 
Other recent works found that sparsity can improve robustness against adversarial attacks~\cite{guo2018sparse,gopalakrishnan2018combating,10.1145/3386263.3407651,sehwag2020hydra,cosentino2019search,2020-verdenius,2020-madaan}.

A larger group of works recently focused on improving the computational efficiency while maintaining the model accuracy. 
Modern networks are computationally expensive to use --- for example, Inception-V3~\cite{2016-szegedy}, a state of the art object recognition network, requires 5.7 billion arithmetic operations and 27 million parameters to be evaluated; and GPT-3~\cite{gpt-3}, an experimental state of the art natural language processing network requires 175 billion parameters (350 GiB assuming 16 bits per parameter) to be evaluated.
Furthermore, training such deep neural models becomes increasingly expensive and the largest language models already require supercomputers for training, potentially costing millions of dollars per training run~\cite{gpt-3}. Thus, it is important to investigate sparsity during the training process to manage the costs of training.

The results we survey show that \emph{today's sparsification methods can lead to a 10-100x reduction in model size, and to corresponding theoretical gains in computational, storage, and energy efficiency, all without significant loss of accuracy}. 
If those speedups are realized in efficient hardware implementations, then the gained performance may lead to a phase change in enabling more complex and possibly revolutionary tasks to be solved practically. 
Furthermore, we observe that the pace of progress in sparsification methods is accelerating, such that even during the last months while we worked on this report, several new methods that improve upon the state of the art have been published.

\parad{style}
We aim to provide an overview of the key techniques and ideas, while covering some of the necessary mathematical background. Due to space constraints, we keep our descriptions brief---we always refer the interested reader to the original papers which describe the ideas in full detail. 
We structure the discussion along various axes: which elements of a neural network are sparsified, when are they sparsified, and how can they be sparsified. Furthermore, we consider sparse training and the need to re-add connections during training to maintain a constant model complexity after sparsification. We also outline the development of results in various areas of sparsification.

\parad{state of community}
In general, the flurry of different techniques, tasks, models, and evaluation settings causes a wide spread in the community. This leads to many incomparable results and makes it hard to determine the state of the art and whether method A is better than method B. Furthermore, we found that nearly every basic approach has been invented at least twice. \citet{2020-blalock} also point at these problems and they propose a common benchmark and methodology to go forward. We aim at summarizing the existing techniques, and first focus on purely qualitative aspects of designing models in Sections~\ref{sec:overview}-\ref{sec:dynamic}. Then, in Sections~\ref{sec:architectures} and \ref{sec:speed}, we explain a selection of architectures implementing combinations of those designs including performance results. Sections~\ref{sec:discussion}-\ref{sec:conclusions} provide a general discussion, list open problems, and conclude the overview. 

\subsection{Overview of Model Compression Techniques}
\label{sec:compress}

We first present the landscape of approaches to compress models in order to improve computational and memory efficiency. We differentiate between six main techniques:
\begin{itemize}
	\item \textbf{Down-sizing models} creates smaller dense networks to solve the same task. Model distillation~\cite{hinton2015distilling} or Neural Architecture Search~\cite{elsken2019neural} are typical examples of techniques to find small dense models.
	\item \textbf{Operator factorization} decomposes operators, for example the matrix multiplication of dense layers, into smaller operators. For matrices, operators can be decomposed via singular value decomposition~\cite{6638949}, while more  general tensors can be decomposed via tensor train decomposition~\cite{zhao2017learning,1993-kanjilal}. 
	\item \textbf{Value quantization} seeks to find a good low-precision encoding for values in the networks, such as weights, activations, or gradients. Various floating point and integer formats can be used to encode data efficiently leading to a smaller number of bits than standard 32 or 64 bit datatypes. 
	\item \textbf{Value Compression} can be used to compress model structures and values (e.g., weights) either with generic entropy-based methods~\cite{2015-han} or loss-bounded type-specific methods using correlation across values~\cite{2019-jin}. 
	\item \textbf{Parameter sharing} can lead to model compression by exploiting redundancy in the parameter space. Such redundancy can also be fostered during the training process~\cite{plummer2020shapeshifter}. 
	\item \textbf{Sparsification} can lead to more efficient models that continue to operate in high-dimensional feature spaces but reduce the representational complexity using only a subset of the dimensions at a time. Practically, such methods can reduce complexity by zeroing out subsets of the model parameters. 
\end{itemize}

All of these methods lead to reduced memory requirements and all schemes, except for  parameter sharing, can also reduce the computational complexity. 
These schemes can be combined into an efficient inference and training approach and various surveys cover subsets of this space in detail~\cite{9043731,choudhary2020comprehensive,cheng2020survey}. 
In this paper, we focus on the most complex and, in our view, most powerful of those techniques: sparsification, also known as ``pruning'' in some contexts. 
\citet{1993-reed} provides an overview of early sparsification techniques until 1993---since then, the literature has evolved significantly.
A second ``AI winter'' in the late 1980s and early 1990s appears to have significantly reduced interest in and funding for artificial intelligence research and development~\cite[Sec.~1.3]{2020-russell},~\cite[Sec.~24.4]{2009-nilsson}, and activity in neural networks subsequently waned for nearly two decades.
Deep learning (re-)started its success story around 2012 with convolutional neural networks for image recognition. Since then, more than
266
papers, comprising 
%
4,089
pages focusing on ideas and techniques for sparsity in deep networks appeared, which we categorize and summarize below. We aim to provide an intuitive and comprehensive overview of the most important ideas. Yet, at a compression rate of 97.9\% and more than 420 citations, we almost surely miss specific ideas or works. 

\begin{figure}[h!]
	\includegraphics[width=0.6\textwidth]{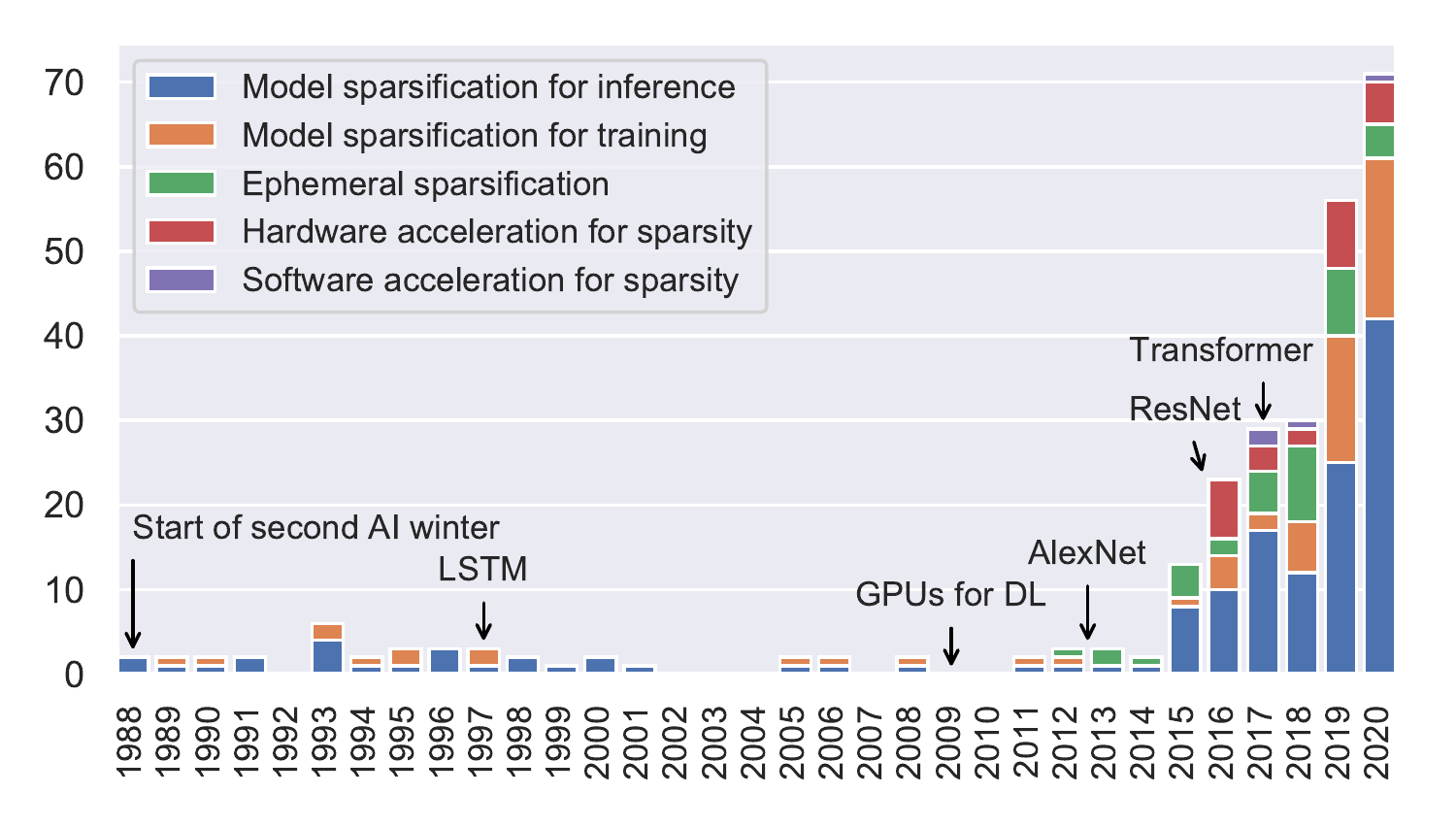}
	\caption{Literature development over the years.}
	\label{fig:papers}
\end{figure}

Fig.~\ref{fig:papers} shows the volume of scientific publications on various aspects of sparsity over the last three decades. The first papers in the late 80's and 90's focus on very small models and their generalization and interpretability properties. The whole field of neural networks was rather inactive during the early 2000's until the breakthroughs in image recognition circa 2012, followed by a resurgence of interest in optimization of sparse networks. During the late 2010's, numerous accelerators and optimization techniques were designed to specifically aim at optimizing sparse deep neural networks. The meaning of the labels will be clarified later in this paper.

\parad{sparsity as big opportunity after ending of Moore's law}
One of the main drivers behind the massive progress in deep learning between the 90's and today was the nearly 1 million times increase in computational capability delivered by Moore's law, Dennard scaling, and architectural specializations with GPUs and specialized machine learning accelerators. 
With the ending of those scaling laws and specialization opportunities, these developments will hit their natural limits and progress may stall.
\emph{We see sparsity as potentially achieving a second significant ``jump'' in computational capability as, even with current methods, it has the promise to increase computational and storage efficiency by up to two orders of magnitude.}
 
\subsubsection{Document structure}
We aim to provide a comprehensive overview to a diverse set of readers. Section~\ref{sec:math-notation} introduces all mathematical background for the different sparsification approaches and can be skipped by experienced readers as well as readers that are mostly looking for intuition. 
Section~\ref{sec:overview} provides an executive summary of how pruning schemes work.
Sections~\ref{sec:removal} and~\ref{sec:growth} dive deeply into different schemes for removal and growth (weight addition) during training and pruning while 
Section~\ref{sec:dynamic} describes details of various ephemeral (per example) sparsification schemes.
We consider examples of pruning for full convolutional and transformer architectures in Section~\ref{sec:architectures}.
Section~\ref{sec:speed} overviews various approaches for improving the performance of sparse models, ranging from software to specialized hardware implementations. 
In Section~\ref{sec:discussion}, we summarize and extrapolate the most significant observations in the field and we provide ten research challenges in Section~\ref{sec:challenges}.

If your goal is to get a quick executive overview of the field, then we recommend studying Sections~\ref{sec:overview} and \ref{sec:discussion} while skimming Sections~\ref{sec:removal},~\ref{sec:growth},~\ref{sec:dynamic}, and~\ref{sec:speed}, especially the overview figures and tables therein.
If your main interest lies in the hardware engineering aspects, then we recommend to at least get the executive overview mentioned before and study Section~\ref{sec:speed} in detail. Similarly, if you are a neural network architect looking for sparsification best practices, we recommend the executive overview in combination with details in Section~\ref{sec:architectures} and the references therein.
Researchers in the field may want to examine the whole document carefully to get a deep overview of all aspects and focus efforts especially on the challenging problems in Section~\ref{sec:challenges}. 
Finally, readers can get a view of each section from the first 1--2 paragraphs to decide whether to dive deeper into each subject.

\subsection{Background and Notation}\label{sec:math-notation}
\parad{informal overview of deep learning models}
We start by providing some background on deep learning inference and training to introduce our notation. Experienced readers may wish to skip to the next section.
Deep learning models (or ``networks'') consist of a graph of parameterizable layers (or ``operators'') that together implement a complex nonlinear function $f$. 
We consider a general supervised learning setting, where we are given a training set, comprised of pairs of input examples $\vec{x} \in \mathcal{X}$ and outputs $\vec{y} \in \mathcal{Y}$. The goal is to learn the function $f: \mathcal{X} \mapsto \mathcal{Y}$, parameterized by weights $\vw \in \R^d$, such that given input $\vec{x}$, the prediction $f(\vec{x}; \vec{w})$ is close to $\vec{y}$. 
We usually assume that $\mathcal{X}$ represents a vector of features describing an element drawn from a true input distribution $\mathcal{D}$ that captures the characteristics of typical inputs but cannot be measured or described concisely (e.g., cat pictures). 
Applying the function $f(\vec{x}; \w )$ is performed by transforming the input $\vec{x}$ layer by layer to generate the output - this process is called \emph{inference}, or in a training setting the \emph{forward pass}. 

\parad{informal overview of deep learning}
The process of finding a network to solve a specific task can be decomposed into two phases: (1) \emph{design} or \emph{engineer} the network structure, and (2) \emph{train} the network's weights. The network structure is traditionally designed manually and not changed during the training process. 
Training iterations start with a forward pass, which is similar to inference but stores the inputs of each layer. 
The quality of the result $f(\vec{x}; \w )$ of the forward pass is evaluated using a loss function $\ell:\mathcal{Y} \times \mathcal{Y} \mapsto \mathbb{R}$ to estimate the accuracy of the prediction, $\ell \big( \vec{y}, f \left(\vec{x}; \vec{w}\right)\big)$, where $(\vec{x}, \vec{y})$ is the sample pair. 
Many loss functions are known, such as the $L_2$ distance or the cross-entropy between the predicted output $f\left(\vec{x}; \vec{w}\right)$ and the expected one $\vec{y}$. 
The following backward pass propagates the loss (``\emph{error}'') from the last layer in the reverse direction. At each learnable (parametric) layer, the backward pass uses the adjoint of the forward operation to compute a gradient $g$ and update the parameters (``\emph{weights}'') using a \emph{learning rule} to decrease $\ell$ (for the current example pair). 
This method is repeated iteratively for many different examples drawn from $\mathcal{D}$ until the function $f(\vec{x}; \w )$ provides the desired accuracy. This accuracy is typically evaluated on a separate set of examples that were not used to train the model in order to measure the \emph{generalization} capabilities of the model to unseen examples drawn from $\mathcal{D}$.

\begin{figure}[h!]
	\begin{subfigure}{.55\textwidth}
		\centering
		\includegraphics[width=0.9\linewidth]{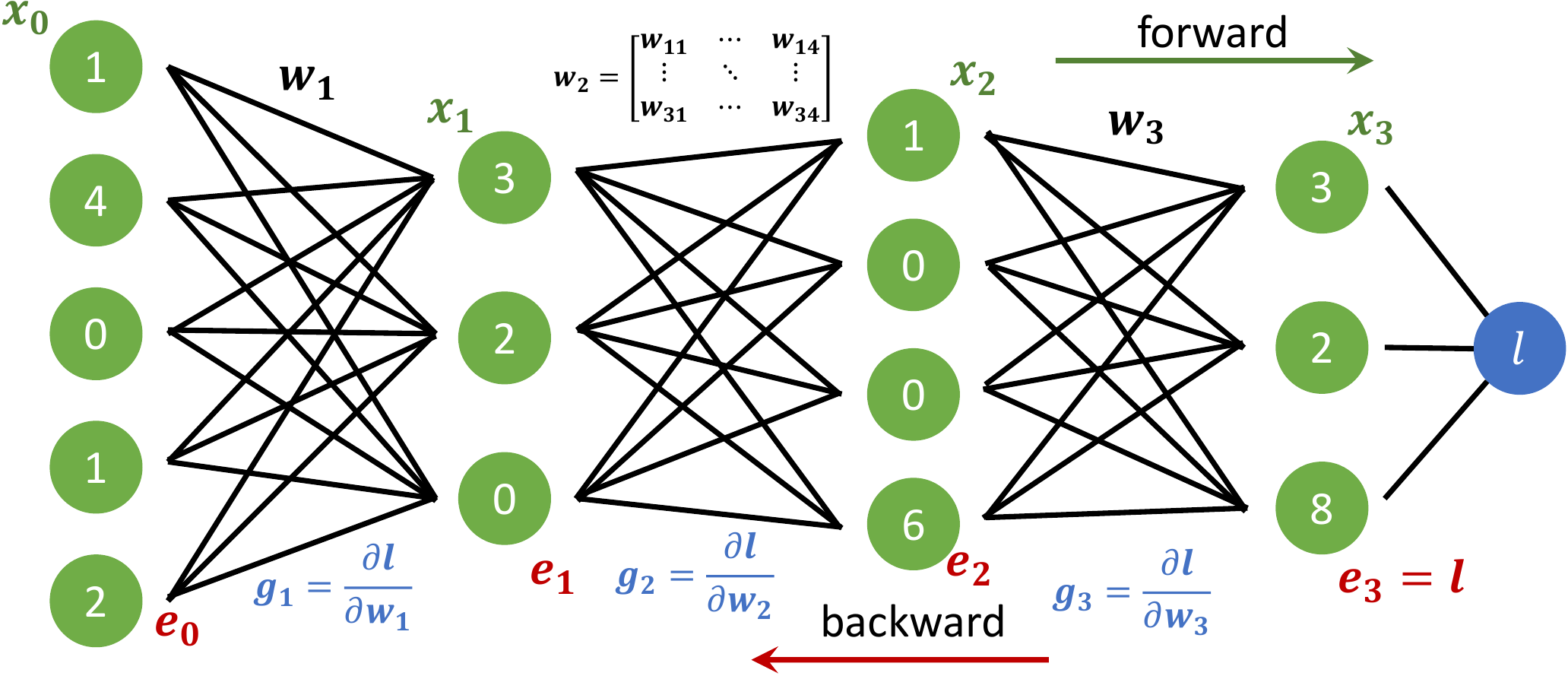}  \hfill
		\caption{Dense MLP training example}
		\label{fig:network_overview}
	\end{subfigure}
	\begin{subfigure}{.4\textwidth}
		\centering
		\hfill\includegraphics[width=0.95\linewidth]{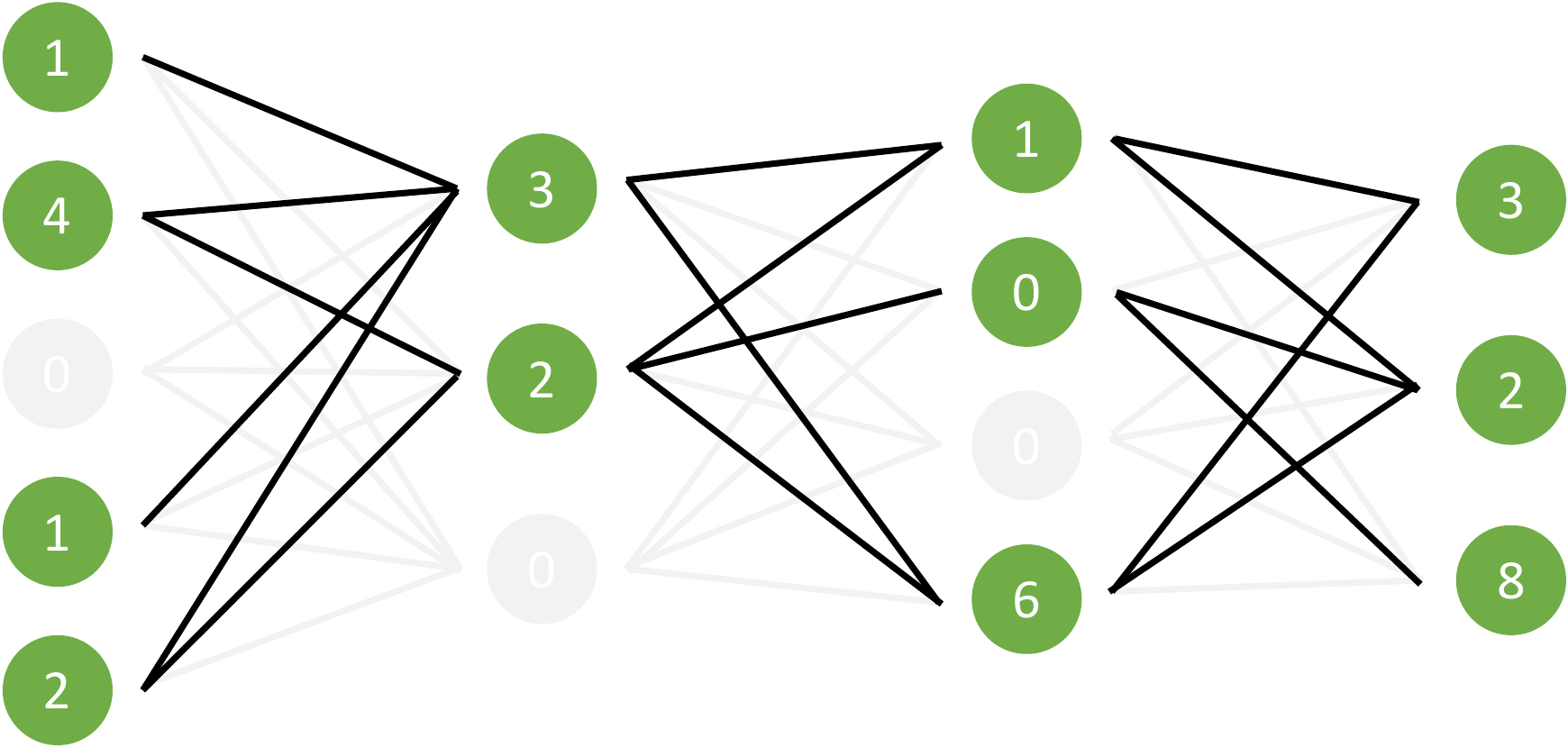}  
		\caption{Sparsified MLP}
		\label{fig:sparse_mlp}
	\end{subfigure}
	\caption{Training, Inference, and Sparsification examples.}
	\label{fig:fig}
\end{figure}
We now introduce some further notation and mathematical background, which will be useful to understand some of the following pruning schemes. Parts of this section follow the notation and general approach of~\cite{2016-molchanov, 2020-singh}. 

\parad{formalize forward pass for MLP}
Let us consider the simple case of a multilayer perceptron shown in Fig.~\ref{fig:network_overview} with the typical input layer $\vec{x_0}$, two hidden layers $\vec{x_1}, \vec{x_2}$, and output layer $\vec{x_3}$, with rectified linear units (ReLU) $\sigma_R(x):=\max\left(0,x\right)$ as \textit{activation} functions. We denote the number of neurons in layer $i$ as $|\vec{x_i}|$. The forward pass can be written as a series of matrix-vector products $f(\vec{x_0}; \w ) = \sigma_R(\vec{w_3}\cdot \sigma_R(\vec{w_2}\cdot \sigma_R(\vec{w_1} \vec{x_0} + \vec{b_1}) + \vec{b_2}) + \vec{b_3})$, where $\vec{x_0}$ is the input (``feature'') vector.
Here, the network function $f(\vec{x_0}; \w )$ is parameterized by weight matrices $\vec{w_1}$ with dimensions $|\vec{x_0}|\times|\vec{x_1}|$, $\vec{w_2}$ with dimensions $|\vec{x_1}|\times|\vec{x_2}|$, and $\vec{w_3}$ with dimensions $|\vec{x_2}|\times|\vec{x_3}|$; and bias vectors $\vec{b_i}$ with dimensions $|\vec{x_i}|$ for layer $i$ (we usually omit biases for brevity). 
Subscripts identify the layer---we omit them for equations that apply to all layers. Sometimes, we treat the concatenation of all weight matrices as a vector---this will be obvious from the context. It is already apparent that the $\mathcal{O}(|\vec{x_i}|\cdot|\vec{x_{i+1}}|)$ storage and compute may overparameterize the model for a large number of neurons. 

Fig.~\ref{fig:sparse_mlp} shows a sparsified version of Fig.~\ref{fig:network_overview}. It shows that the third input feature and all its adjacent weights are removed (grayed out). Furthermore, two hidden neurons and their weights as well as various other weights have been removed. Removing neurons or input features corresponds to removing rows or columns in the layer weight matrices while single weights remove elements of the matrices.

\subsubsection{The Deterministic Formulation for Training}  
Training a deep neural network minimizes a loss. In the deterministic case, the (empirical) training loss $L$ is defined as the average loss over training examples, i.e., $L(\vec{w})=\frac{1}{N} \sum_{n=1}^{N} \ell\big(\vec{y}[n], f\left(\vec{x}[n]; \vec{w}\right)\big)$. In the following, we fix $d \geq 1$ to be the total number of parameters in the model, and we will omit the indexing by sample when clear from context. The loss function $L$ will be a $d$-dimensional loss function over the model parameters $L: \mathbb{R}^d \rightarrow \mathbb{R}$. 

The most common training scheme is stochastic gradient descent (SGD), which is based on a first-order approximation to the loss function $L$. 
This method utilizes automatic differentiation (AD) to compute the derivative (``gradient'') of the loss with respect to the weights in a layer $\vec{g_1}=\frac{\partial L}{\partial \vec{w_1}}$ and $\vec{g_2}=\frac{\partial L}{\partial \vec{w_2}}$ at the specific example $\vec{x}$.
Reverse mode (aka.~adjoint) AD stores the intermediate results of the forward pass and applies the loss function $L$ that returns an error (``distance'') with respect to the desired model output.
This can be done by applying the chain rule to the compound function $f(\vec{x}; \w )$ and propagating the error backward through all operators. For example, the gradient of the second layer is $\vec{g_2} = \frac{\partial L}{\partial \vec{w_2}}=\frac{\partial L}{\partial \vec{e_2}}\frac{\partial \vec{e_2}}{\partial \vec{w_2}}$. The gradients are then used with a learning rule function $R$ to update the weights for the next iteration: $\vec{w}^{(i+1)} = \vec{w}^{(i)} + R(\vec{g},\vec{w}^{(i)})$.

\paragraph{The Jacobian matrix.} 
The Jacobian of an arbitrary function $F : \mathbb{R}^d \rightarrow \mathbb{R}^m$ is the matrix of first-order partial derivatives of a vector-valued function with respect to its inputs. For example, the Jacobian matrix for the loss function $L: \mathbb{R}^d \rightarrow \mathbb{R}$ with respect to the weights is a $1\times d$ matrix of partial derivatives with respect to each individual weight. If we write $w_1$ for the first individual weight and $\vec{w_1}$ for the set of weights in the first layer (similarly for gradients), then the Jacobian matrix is defined as $\mathbf{J}=\nabla_\vw L=
\left[ \frac{\partial L}{\partial w_1} \, \frac{\partial L}{\partial w_2} \cdots \frac{\partial L}{\partial w_d}  \right] = \left[ g_1 g_2 \ldots g_d \right]$.
More generally, the Jacobian also arises when we consider the matrix of partial derivatives for a specific layer's outputs with respect to its inputs. 
Intuitively, the Jacobian matrix encodes the rate of change of a given vector-valued function's outputs with respect to its inputs. 

\paragraph{{The Hessian matrix.}} For a twice-differentiable loss $L$, the Hessian matrix is the matrix of second-order derivatives of the loss function with respect to the weights, mathematically expressed as $\hess=\nabla^2_\vw L$. Intuitively, its role is to express the local geometry (``curvature'') of the loss around a given point $\vw$, leading to a faithful approximation of the function in a small neighborhood $\delw$ around the point $\vw$. The second-order approximation of the function, which includes the first-order (gradient) term and the second-order (Hessian) term, is also referred to as the \emph{local quadratic model} for the loss. Following the Taylor expansion, where we assume that the higher-order terms are negligible, this leads to the approximation 
$$L(\w + \delw) \approx  L(\w) + {\gradw}\, \delta \mathbf{w}+\frac{1}{2} \delta \mathbf{w}^{\top} \mathbf{H} \; \delta  \mathbf{w}.$$
For clarity, note that here we take $\w$ to be a column vector, which implies that each term in the above expression is a scalar. 

\subsubsection{The Probabilistic Formulation} 
The above deterministic formulation inherently assumed a deterministic ``correct'' output label corresponding to each input example. 
However, it is just as reasonable to consider that each input example $\vec{x}$ has some probability of being assigned a given label $\vec{y}$, rather than the output being fixed.

We can formalize this intuition following~\citet{2015-martens}.  Given input examples $\vec{x} \in \mathcal{X}$ and outputs $\vec{y} \in \mathcal{Y}$, 
we assume that input vectors $\vec{x}$ are drawn from a distribution $Q_x$, and that the corresponding outputs $y$ are drawn from a conditional distribution 
$Q_{y|x}$, leading to an underlying joint probability distribution defined as $Q_{\vx,\vy} = Q_{\vx} \,Q_{\vy| \vx}$. 
We will assume that the marginal probability distribution over input samples $Q_{\vx}$ is well-approximated by the empirical distribution $\widehat{Q}_\vx$ over the inputs in our training set. 
Intuitively, this means that we trust the sampling distribution used to generate the input dataset to be representative of the true distribution. 

In this context, the goal of learning is to minimize the distance  between the target joint distribution $Q_{\vx, \vy}$, and a learned joint  distribution $P_{\vx, \vy}(\vw)$, where $\vw$ is the model.  It is standard for this distance to be measured in terms of the Kullback-Leibler (KL) divergence between distributions. Alternatively, we can cast this as the task of predicting the output $\vy$ given an input $\vx$, i.e., training a model $\vw$ to learn the conditional distribution $P_{\vy|\vx}(\vw)$, where $P_{\vy|\vx}(\vw)$ is the probability of a given output given a certain input, which should be close to the true distribution $Q_{\vy| \vx}$. In the following, we omit the explicit dependency of $P_{\vy|\vx}$ on $\vw$ when clear from context. 
In this formulation, we can obtain an equivalence between the standard loss we considered above and the negative log-likelihood of the probability density function corresponding to the output distribution of the model with parameters $\vw$, which we denote by $p_\vw$. Formally, for a sample  $(\vx_n \vy_n)$ in the probabilistic formulation, we have:

$$\ell\big(\vy, f(\vx;\vw)\big)  =- \log \big(p_\vw(\vy| \vx)\big).$$

\paragraph{The Fisher Matrix.} Intuitively, the role of the Fisher matrix is very similar to that of the Hessian matrix, but in the \emph{probabilistic} setting, where our notion of distance is the KL divergence between the model's output distribution and the true output distribution. 
More precisely, assuming the probabilistic view, the Fisher information matrix $F$~\cite{ly2017tutorial} of the model's conditional distribution $P_{y|x}$ is defined as 
\begin{equation}\label{eq:fisher}
F=\mathrm{E}_{P_{\vx, \vy}}\left[\nabla_\vw \log p_{\vw}(\vx, \vy) \, \nabla_\vw \log p_{\vw}(\vx, \vy)^{\top}\right]\,.
\end{equation}
It can be proved that the Fisher matrix in fact satisfies $F=\mathrm{E}_{P_{\vx, \vy}}\left[- \nabla^2_\vw \log p_\vw(\vx, \vy)\right]\,$. Matching the original intuition, we can express \(P_{\vy, \vx}=Q_{\vx} P_{\vy | \vx} \approx \widehat{Q}_{\vx} P_{\vy | \vx}\). 

Further, it is known~\cite{ly2017tutorial} that, if the model's output conditional distribution $P_{\vy|\vx}$ matches the conditional distribution of the data $\widehat{Q}_{\vy|\vx}$, then the Fisher and Hessian matrices are in fact equivalent. 
In practical terms, this means that, if $\vw$ is an accurate set of parameters for the model, we can approximate the Hessian matrix at $\vw$ with the Fisher matrix. 
In turn, this is useful since the Fisher matrix can be more efficiently approximated, as we will see below.  

\paragraph{The Empirical Fisher.} In practical settings, it is common to consider an  approximation to  the Fisher matrix introduced in Eq.~\eqref{eq:fisher}, where we replace the model distribution $P_{\vx, \vy}$ with the empirical training distribution $\widehat{Q}_{\vx, \vy}$. Then we can simplify the expression of empirical Fisher $\hat{F}$ as follows, 
{
	\begin{equation*}
	\begin{aligned}
	\hat{F} &=\mathrm{E}_{\widehat{Q}_{\vx}}\left[\mathrm{E}_{\widehat{Q}_{\vy | \vx}}\left[\nabla \log p_{\vw}(\vy | \vx) \nabla \log p_{\vw}(\vy | \vx)^{\top}\right]\right] 
	\stackrel{(a)}{=} \frac{1}{N} \sum_{n=1}^{N} \underbrace{\nabla \ell\left(\vy_{n}, f\left(\vx_{n}; \vw\right)\right)}_{\nabla \ell_n} \nabla \ell\left(\vy_{n}, f\left(\vx_{n}; w\right)\right)^{\top}, 
	\end{aligned}
	\end{equation*}
}%
where (a) uses the equivalence of the loss between the probabilistic and deterministic settings. 
In the following discussion, we will use a shorthand $\ell_i$ to denote the loss for a particular training example $(\vx[i], \vy[i])$, and refer to the \emph{true Fisher} when describing the matrix defined in Eq.~\eqref{eq:fisher}. Thus, the above formula describes a fairly popular approximation, which equates the Fisher matrix with the  \emph{empirical} Fisher. For a more detailed exposition on various aspects of this topic, we refer the reader to \citet{2015-martens, ly2017tutorial, 2019-kunstner, 2020-singh}. 

\subsubsection{The Bayesian Formulation} \label{sec:bayesian}
We now provide a brief primer on Bayesian inference, which will be useful to understand the variational pruning approaches presented in the later sections. 
Our presentation follows~\cite{2017-molchanov}. 

We start from the probabilistic formulation above, in which, given a dataset $S=\lbrace (\vx[i], y[i] ) \rbrace_{i=1}^N$ our goal is to identify a set of parameters $\w$ which approximates the ``correct'' distribution of outputs $p(y[i] | \w, \vx[i])$ for any given input $\vx[i]$. 
In Bayesian learning, it is assumed that we have some prior knowledge on $\w$, in the form of a \emph{prior} distribution over models, $p(\w)$. 
After observing some of the data, we can form the \emph{posterior} distribution by following Bayes' rule 
\[ p(\w | S) = p( S \, | \w) p(\w) / p(S).\] 
This process is called Bayesian Inference. However, computing the posterior distribution is often not possible in practice, as it requires computing the marginal likelihood $p(S)= \int p(S | \w) p(\w) d\w$, which is an intractable integral for most complex models. Therefore, certain simplifying assumptions are usually made, to enable an efficient approximation of the posterior distribution.

One specific technique for Bayesian Inference that relies on such simplifying assumptions is Variational Inference.
Here, the posterior distribution $p(\w | S)$ is approximated by a parametric distribution $q_{\phi} (\w)$.
The quality of this distributional approximation is measured in terms of the KL divergence $D_{{KL}}(q_{\phi}(\w) \| p(\w | S))$, and the task of finding $p(\w | S)$ is translated into an optimization problem in the space of variational parameters $\phi$. 
In this context, the optimal value of $\phi$ can be found by maximizing the following \emph{variational lower bound} of the marginal log-likelihood of the data:

\begin{equation}
\mathcal{L}(\phi) = \sum_{i = 1}^{N} \mathbb{E}_{q_{\phi}}[\log p(y[i] | \vx[i], \w)] - D_{{KL}}(q_{\phi}(\w) \| p(\w)).
\label{eq:elbo}
\end{equation}

The first term is called the expected log-likelihood, which is often denoted by $L_S(\phi)$, representing the model's loss, whereas the second term acts as a regularizer, enforcing that the parametric distribution $q_{\phi}(\w)$ should stay close to the prior $p(\w)$.

One important issue with the above framework is that, for complex models, optimizing the above variational lower bound is intractable, due to the integration required for computing $L_S(\phi)$. Instead, it is common to estimate $L_S(\phi)$ by sampling, and optimize the lower bound stochastically. A series of technical advances generally known as ``reparametrization tricks'' allow to obtain unbiased, differentiable, minibatch-based Monte-Carlo estimators of the expected log-likelihood term above for large-scale models.\footnote{The main idea is to represent the parametric noise $q_{\phi}(\w)$ as a deterministic differentiable function $\w = g(\phi, \varepsilon)$ of some non-parametric noise $\varepsilon \sim p(\varepsilon)$. 
	This trick allows one to obtain an unbiased estimator of the gradient of the log-likelihood term, $\nabla L_S(\phi)$.} We refer the interested reader to~\cite{2013-kingma, 2014-rezende, 2015-kingma, 2017-molchanov} for details. 

\paragraph{Variational Dropout.} 
To illustrate this technique, we will use the same notations as \cite{2017-molchanov} and consider a single fully-connected layer with $I$ input neurons and $O$ output neurons before the non-linear activation function. 
Taking $M$ to be the minibatch size, we denote the ${M\times O}$ output matrix by $B$, the ${M\times I}$ input matrix as $A$, and the $I\times O$ layer weight matrix as $W$. Notice that $B = A W$. 

Dropout~\cite{2012-hinton} is a popular regularization method for neural networks, which injects multiplicative random noise $\Xi$ to the layer input $A$, at each iteration of the training procedure. Mathematically, 
\[
	B = (A\odot \Xi)W, 
\]
where the entries of $\Xi$ denoted by $\xi_{mi}$ follow a given distribution $p(\xi)$. 
The original variant of dropout used a constant parameter $p \in (0, 1)$ called \emph{dropout rate}, and drew the random variables as $\xi_{mi} \sim \text{Bernoulli}(1-p)$. 

\citet{2014-srivastava} reported that Gaussian dropout, where the noise is drawn from a continuous distribution $\xi_{mi} \sim \mathcal{N}(1, \alpha = \frac{p}{1-p})$, 
works as well as the discrete counterpart. Interestingly, this procedure has a non-trivial Bayesian interpretation, as was shown in \cite{2015-kingma}.
 
Specifically, applying Gaussian noise $\xi_{mi} \sim \mathcal{N}(1, \alpha)$ to a weight $w_{ij}$ is equivalent to sampling the weight's value from a parameterized normal distribution centered at $w_{ij}$, denoted as $q ( w_{ij} \, | \, \theta_{ij}, \alpha ) \sim \mathcal{N}( w_{ij} | \theta_{ij}, \alpha \theta_{ij}^2 )$. 
Thus, instead of viewing each $w_{ij}$ as a parameter, each weight can be seen as a random variable parameterized by $\theta_{ij}$, which controls the weight's variance. 

Following this interpretation, Gaussian Dropout training can be seen as equivalent to standard stochastic optimization of the expected log-likelihood over the parameters $\theta_{ij}$, in the special case where we draw a single sample of the weights $W \sim q(W | \theta, \alpha)$ per minibatch to estimate the expectation, and where we use a log-uniform prior distribution over the weights. 
Sparse Variational Dropout \cite{2017-molchanov} extends this idea, and explicitly uses $q(W|\theta, \alpha)$ as an approximation for the posterior distribution. 
Thus, the parameters $\theta$ and $\alpha$ of the distribution $q(W|\theta, \alpha)$ can be optimized via stochastic variational inference. This means that $\phi = (\theta, \alpha)$ are the so-called \emph{variational parameters}, introduced above. To avoid the problem of high variance of stochastic gradients for large values of $\alpha_{ij}$ as reported in \cite{2015-kingma}, \citet{2017-molchanov} introduce an \emph{additive noise reparameterization}, in which the optimization is done directly over $(\theta, \sigma^2)$, with $\sigma^2=\alpha \theta^2$, instead of $(\theta, \alpha)$. 

Practically, Variational Dropout provides a way to train the dropout rate $\alpha$ by optimizing the variational lower bound we introduced above. 
Interestingly, however, the dropout rate becomes a variational parameter to be optimized, and not a simple hyper-parameter. This allows one to train individual dropout rates for each layer, neuron, or even weight. While the basic technique was introduced by \citet{2015-kingma}, it was \citet{2017-molchanov} who first investigated the effects of training individual dropout rates, and showed that Variational Dropout can effectively sparsify DNNs. We discuss this latter paper and its follow-ups in Section~\ref{sec:varbayes}.

\subsubsection{Convolutional Layers as Designed Sparsity}
\label{sec:conv}
\begin{figure}[h!]
	\includegraphics[width=0.9\textwidth]{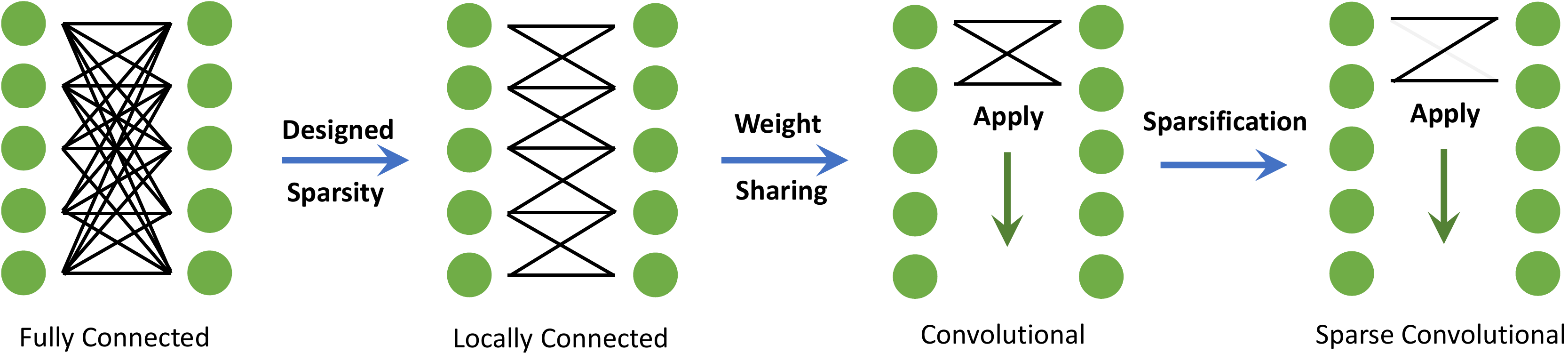}
	\caption{Convolutional operators as sparse fully-connected operators for a single input and output channel.}
	\label{fig:conv}
\end{figure}

Particularly common in deep learning are convolutional operators. Convolutions perform a weighted average over local regions of neurons, incorporating local information and reducing the number of weights at the same time. Convolutional Neural Networks (CNNs) have been proven to be highly successful for image classification~\cite{2016-he}, segmentation~\cite{2017-he-mask}, and many other tasks. 
The convolution operator itself and its variants can be seen as a sparse version of fully connected layers (Fig.~\ref{fig:conv}).
Instead of connecting every pair of neurons in the input and output layers, we prune the connections to contain only local surroundings based on the operator's convolution kernel size, strides, padding, and other factors such as dilation. 
The new operator contains a unique \emph{filter} for each output neuron, also known as a Locally Connected Network (LCN)~\cite{2010-ngiam}, which are used for specializing filters for different spatial regions~\cite{2020-groenquist} as shown in the 2nd part of Fig.~\ref{fig:conv}.
\citet{1996-olshausen} even argue that sparsity is essential property to encode vision operations. 

In order to provide translational equivariance, these operators are sparsified yet again by way of \textit{weight sharing}, reusing the local filters in each output neuron as shown in the 3rd part if Fig.~\ref{fig:conv}. In a typical convolutional layer, the input is divided into $C_{in}$ ``channels'' and the output into $C_{out}$ channels or ``features'', multiplying and summing each input channel with a unique set of $C_{out}$ filters. This yields the formula for the convolutional operator: $o_{j,k,l}=\sum_{m=0}^{C_{in}-1}{\sum_{k_y=0}^{K_y-1}{\sum_{k_x=0}^{K_x-1}{x_{m,k+k_y,l+k_x} \cdot W_{j,m,k_y,k_x}}}}$ for a filter size $K_x\times K_y$. Fig.~\ref{fig:conv} shows only one input channel and one output channel for simplicity. 
As we will discuss in the following, further sparsity can be introduced in CNNs, as well as other DNN classes as shown in the last part of Fig.~\ref{fig:conv}.

\section{Overview of Sparsity in Deep Learning}\label{sec:overview}

The utility of sparsification lies in two very different areas: (1) improved generalization and robustness and (2) improved performance for inference and/or training. 
We now provide a general overview of sparsification in deep learning, starting with an observation of typical sparsity-accuracy tradeoffs.
We then discuss sparse storage formats, a taxonomy of element removal, and sparsification schedules. 
All discussions apply to both inference and training.

\subsection{Generalization}

\begin{figure}[h!]
	\includegraphics[width=0.6\textwidth]{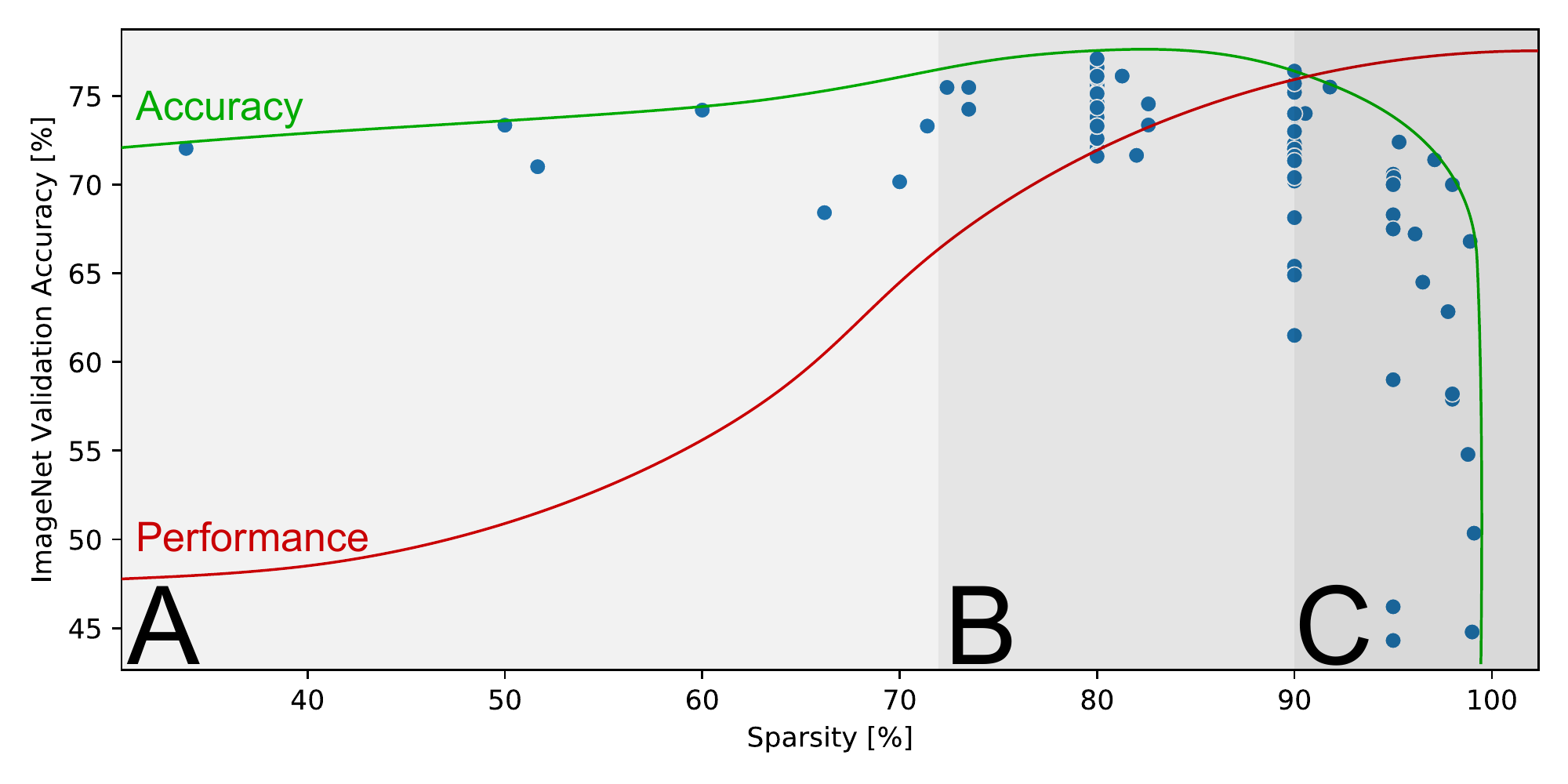}
	\caption{Typical test error vs. sparsity showing Occam's hill (network: ResNet-50 on Top-1 ImageNet).}
	\label{fig:generalization_err}
\end{figure}

Generalization performance is one of the most important aspects of a deep learning model. It measures how well the model performs for unseen data drawn from the same distribution as the training data but was not used for training. 
Most, if not all, sparsification follows Occam's hill~\cite{rasmussen2001occam} shown as a sketch (green line) in Fig.~\ref{fig:generalization_err}: As we start to sparsify, initially the accuracy increases due to the reduction of learned noise. Intuitively, the smaller model forms a stronger regularizer forcing the learning algorithm to ``focus'' on more important and general aspects of the model (Part A in the figure). Then, the model reaches an often extended range of sparsities where the performance remains stable and maybe slightly decreases (Part B). Eventually, with high sparsity, the quality quickly degrades (Part C).

If we observe the computational performance of the model, we often see a curve similar to the red line in Fig.~\ref{fig:generalization_err}: initially, for low sparsity, performance grows slowly due to overheads in storing sparse structures and controlling sparse computations. Then, for moderate and high sparsity, we see a sustained growth of performance before it usually levels off at extremely high sparsities where storage and control overheads dominate. For most practical purposes and sparsities, the performance increases with growing sparsity, the area of diminishing returns only applies to extreme sparsities which deep learning models have yet to reach. In general, achieving highest performance at a specific sparsity level is complex---most techniques to store and exploit sparsity are only efficient within a limited sparsity interval and/or distribution of non-zero elements. 

\subsection{Performance and model storage}\label{sec:perf}

\parad{intro}
Sparsification reduces the necessary operations to evaluate a model as well as the memory necessary to store the model by removing nonessential elements. In some cases, for example, when whole neurons or filters are removed, we can use associativity and distributivity of linear algebra to transform a sparsified structure into a smaller dense structure. 
However, if we remove random elements of a weight matrix, we need to store the indices of the remaining non-zero elements. 

The storage overheads for indexing $m$ non-zero elements in a space of size $n$ vary from bitmaps with $n$ bits to absolute coordinate schemes using $m\log(n)$ bits. Many different formats cover the whole space and the optimal scheme depends on the sparsity, the structure, and the required access patterns (e.g., streaming, transposed, or random access). More generally, finding space-optimal indexing schemes falls into the class of integer compression problems and hundreds of sparse matrix indexing techniques exist~\cite{10.1145/356616.356618}. Here, we focus on a small illustrative subset.

\begin{figure}[h!]
	\includegraphics[width=0.9\textwidth]{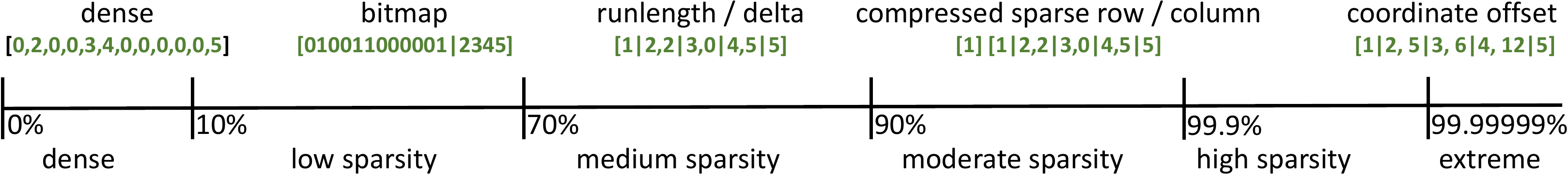}
	\caption{Simple sparse storage formats.}
	\label{fig:sparse_storage}
\end{figure} 
\parad{overview/summary of schemes}
Let us assume we have to store the positions of $m$ elements, each of size $k$ bits in a space of $n$ elements, i.e., $m\leq n$. Fig.~\ref{fig:sparse_storage} overviews a sketch of the schemes described below and shows a range of sparsity where they are most beneficial. The exact scheme depends on many architectural factors and also the exact size of each weight. The simplest scheme stores one bit per element in a \textbf{bitmap} (BM) that stores a map with $n$ bits, each bit indicating whether an element is present. It is efficient for relatively dense structures and requires $o=n$ additional bits.
The next simpler scheme, \textbf{coordinate offset} (COO), stores each non-zero element together with its absolute offset. This scheme lives at the other end of the sparsity spectrum and is most efficient for hyper-sparse structures because it requires $o=m\lceil \log_2n\rceil$ additional bits. 
This offset scheme can be extended with \textbf{runlength encoding} (sometimes also known as \textbf{delta coding}) where only the difference between two elements is stored. If the maximum difference between the indices of two neighboring elements after sorting by index is $\hat{d}$, then those can be encoded with $o=m\lceil\log_2\hat{d}\rceil$ bits. 
If the offsets vary highly, then we could use a \textbf{zero-padded delta offset} scheme where we reduce the bit-width to $\lceil\log_2\bar{d}\rceil$. Here, $\bar{d}<\hat{d}$ represents the expected difference---for all elements that are larger than $\bar{d}$ apart, we add zero values in $\bar{d}$ intervals. The overhead now depends on the distribution of distances and this scheme works best when little padding is necessary. 

In the high-sparsity regime, schemes known from scientific and high-performance computing such as \textbf{compressed sparse row} (CSR), \textbf{compressed sparse column} (CSC), and more general fiber-based schemes can store indices of matrices and tensors, respectively. 
We exemplify these \textbf{dimension-aware schemes} using CSR: CSR represents the indices in an $n=n_c\times n_r$ matrix using column and row index arrays. The column array is of length $m$ and stores the column indices of each value in $\lceil\log_2n_c\rceil$ bits. The row array is of length $n_r$ and stores the offsets of each row in the value array in $\lceil\log_2m\rceil$ bits. The overhead is $o=m\lceil\log_2n_c\rceil + n_r\lceil\log_2m\rceil$ and other dimension-aware schemes are similar.
 
\parad{example}
Let us consider an example with $n_c=n_r=10^4 \rightarrow n=10^8$, $k=8$ and $m$ ranging from 100-0\%. The storage overhead for bitmaps is lowest for rather dense representations. No sparse storage scheme offers benefits for less than 10\% sparsity. The bitmap index fares best between 10-70\% sparsity and the delta encoded scheme (assuming $\hat{d}<1000$) is best for sparsity higher than 80\%. The offset index and dimension-aware schemes could work best in very high sparsity and hyper-sparse environments with very high $\bar{d}$ but it is unclear if such high sparsity is to be expected for deep models. The highest sparsity reported in the literature to date is up to 99.9\%~\cite{2020-lin}. 

\subsection{What can be sparsified?}\label{sec:candidates}

\begin{figure}[h!]
	\includegraphics[width=0.9\textwidth]{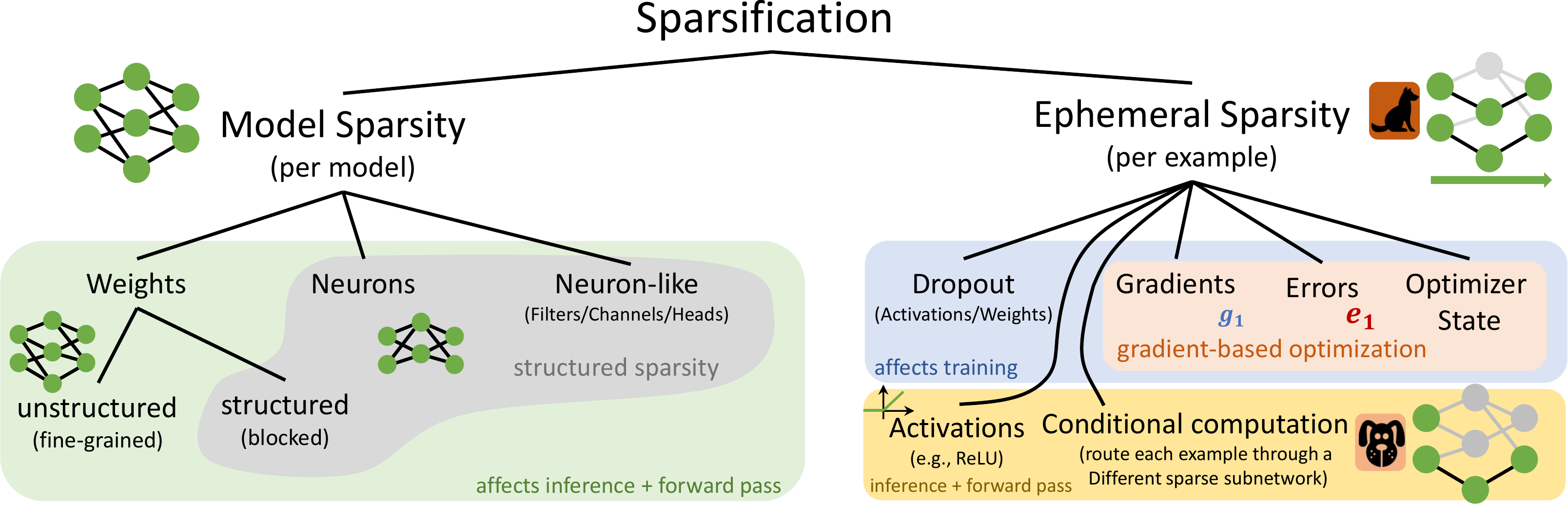}
	\caption{Overview of DNN elements to sparsify.}
	\label{fig:sparsification_elements}
\end{figure}

\parad{intro}
We now provide a summary of which elements of a deep learning model can be sparsified. Fig.~\ref{fig:sparsification_elements} shows an overview. First, we differentiate between \emph{model} (also \emph{structural}) and \emph{ephemeral} sparsification. Model sparsification changes the model and can be considered as a generalization of neural architecture search (NAS). NAS summarizes a class of methods to automatically find better deep learning models and \citet{elsken2019neural} provide an overview. 

\parad{structural sparsification}
Model sparsification changes the model but does not change the sparsity pattern across multiple inference or forward passes. The two main elements, weights and neurons can be sparsified. Elements in specialized layers, such as filters in convolutional layers or heads in attention layers are similar to neurons in the context of pruning and can be removed as well. 
Neuron, filter, and head sparsification reduces simple parameters of the model, can shrink it substantially, and results in a new model that is essentially dense (i.e., can efficiently be executed on the same hardware as the original model)~\cite{2017-sharma}. 
If we sparsify arbitrary weights, the resulting model may be unstructured and we may need to remember indices as described before. This adds overheads for index structures and leads to less efficient execution on hardware that is optimized for dense computations. However, weight sparsification ``is very fine-grained and makes pruning particularly powerful.''~\cite{1997-prechelt}. Thus, approaches for structured weight sparsification have been developed to reduce indexing overheads and improve efficiency of execution. These approaches typically store contiguous blocks of the weights instead of single elements. 
We overview model sparsification techniques in Sections~\ref{sec:removal} and~\ref{sec:growth}.

\parad{ephemeral sparsification}
Ephemeral sparsification is a second class of sparsification approaches---it is applied during the calculation of each example individually and only relevant for this example. The most obvious structural sparsification applies to activations---in fact, the well-known ReLU and SoftMax operators lead to a natural sparsification. Both set values to zero by a fixed threshold (rounding in case of SoftMax). One can also consider random activation sparsity as in dropout~\cite{10.5555/2627435.2670313} (see Section \ref{sec:dropout}) or top-$k$ sparsification as used in~\cite{makhzani2015winnertakeall,ahmad2019dense}. A second set of ephemeral sparsity elements are related to the gradient-based training values. The back-propagation phase of SGD uses errors and gradients to update the weights. Both can be sparsified to only update weights partially (see Section~\ref{sec:gradients}). This can have a similar effect to ephemeral  sparsification in the forward pass and lead to significant performance improvements, especially in distributed settings. An option here is to delay the communication/update of small local gradient contributions until they are significant~\cite{2019-renggli}. 
Another important class of ephemeral techniques is conditional computation, where the model dynamically decides a sparse computation path for each example. 
We overview ephemeral sparsification techniques in Section~\ref{sec:dynamic}.

\subsection{When to sparsify?}

\begin{figure}[h!]
	\includegraphics[width=0.9\textwidth]{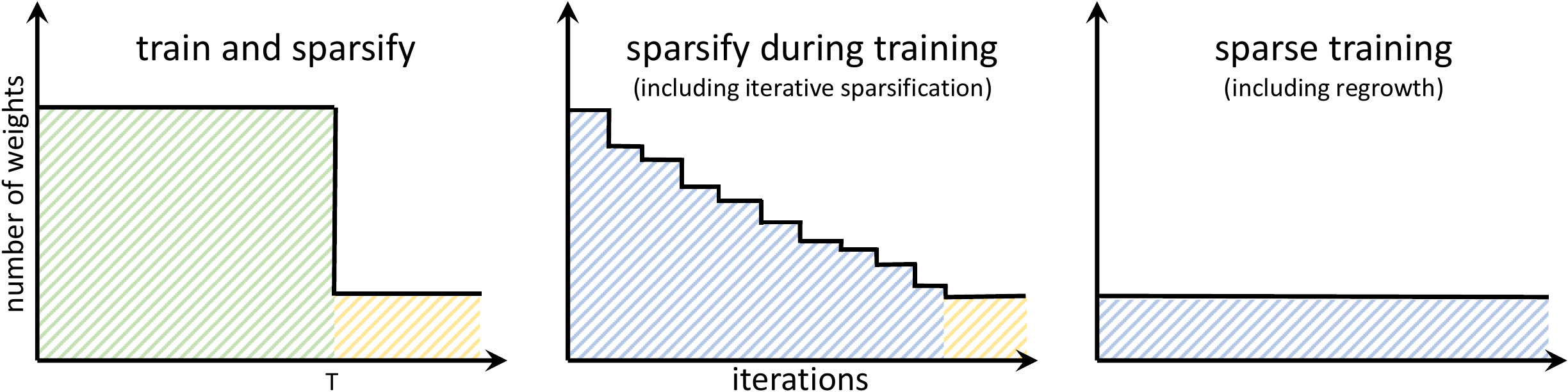}
	\caption{Overview of structural sparsification schedules.}
	\label{fig:training_overview}
\end{figure}

\parad{intro}
While ephemeral sparsity is dynamically updated for each example and configured with a small number of parameters during inference and training, model sparsity follows a more complex NAS-like procedure. Model sparsity is thus often trained with a \emph{schedule}. We differentiate three different classes of training schedules illustrated in Fig.~\ref{fig:training_overview}. Each of those schedules could be used iteratively in an outer train-sparsify loop~\cite{2015-sun}.  

\subsubsection{Sparsify after training}
\parad{train-sparsify}
The \textbf{train-then-sparsify} is the most common schedule type and uses a standard dense training procedure that is run to convergence in $T$ iterations (green area in Fig.~\ref{fig:training_overview}) followed by a sparsification of the fully trained model. Beginning from the earliest works~\cite{1989-janowsky}, the model is typically re-trained (``fine tuned'') after the sparsification to reach significantly higher accuracy (yellow area in Fig.~\ref{fig:training_overview}). 
This schedule type aims at improving performance and/or generalization during inference. It provides the best baseline for model quality because we can always compare the sparsified model quality with the original dense model. Furthermore, since we are starting from a dense model, training does not change such that existing hyperparameter settings and learning schedules can be re-used. Some early works even show that pruning before the model has converged can reduce the final accuracy~\cite{1996-engelbrecht}.

\subsubsection{Sparsify during training}\label{sec:sparsify-during-training}
\parad{sparsify during training}
The \textbf{sparsify-during-training} schedule starts sparsification of the model before it has been trained to convergence and is usually cheaper than the train-then-sparsify schedule. 
Furthermore, training a dense model to convergence may allow for overfitting that is hard to correct with pruning alone. 
Schedules that gradually sparsify during training may follow a pruning schedule that also corrects for approximation errors due to premature pruning in early iterations.
Such schemes often train the dense model for some iterations before sparsification starts and end with a sparse trained model. Early work~\cite{1992-finnoff} advocates a fixed schedule to sparsify during the training before the model converges to improve the quality of solutions using early stopping. In general, sparsifying during training already reaps potential performance benefit of sparsity early on but could lead to less efficient convergence and is often more brittle to configure via hyperparameters~\cite{1995-ghosh}. Furthermore, this approach needs to hold the dense model in memory at the beginning of the operation and thus does not enable the use of smaller-capacity devices.

\parad{full vs. partial gradients}
Some methods take advantage of this limitation and do not reduce the memory consumption during the training process. Instead of deleting pruned weights and gradients, they use binary masks to determine the presence or absence of weights and update even masked weights during backpropagation to enable better weight regrowth/selection (see Section~\ref{sec:dynamic}). 
For example, \citet{2019-wortsman} and \citet{2020-lin} keep the full weights around to implement an efficient search through different sparse architectures by turning weights on and off during training.

\parad{pruning schedule based on generalization error}
The sparsification schedule, i.e., how fast to prune how many elements, is of central importance to this method. \citet{1997-prechelt} observes that a fixed pruning schedule can reduce the generalization ability of the network substantially.
He also observes that the distribution of weight values during training is roughly normal with the mean and variance increasing during the process. Pruning reduces the variance and raises the mean, then during early training the variance increases and the mean decreases before training proceeds as before with increasing mean and variance. 
Prechelt uses the generalization loss to characterize the amount of overfitting and adjust the pruning rate dynamically during training. The pruning rate increases with growing generalization loss and saturates at a maximum value. 
This method demonstrates a significant gain in generalization ability for well-tuned static-dynamic schedules.

\parad{iterative thresholding}
Another approach, Iterative hard thresholding (IHT), is a technique where training schedules of dense and sparse iterations are combined~\cite{2016-jin}. IHT iterates the following two steps: (1) prune all but the top-$k$ weights by magnitude (implements an $L_0$ constraint, see Section~\ref{sec:regularization}) and fine-tune the sparsified network to the task for $s$ iterations, and (2) re-enable the pruned weights and train the dense network for $d$ iterations. The outer loop is running for $i$ iterations with a total of $s_i$ sparsified and $d_i$ dense training steps. The first step regularizes the network while the second step relaxes the optimization to ``learn better representations''~\cite{2016-jin}. \citet{2017-han} use a similar scheme where they run three steps during training: (1) (traditional) dense training to convergence, (2) magnitude-pruning followed by retraining, and (3) dense training. All steps are performed for multiple iterations but the overall scheme is not repeated. They show that this dense-sparse-dense scheme leads to significantly higher generalization performance.  
\citet{2018-carreira} use a similar scheme of sparsification followed by training. They only re-enable a subset of the weights while others are masked out by a learned mask using a penalty term. They argue that magnitude-based pruning (see Section~\ref{sec:sparse-magnitude}) arises naturally in their scheme but the ``soft pruning'' approach selects better weights allowing for higher sparsity. 
All those schemes aim to improve the ``learnability'' of the model by supporting the standard stochastic gradient descent (SGD) algorithm.

\paragraph{SGD training dynamics and sparsity}\label{sec:early_structure_adaptation}
\parad{training dynamics and sparsity}
Similarly to reduced neuroplasticity as biological brains age~\cite{jones2006cognitive}, studies of deep neural networks show that the importance of elements is determined relatively early on in training. 
Specifically, \citet{shwartzziv2017opening} argue that SGD-based training of deep neural networks happens in two phases: (1) a drift phase that quickly minimizes the empirical risk (training error), and (2) a diffusion phase that compresses the internal representation. Similarly, \citet{achille2019critical} describe two phases of training where the first phase discovers the important connections and their topology between layers and the second phase fine-tunes this relatively fixed pattern.
\citet{2019-michel} show that the most important heads in transformers (see Section~\ref{sec:tformers}) are identified in the first 10 epochs.
\citet{2019-ding} observe that identifying weights for later elimination happens early in the training process and weights are rarely re-added late in the process. 
We call this phenomenon \emph{early structure adaptation} in the following. 

\parad{early bird tickets with low-cost training}
\citet{2020-you} and \citet{2019-golub} directly utilize early structure adaptation during the training process where they freeze the sparsity pattern after some iterations. \citet{2020-you} propose to use low-cost approximate training to identify the best sparse structure before starting the actual training of the network. Their work is inspired by \citet{li2020explaining}, who show that a large learning rate in earlier iterations helps the model to memorize easy to fit patterns that are later refined. Specifically, they show that for structured pruning of feature maps in convolutional networks, quick training at low precision and large learning rates leads to a good approximation of the sparse network structure. In general, early structure adaptation is reflected in learning rate schedules and most sparsification schemes use large learning rates for denser models and drastically reduce the rate with growing sparsity.

\subsubsection{Sparse training}\label{sec:sparse_train}
\parad{fully-sparse training}
The \textbf{fully-sparse training} schedule starts with a sparse model and trains in the sparse regime by removing and adding elements during the training process. \citet{2008-narasimha} showed early that this scheme can even outperform separate growing or pruning approaches for neuron-sparse training of simple MLPs. Weight-sparse training often uses complex hyperparameter settings and schedules. However, it enables to train very high-dimensional models whose dense representations would simply not fit into the training devices. 

\parad{static vs. dynamic structure}
We differentiate between \textbf{static} and \textbf{dynamic sparsity} during sparse training. Dynamic sparsity combines pruning and regrowth of elements during the training process, while static sparsity prunes once before the training starts and does not update the model structure during training. 

\paragraph{Dynamic sparsity during training}
We start with schemes that iteratively prune and add (regrow) elements during the training phase. A general overview of pruning techniques is provided in Section~\ref{sec:removal} while growth techniques are described in Section~\ref{sec:growth}. Dynamic sparse training can use any combination of those schemes---we highlight some successful approaches below. 

\parad{flexible pruning schedules with varying sparsity}
The number of elements and the sparsity does not necessarily have to remain constant throughout training. NeST~\cite{2019-dai}, for example, uses a training schedule that is inspired by the development of the human brain~\cite{2017-hawkins}. 
It uses three stages to arrive at the final network architecture: (1) a random seed architecture (``birth brain''), (2) a growth phase (``baby brain'') where neurons and connections are added, and (3) a pruning phase (``adult brain'') where weights and neurons are removed. 
SET~\cite{2019-mostafa} combine magnitude pruning and random regrowth to maintain a balanced parameter budget throughout training. This and many other schemes focus on different ways to regrow connections, those are outlined in Section~\ref{sec:growth}.

\paragraph{Fixed sparsity during training}
Networks can also be trained with a fixed sparsity structure determined before training starts. This structure can either be hand-tuned such as ``structured sparsity'' for transformers~\cite{child2019generating}, sparsity determined in a pre-training phase~\cite{2020-you}, or data-independent (randomly initialized) sparsity~\cite{2017-changpinyo,2017-prabhu,2017-bourely,2020-su}. 

\parad{random initialization}
\citet{2018-liu} question the hypothesis that one must train an overparameterized model and then prune it in order to achieve acceptable accuracy. They show that for neuron and filter removal (structured sparsity), training a smaller model with standard random weights suffices.  
They show examples for CNNs on CIFAR-10 and ImageNet. They achieve state-of-the-art for neuron pruning but fail for weight pruning (unstructured sparsity) on the large ImageNet dataset where fine-tuning still improves performance.
They conclude that one can train sparse models from scratch without pruning if the architecture and hyperparameters are chosen well and that such sparse training may improve performance.

\parad{single-shot pruning before training}
SNIP's~\cite{2019-lee} single shot network pruning approach identifies unstructured sparsity in the network in a data-driven way \emph{before} training. Specifically, the scheme aims to classify (the initial random) weights as important based on an influence to the loss metric proposed nearly 30 years earlier by \citet{1988-mozer}: $I^{(1)}_{w} = \left| \frac{\partial L}{\partial {w}} {w}\right|$, where $I_{w}$ represents the importance of weight ${w}$ evaluated for a single batch. They suggest to choose the batch size equal to the number of result classes. Then, the least important weights are removed and the network is trained in a standard way. 
ESPN~\cite{2020-cho} uses a similar technique but trains the network for a small number of iterations before sparsification in order to quickly establish more structure using early structure adaption in DNN training.

\parad{preserve gradient flow}
\citet{2020-wang} observed that for sparsities above 99\%, SNIP eliminates nearly all weights in some layers, effectively creating a bottleneck. Following this observation, they note the importance of ``gradient flow'', the ability to propagate gradients through the network. They observe that SNIP can hinder gradient flow and performs worse than random pruning at high sparsity~\cite{2020-jorge}, because it considers the gradient for each weight in isolation. \citet{2020-tanaka} even show cases where SNIP disconnected networks, rendering them untrainable, by removing all weights of a layer, a phenomenon they name ``layer collapse''. \citet{2020-wang} detect bottlenecks through a reduction in the norm of the gradient. They propose Gradient Signal Preservation (GraSP), a scheme that considers gradient flows and only prunes weights that decrease the gradient norm (i.e., slow the training of the whole network) least \emph{after} being pruned. GraSP redefines SNIP's gradient-magnitude product of importance to the Hessian-gradient-magnitude product: $I^{(2)}_{w} = \delw \mathbf{H} g_\vec{w}$, with $\delw$ being a selection vector for $w$: $\delw = (0, \ldots, -w, \ldots, 0)$. They also show that GraSP improves upon SNIP in very sparse regimes. A similar observation of a ``minimal layer (junction) density'' to maintain a given accuracy was made earlier by \citet{2018-dey}.

\citet{2020-verdenius} criticize the complexity of GraSP and introduce the small-step iterative SNIP-it for unstructured and SNAP-it for structured pruning, all before training. They follow the intuition that some elements that may be of medium importance initially, gain importance with increased pruning, roughly following the gradient flow argument. By iteratively removing elements according to $I^{(1)}_{w}$ followed by a re-assessment of the importance scores similar to SNIP, information bottlenecks are prevented at a much lower complexity than GraSP. 
\citet{2020-jorge} derive a similar iterative algorithm as well as a variant that slightly improves performance by reanimating weights excluded in earlier iterations. They suggest to use more data during the structure finding phase and 
show a 5x improvement of performance over GraSP while achieving similar quality.
This scheme achieves state-of-the art results today but leads to a lower accuracy than pruning of fully-trained ResNets.
\citet{2020-verdenius} also found that random initialization is a very strong baseline, hinting at the idea of data-free initialization methods, which we discuss next.

\parad{data-free initial pruning with signal propagation}
The authors of SNIP complemented their initial pruning scheme with a \emph{data-free} pruning that only considers the structure of the network~\cite{2019-lee-init}.
They consider the ``signal propagation'' across layers: better signal propagation leads to better properties during training, which leads to better networks (loss minima). Starting from a random pruning, they propose to increase the signal propagation through each layer by adjusting the initial weights using a gradient descent method. This method initializes  weight matrices $\vec{w}$ to full rank such that the combination of sparse topology and the weight is layer-wise orthogonal.
The authors argue and show empirically that such randomly structured but orthogonally initialized networks can be trained to achieve the same or higher accuracy than dense networks with the same number of parameters. \citet{2020-hayou} provide additional theoretical evidence for the efficacy of this initialization scheme and show how ResNets can be effectively initialized. 
\citet{2020-verdenius} and \citet{2020-jorge} also use this scheme for initializing networks pruned in a data-dependent way. 
With such data-free schemes, the pruning ratio still needs to be fine-tuned per layer. \citet{2020-su} propose a fixed sparsity schedule (``smart-ratio'') for ResNet and VGG that decreases for larger layers. Other networks would need to be tuned accordingly. 

\parad{data-free based on synaptic flow}
\citet{2020-tanaka} propose to overcome layer collapse by ensuring a minimal flow through the sparse network. They also show that iterative magnitude pruning avoids layer collapse, providing additional support for Verdenius' and Hayou's iterative schemes. They use the $L_1$ path norm in addition to SNIP's gradient-magnitude product to avoid layer collapse and reach extreme sparsity. It remains unclear whether the performance-accuracy tradeoff at those sparsity levels (for which layer collapse would happen) justifies the cost of avoiding it.

\parad{neural tangent transfer}
Another fixed sparsity training method, Neural Tangent Transfer~\cite{2020-liu}, uses a dense teacher to derive a sparse model without requiring labels that follows a similar training trajectory as the dense one.

\subsubsection{Ephemeral sparsity during training}

Most efficient training methods would take advantage of both ephemeral and model sparsity during training (see Section~\ref{sec:dynamic} for an overview). In an empirical study,~\citet{2020-raihan} observe that training is less robust with respect to sparsifying activations in the forward pass and gradients in the backward pass. Based on those findings, they design the SWAT method that eliminates small weights during the forward pass and both small weights and activations during the backward pass using a simple top-$k$ method.

\subsubsection{Sparsify for transfer learning and fine tuning}

\parad{pruning at transfer learning time}
In transfer learning, large pre-trained and somewhat generic networks are specialized to an often narrower task than the original broad training goal. This specialization is another opportunity for pruning and potentially parameters can be pruned during the process~\cite{2016-molchanov,2019-mehta}. The schedule for such pruning during fine-tuning is similar to the train and prune schedule: a model is trained to convergence and then pruned. However, the difference is in the training dataset and corresponding distribution. The dataset used for fine-tuning is different from the original dataset---often it corresponds to a specific subset, but sometimes it could represent a distributional shift. 
So in some sense, the pre-trained network can be seen as a more intelligent (non-random) weight initialization as basis for a shorter learning process. Also, data sets for fine-tuning are often much smaller. 

Given these characteristics, different pruning mechanisms are used in practice. Specifically, \citet{2016-molchanov} and \citet{2020-sanh} use first order (gradient-based, see Section~\ref{sec:1storder}) pruning for transfer learning to capture the change from the pre-trained weights to the new weights. \citet{2019-mehta} use magnitude-based pruning to transfer sparse networks during fine-tuning.  Those and related methods are summarized in Section~\ref{sec:1storder}.
\citet{2020-chen} showed that task-specific fine-tuning of the BERT transformer network can result in 40-90\% sparsity in final weights using iterative magnitude pruning. They found that most fine-tuned networks have a task-specific structure while the masked language modeling task that was used for pre-training generates universal sparse networks that even transfer well to other tasks. 

\citet{2018-manessi} investigate how well sparse models can be used to transfer their knowledge to other tasks. They show that for various image recognition tasks, moderately sparse models transfer well with either negligible accuracy loss or even a small gain in one example.

\subsubsection{General Sparse Deep Learning Schedules}\label{sec:schedules}
\begin{figure}[h!]
	\includegraphics[width=\textwidth]{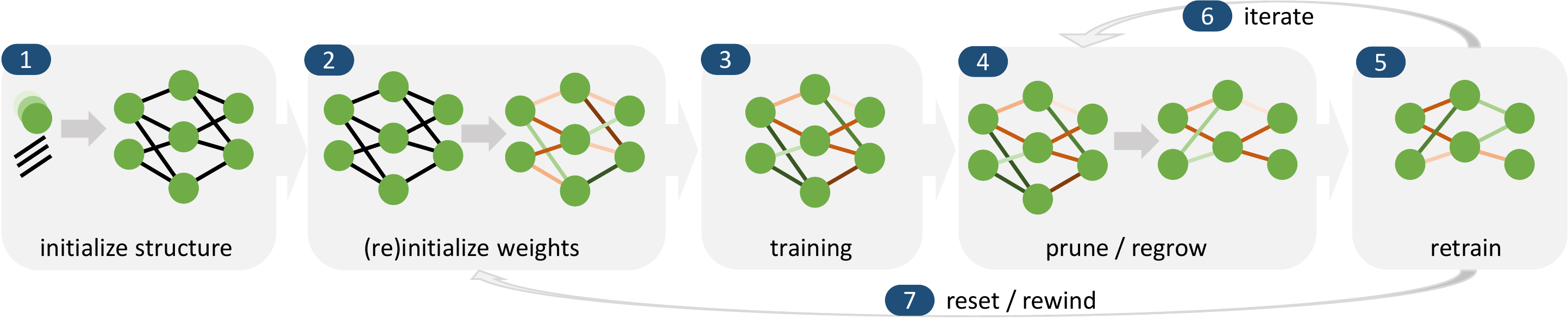}
	\caption{Overview of sparsification schedules. Different weight values are indicated by different colors, the darker the lower the magnitude (black=zero), red indicates positive weights, green indicates negative weights.}
	\label{fig:schedules_overview}
\end{figure}

\parad{general schedule for sparsity in dnns}
Fig.~\ref{fig:schedules_overview} shows a prototypical training algorithm for a pruned network.
%
%
The sparse training process can be described as a series of steps, each can be skipped and some steps can be iterated multiple times. 
Step (1) initializes the network structure, this can either load a description of the network structure from disk or be built using a framework as is usually done for dense networks. 
However, it could also generate a random network structure or use a sparse network construction strategy such as SNIP (see Section~\ref{sec:sparse_train}).

Step (2) initializes the weights of the network, typically randomly or in transfer learning settings with pre-trained weights. For sparse networks, one could use specialized initialization strategies such as synaptic flow (see Section~\ref{sec:sparse_train}). Different weight values are indicated by different colors in Fig.~\ref{fig:schedules_overview}, the darker the lower the magnitude (black=zero), red indicates positive weights, green indicates negative weights. 

Step (3) trains the network for a defined number of iterations or until convergence. This training can be done with an unmodified dense training schedule or with a sparsity-inducing schedule (e.g., regularization, see Section~\ref{sec:regularization}). This initial training may be run until convergence or stop early for iterative methods. 

Step (4) prunes and regrows various elements (see Section~\ref{sec:candidates}) using the different techniques explained in Sections~\ref{sec:removal} and~\ref{sec:growth}, respectively. 

Step (5) may retrain the network either for a fixed number of iterations or to convergence (this step is relatively often skipped but generally improves model accuracy).

Steps (6) and (7) indicate possible loops in the training process. Step (6) is often used in iterative training/sparsification schedules to achieve highest quality. Step (7) could be used to reset weight values, which is sometimes done (see Section~\ref{sec:lottery}).

\paragraph{Why retraining?} 
Even though many pruning schemes pick the least important elements, the degradation of model quality greatly varies (see Section~\ref{sec:architectures}). \citet{1989-janowsky} point out that ``There is no a priori reason why their initial values should remain optimal after the pruning process''. In fact, many works have shown that retraining immediately following each pruning step and fine-tuning after the last pruning step are both crucial for well-performing sparsification schedules. 

In particular, we observe that many methods follow the pruning (or weight masking) step with re-training the resulting sparse network, a process also known as ``fine-tuning.'' 
When the sparsification is performed in multiple steps (usually called gradual or iterative pruning), then several fine-tuning periods may be applied. 

\begin{figure}[h!]
	\centering
		\includegraphics[width=\linewidth]{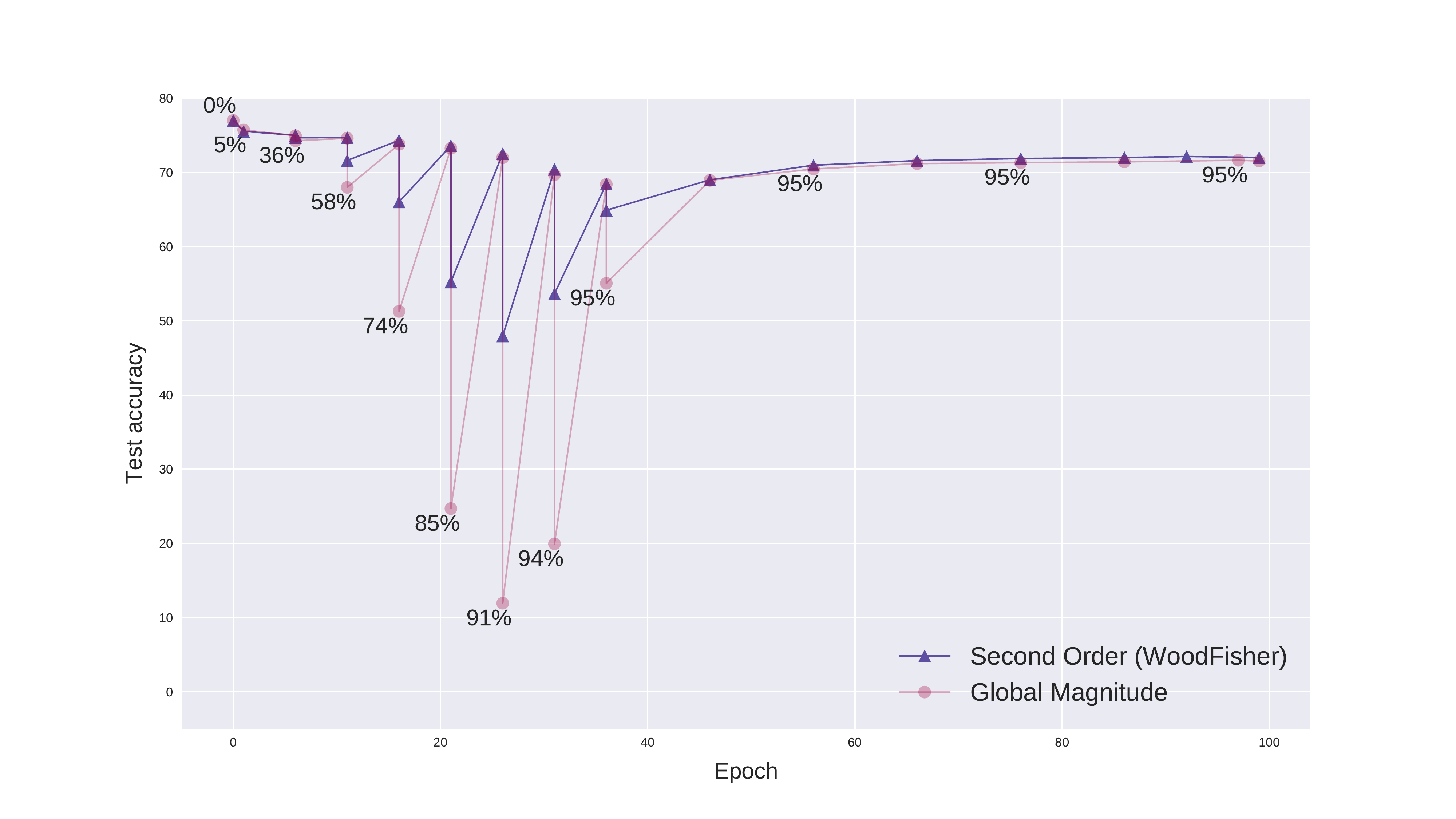}
				\caption{An illustration of a standard gradual pruning schedule including fine-tuning periods, applied to \textsc{ResNet-50} on the ImageNet dataset. The graph depicts the evolution of the validation accuracy for two different methods (global magnitude pruning and WoodFisher~\cite{2020-singh}) across time.}
		\label{fig:gradual-imagenet}
\end{figure}

The approach of choosing which elements to remove based on the difference in loss immediately observed after removal inherently assumes that the accuracy after fine-tuning correlates perfectly with the accuracy \emph{before} fine-tuning, i.e., immediately after pruning was applied. 
This assumption was validated to some extent in the analysis of~\cite{2019-he}, which exhibited a correlation between the two accuracies. 
However, other references, notably~\cite{2020-singh}, observe that the SGD fine-tuning process can serve to ``level'' the performance of various schemes, to the extent that large gains in terms of quality immediately following the pruning step for a specific method can be erased to a large extent after fine-tuning. 
Fig.~\ref{fig:gradual-imagenet} provides an illustration of this phenomenon, as well as of the structure of a gradual pruning schedule. %

Specifically, the sparsity targets shown on the graph are increased progressively, starting at 5\%, until they reach the final 95\% target. 
Fine-tuning periods of fixed length are applied between pruning steps, and a longer fine-tuning period follows the last pruning step. 
Observe the loss of accuracy immediately following the pruning steps, for both methods. Further, notice the significantly better performance of the second-order WoodFisher method immediately following a pruning step, but also the fact that the difference between the methods largely levels off before the next pruning step, due to SGD fine-tuning. 
Ultimately, the second-order method does achieve higher accuracy than the magnitude-based one (by $0.4\%$ Top-1 validation accuracy), but this difference is lower than what one may expect based on the difference immediately following the pruning step.

\parad{intro}
\paragraph{Update frequency of sparse model structures}
All methods described above allow to choose a sparsification frequency through the number of iterations in the (re)training steps. While ephemeral sparsification schemes are applied to each example in each minibatch, structural changes to the model often benefit from delays to reduce noise (cf. momentum) and amortize the often expensive rearrangement of data structures over multiple examples. This is consistent with biological brains where neurotransmitters are activated at high frequency while plastic structural changes happen relatively infrequently (e.g., during sleep~\cite{de2017ultrastructural,diering2017homer1a}).

\parad{hyperparameters}
Tuning the right update frequency for structural changes is crucial to the performance of the final model~\cite{2016-jin}. There have not been many structured studies on how to tune this new hyperparameter but it seems related to the choice of minibatch size and ideas such as gradient noise~\cite{mccandlish2018empirical} may be a good starting point. \citet{2020-raihan} show that a higher update frequency is better for training based on ephemeral weight and activation sparsity. Many works also consider tempering hyperparameters on a specific schedule during training (e.g., sparsification probability~\cite{2016-guo}), similarly to other hyperparameters (e.g., learning rate schedules).

\subsection{Ensembles}

One interesting use-case for sparsification is to enable ensemble models with a limited parameter and compute budget. Instead of having a single model within the budget, one could train an ensemble of multiple smaller models and average or otherwise combine their outputs to make a final selection. \citet{2014-collins} show that 2--3 ensemble models can improve the performance of image recognition tasks over a single model with the same parameter budget. 

\section{Selecting Candidates for Removal}\label{sec:removal}

The core operation in any sparsification scheme is to select candidate elements to be removed. 
The most intuitive and most precise data-driven way to select elements for removal is to evaluate the network with and without the elements in question~\cite{2001-suzuki}. 
However, this simple leave-some-out approach to just train the network with and without the neurons or weights removed poses obvious scalability challenges as it needs to train $n \choose k$ networks with $n$ elements total and $k$ removal candidates. 
Another simple method is to select elements to be removed at random, which is related to the theory of compressive sensing and can be quite effective in some settings~\cite{2017-changpinyo,2018-mittal}.
However, guiding the removal by some metric of importance has been shown to perform best to achieve compressed models with high sparsity in practice. In the following, we provide an overview of such selection methods. 

\begin{figure}[h!]
	\includegraphics[width=\textwidth]{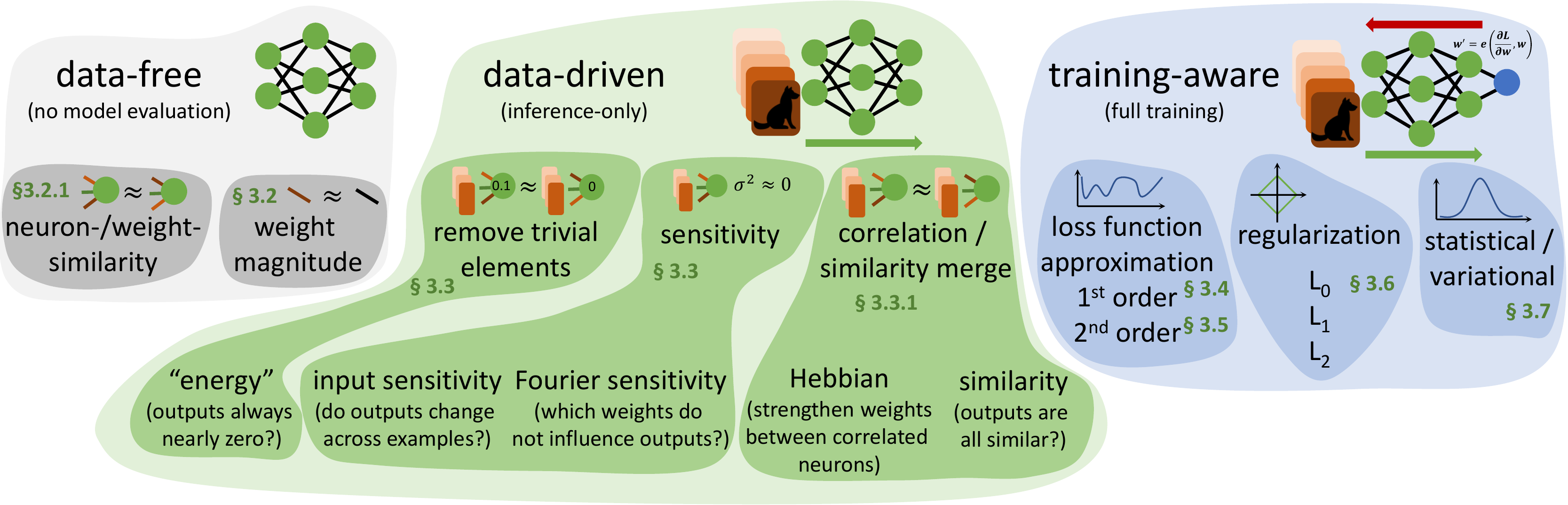}
	\caption{Overview of schemes to select candidate elements for removal during sparsification}
	\label{fig:sparsification_techniques}
\end{figure}

The various schemes for element removal form the basis of different sparsification methods. Unfortunately, comparative studies such as \citet{2019-gale} have not identified a clear winner, thus, we aim to provide a comprehensive overview of the known methods. We will not quantify the efficacy of each scheme here, because this depends on the exact setting of network architecture, hyperparameters, learning rate schedule, learning task etc., and different works can hardly be compared. 
Instead, we will focus on the intuition behind each scheme, and describe specific results in their experimental context for some network architectures in Section~\ref{sec:architectures}. We provide a set of references for each method for more details. 
Fig.~\ref{fig:sparsification_techniques} provides a coarse classification of existing methods to select candidates for removal and a roadmap for this section. 

\subsection{Structured vs. unstructured element removal}

\parad{intro}
As discussed in Section~\ref{sec:perf}, fine-grained unstructured weight sparsity requires storing the offsets of non-zero elements and handling the structure explicitly during processing. Both add significant cost to processing and storing sparse deep neural networks. 
Structured sparsity constrains sparsity patterns in the weights such that they can be described with low-overhead representations such as strides or blocks.  
This reduces the index storage overhead and simplifies processing (see Section~\ref{sec:speed}). Structured sparsity promises highest performance and lowest storage overheads but it may lead to worse models because it may limit the degrees of freedom in the sparsification process. 

One simple example of structured sparsity is the removal of whole neurons in a fully-connected layer: the resulting computations for the forward or backward pass after removing a neuron are simple dense matrix multiplications from which a whole row/column was removed (weights of all incoming and outgoing connections). A similar argument applies to the removal of convolutional filters~\cite{2015-polyak} and transformer heads~\cite{2019-michel}. 
%

\parad{strided sparsity}
Strided sparsity~\cite{2015-anwar} considers structured weight sparsification at the granularity of channels (removing whole feature maps in a layer), kernels (removing all connection between two features in consecutive layers), or a strided kernel structure (remove all connections between features with a particular stride). For example a stride-2 weight vector could be $w=[0.2, 1.9, 0, 1.3, 0, 0.3, 0, 1.2, 0, 0.4]$ where after an initial offset of one, every other element is zero. The storage of this vector would simply require to memoize the offset, stride, and non-zero elements, e.g., $\hat{w} = [1, 2, 0.2, 1.9, 1.3, 0.3, 1.2, 0.4]$.

\parad{im2col sparsity}
Convolutional layers can not only benefit from structured sparsity by dropping whole filters or kernels. If we write the convolution operator in matrix form (sometimes called im2col~\cite{im2col}), we can sparsify groups in those matrices. Here, each input map may have a different non-zero structure which is shared across all output maps. \citet{2015-lebedev} showed that this scheme, together with a regularizing training procedure and magnitude-based pruning, can sparsify filters effectively. They also find that the resulting filters are shrunk towards the center and remain largely circular. \citet{2020-meng} learn filter shapes using $L_1$ regularization.
A similar scheme sparsifies the connections between filters---not all output filters in layer $i$ are connected to all input filters in layer $i+1$. Specifically, \citet{2017-changpinyo} choose fixed random connectivity between the filters at each layer.

Structured pruning often uses similar schemes to unstructured pruning, sometimes with minor modifications to prune whole sets of weights. For each of the following pruning methods, we will outline its extension to structured sparsity if it is not obvious. 

\subsection{Data-free selection based on magnitude}\label{sec:sparse-magnitude}

\parad{intro}
One of the simplest, but also most effective, selection schemes is removing weights with the smallest absolute magnitude. This intuitive approach of removing small weights has been discussed ever since in the early 90's as a simple and effective technique~\cite{1993-hagiwara} and always fares surprisingly well~\cite{1995-thimm,2019-gale}. It is often used together with re-training the sparsified network~\cite{2015-han} and training schedules where the sparsity is gradually increased over time~\cite{2017-zhu}. It can be applied to either individual weights or arbitrary groups of weights using $\sum |W_i|$ for structured pruning (e.g., blocks or rows/columns for whole neuron pruning). As we will see in the next section, this scheme even has a strong theoretical justification, under some assumptions. 

\begin{figure}[h!]
	\centering
	\begin{subfigure}[b]{.32\linewidth}
		\centering
		\includegraphics[width=\linewidth]{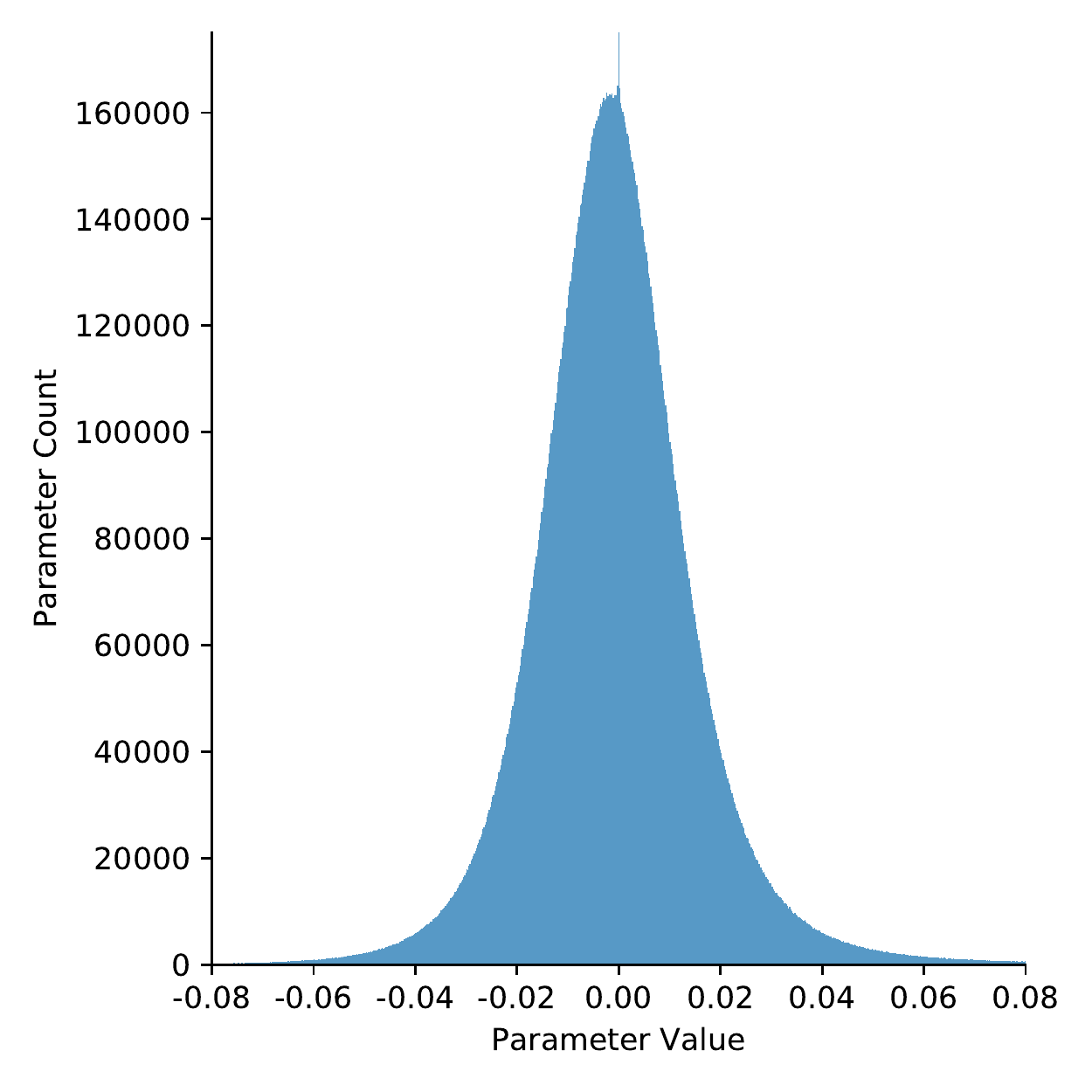}
		\caption{Dense Network (76.0\%)}
	\end{subfigure}
	\begin{subfigure}[b]{.32\linewidth}
		\centering
		\includegraphics[width=\linewidth]{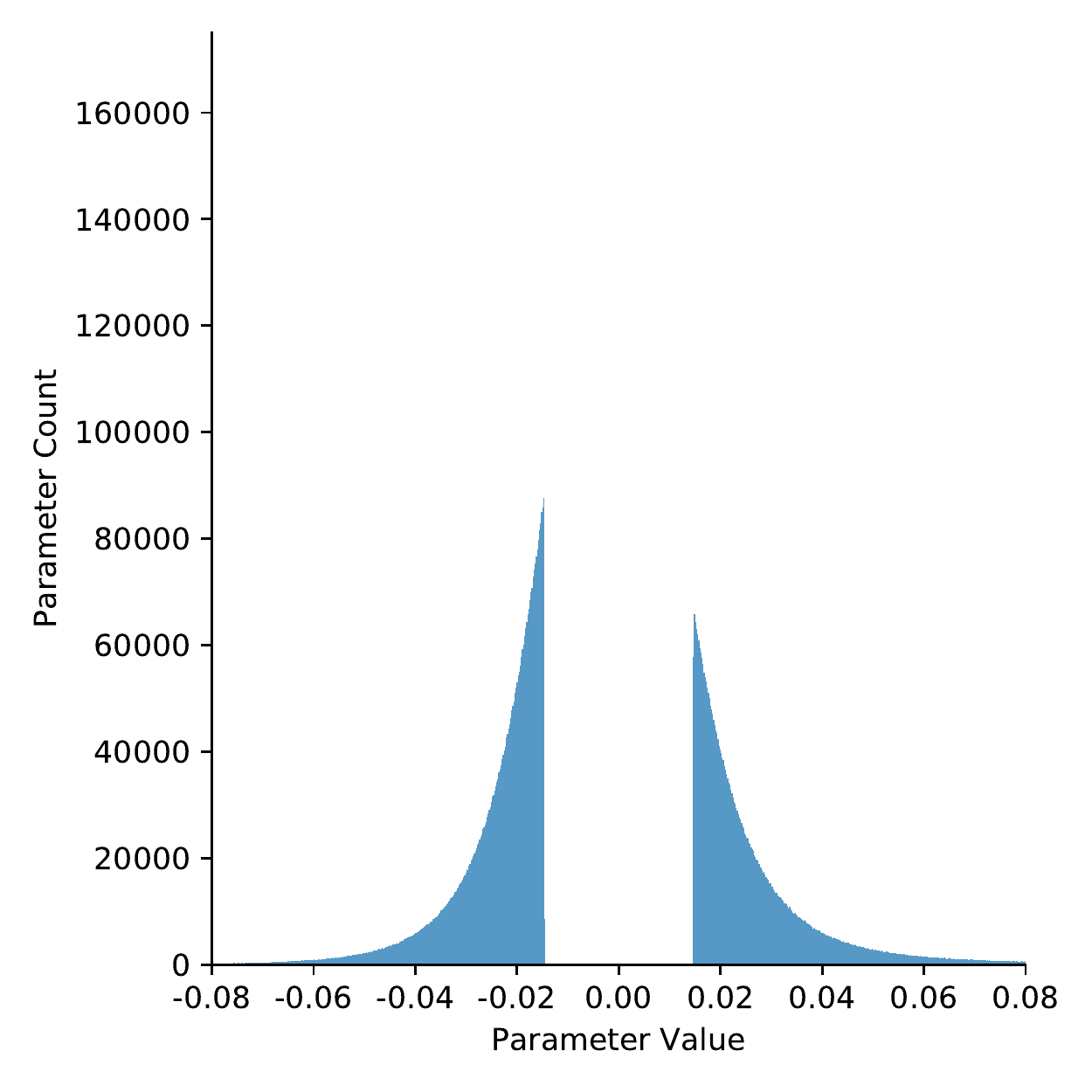}
		\caption{70\% Pruned (36.1\%)}
	\end{subfigure}
	\begin{subfigure}[b]{.32\linewidth}
		\centering
		\includegraphics[width=\linewidth]{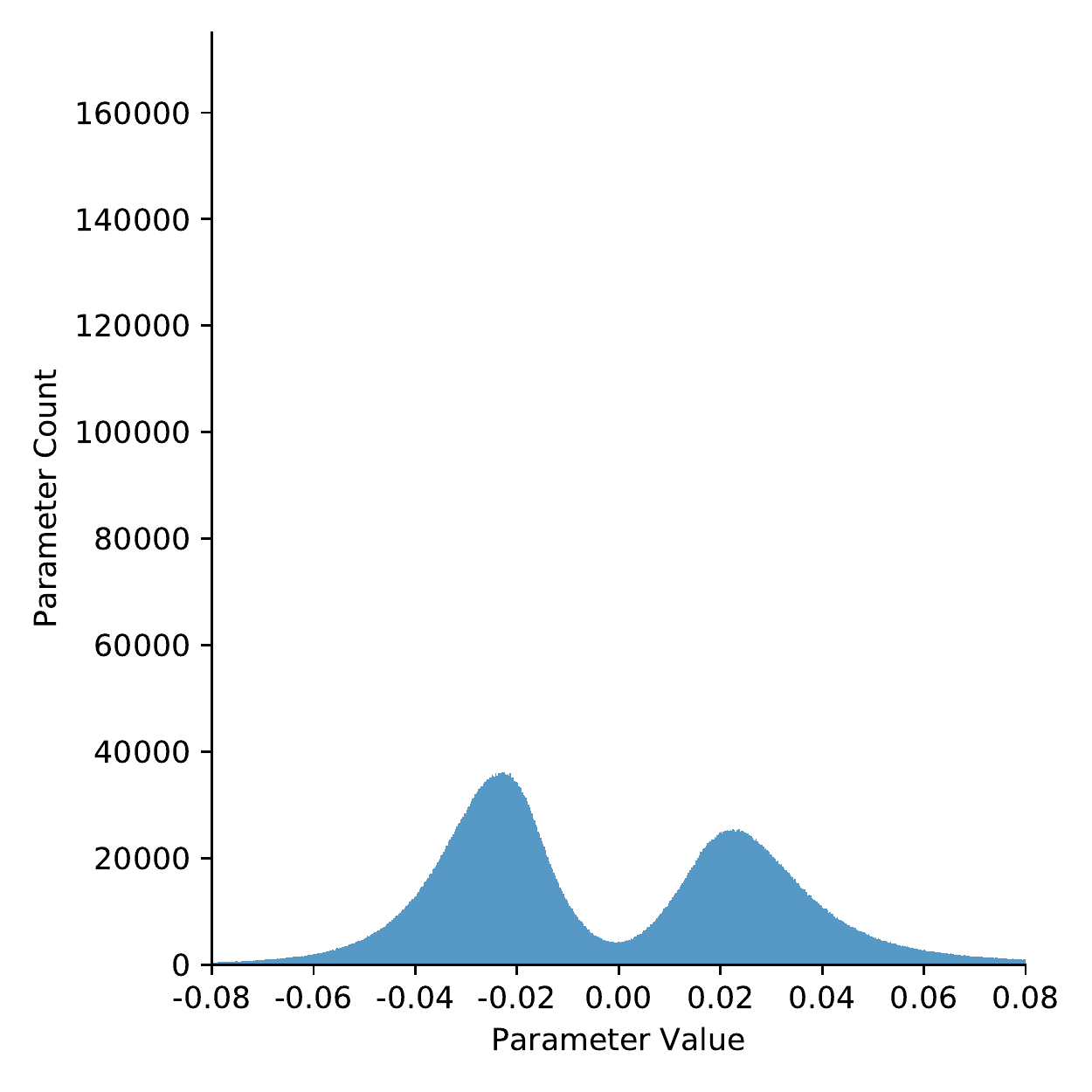}
		\caption{After 3-epoch Retraining (71.4\%)}
	\end{subfigure}
	\caption{Magnitude pruning of weights for ResNet-50 and Top-1 ImageNet validation accuracy.}
	\label{fig:magnitude_pruning}
\end{figure}
As the weight values usually follow a normal distribution with a zero mean, pruning by magnitude can remove the bulk of the weights around zero as shown in Fig.~\ref{fig:magnitude_pruning}. Part (a) shows the weight distribution before pruning, part (b) right after pruning with the condition $|w|\leq x$ for $x=0.17$, and part (c) after retraining. 

An obvious question is how to choose the magnitude $x$ below which to prune. Besides fixing a weight budget and keeping the top-$k$ weights globally or per layer, one could learn sparsification thresholds per layer. 
\citet{2020-kusupati} propose a method to learn those thresholds during the normal SGD step. They replace the original weights $w$ with thresholded weights $w'= sgn(w)\cdot ReLU(|w| - \alpha_l)$, where $\alpha_l$  is a learnable pruning threshold per layer. The loss is computed with respect to $w’$ and layer-wise $\alpha_l$ are learned via SGD. This scheme can easily be extended to structured sparsity as noted above. 
Another approach uses a reinforcement learner to derive the best values for each layer. \citet{2019-he} proposed a DDPG agent~\cite{lillicrap2019continuous} to optimize for different scenarios such as a resource constraint or a target accuracy.

\parad{training}
Magnitude pruning is often used during sparse training schedules to maintain an approximately constant connection density during training~\cite{2018-mocanu, 2019-dettmers, 2016-guo}. 
\citet{2018-bellec} slightly modify the scheme to fix a weight to zero if the SGD optimizer would flip its sign during training. 

\parad{applications}
\citet{2015-han} popularized magnitude pruning for modern deep neural networks as part of neural network compression for inference.
\citet{2017-li} prune whole filters with the smallest sum of absolute weights in convolutional layers. 
Several works~\cite{1997-stroem,2016-see,2017-narang} use magnitude pruning to prune recurrent neural networks as well as sparse training. 
Works related to the lottery ticket hypothesis also use magnitude pruning (see Section~\ref{sec:lottery}).

\subsubsection{Other Data-free methods}\label{sec:data-free}

\parad{intro}
Magnitude pruning is not the only scheme that does not consider training examples. Various other schemes solely base pruning decisions on the structure of the network. Since these methods do not depend on examples, they can be used as a pre- or post-processing step for data-driven methods. 

\parad{neuron weight correlation}
A simple scheme compares sets of weights between different neurons. Specifically, if a fully connected layer has $N$ output neurons, we create an $N\times N$ matrix and compare the input weights between all neurons. Now we can simply merge $k$ similar neurons into a single neuron, multiply all weights by $k$, and add all biases. \citet{2015-srinivas} showed that this method works well for small networks but prunes less for large networks. 
Coreset pruning~\cite{2020-mussay} enables a precise tradeoff between sparsity and approximation error. The authors show improved accuracy at 90\% sparsity for very small example networks. 

\parad{discussion why we need others}
While data-free methods, especially magnitude pruning, are often very effective and can provide state-of-the-art results, several works have shown that more precise methods can achieve significantly better results, especially at high sparsity~\cite{2020-sanh}. Furthermore, data-free schemes often require expensive retraining to recover an accuracy as close to the original performance as possible. An obvious way to improve precision is to consider the influence of the training data (and implicitly its distribution) in the pruning selection. This leads us to the class of data-driven pruning schemes. 

\subsection{Data-driven selection based on input or output sensitivity}
\label{sec:ddsens}

\parad{overview}
This class of selection methods considers the statistical sensitivity of the output of neurons or the whole network with respect to the training data. In those methods, a set of examples (potentially all of the training data) is used to determine directly which elements should be removed to maintain or improve prediction accuracy while sparsifying the network. Elements with very small or zero change with respect to deviation of the input examples contribute less in the entire network since their outputs are approximately constant to the variation in their inputs. Thus, such a sensitivity measure can be employed to define the relevance of an element for the function of a network and low-relevance elements can be removed.

\parad{least output change, using linear systems}
The first scheme follows this intuition and removes neurons that show very little variation in their output across various input examples~\cite{1988-sietsma}. After removing, we add their output to the next neurons' biases. Similarly, if two neurons in a layer always produce the same (or opposite) output for all inputs, we can remove one of those and adjust the other one's outgoing weights without changing the overall function. 
\citet{1997-castellano} generalize this scheme and formulated it in terms of solving a linear system of equations to change the weights after removing a neuron in order to minimize the change of output values across the dataset. They compute new weights for all units that consumed the output of the removed unit to minimize the change in their inputs. They pose the problem using a least-squares metric and optimize it with a conjugate gradient method. Their scheme considers networks where layers can be skipped and it does not require hyperparameter tuning. They also mention the possibility to remove individual weights and later develop a similar scheme to prune input nodes (``features'')~\cite{2000-castellano}.
In a similar scheme, \citet{2000-chandrasekaran} model the outputs of hidden units as linear combinations of outputs of other neurons. A neuron can be pruned if its output is well approximated by a linear combination of other units' outputs. 

Such schemes can also be applied to filters in convolutional networks. 
\citet{2017-luo} phrase the filter pruning problem in terms of its output sensitivity to the following layer. They prune filters that, across the whole minibatch, change the output of a layer least. They define an optimization problem and solve it using a simple greedy strategy.
\citet{2018-yu} define an importance metric that aims to minimize the error in the input to the fully connected classification layers (the ``final response layer'') in CNNs. This captures information flows that span multiple layers. 
\citet{2019-ding-c-sgd} use ``centripetal SGD'' to train the network towards similar filter weights that can later be pruned. 
One could also use a geometric interpretation and find filters that are close to the geometric median of all filters~\cite{2019-he-fpgm}. 

\parad{output sensitivity}
A simple generalization is to consider the sensitivity of neuron outputs (either model or layer) \emph{with respect to elements in earlier layers} (including inputs). \citet{2005-zeng} define a direct measure of the output sensitivity of a neuron with respect to deviations in its inputs. They multiply this sensitivity by the sum of the absolute outgoing weights of the neuron to compute the relevance for pruning. The weights are included because they amplify the sensitivity as input to the next layer.
\citet{1996-engelbrecht} define different measure using a \emph{sensitivity matrix} $S$ that captures the change of a neuron $i$ in a layer $k$ with respect to small perturbations of a neuron $j$ in an earlier layer $l$. They first define the sensitivity with respect to a single example as $S_{ij,lk} = \frac{\partial f_{k,i}}{\partial f_{l,j}}$, where $f_{k,i}$ is the output of neuron $i$ in layer $k$. To consider all training examples, they summarize the matrix using a mean square method into an average sensitivity matrix, which is then used to prune neurons that have low significance with respect to all output neurons. 
\citet{2018-tartaglione} later apply a similar scheme to weights but instead of determining the output sensitivity at the end of training, they use a weight update rule that penalizes a weight's absolute magnitude by output sensitivity during training. These simple sensitivity metrics can be seen as early predecessors of the methods basing on a first order Taylor expansion of the loss function (see Section~\ref{sec:1storder}).

\parad{contribution variance}
A related scheme is \emph{contribution variance} which is based on the observation that some \emph{connections} have very similar outputs across examples in the whole training set~\cite{1995-thimm}. Thus, if a connection (a source neuron multiplied by the weight) has little variance across all training examples, then it can be removed and added to the bias of the target neuron. 
\parad{activation magnitude ``energy''}
\citet{1993-hagiwara,1994-hagiwara} proposes an even simpler scheme to prune neurons based on their ``energy consumption'', basically the value of activations throughout training. They prune ``low-energy'' neurons during training and refine the network with a simple magnitude-based weight pruning.
A similar scheme prunes neurons whose activations are mostly zero for many examples---\citet{2016-hu} define the intuitive ``Average Percentage of Zeros'' (APoZ) pruning criterion. This scheme works well for ReLU activation functions that set negative values to zero. This scheme only distinguishes zero and non-zero values. DropNet~\cite{2020-tan-drop} uses the average magnitude of activations for pruning to achieve higher fidelity.

\parad{weight variation using Fourier amplitude sensitivity}
One could also consider the \emph{variation of model output depending on variation of each weight} in a spectral sense. Here, the relevance of a weight can be assessed by its contribution to the variance of the model output. \citet{2006-lauret} propose to use the ``Fourier Amplitude Sensitivity Test'' (FAST) for determining the relevance of weights. The main idea is to simulate periodic oscillation with frequency $\omega_i$ of each weight $i$ in a fixed interval $[l,u]$. Large Fourier amplitudes at the weight’s frequency $\omega_i$ and its harmonics indicate that the output is sensitive to the weight. Then, perform simulation runs to compute the contribution of each weight variation to the total output through this analysis  and remove neurons whose weights are contributing less than 5\% of the total output variance. The number of necessary simulation runs to disentangle the weights grows linearly with the number of weights. 
\citet{2012-han} combine a similar scheme to prune neurons in a single hidden layer. In order to find the best number of neurons for the model, they prune neurons based on their output variance across samples from the input distribution. They use FFTs to determine the change in output for inputs that vary within the input distribution. They add neurons and improve model capacity if the mean-square training error exceeds a bound. 

One benefit of FAST is that it suffices to have upper and lower bounds on the features to roughly approximate the input distribution---detangling the selection process from the data.
\parad{interval adjoint}
\citet{2020-afghan} also only rely on the the size of the interval that the input values live in. Together with the maximum partial derivative of that input with respect to a specific output, they define a measure of significance for neurons to make pruning decisions. 

\parad{input sensitivity}
Many of the schemes above can be applied to any neuron in any layer. However, some study the ``feature selection problem'' to prune input neurons (``features''). 
Many datasets have inputs with very little information, for example, the four corner pixels in the digit-recognition task for MNIST play a very small role in the actual task output. \citet{1995-engelbrecht} propose a sensitivity analysis to identify input neurons that are of little relevance and can be pruned. For this, they start from a fully-trained network and compute each output's sensitivity with respect to each input $s^{(e)}_{ij}=\frac{\partial o_i}{\partial x_j}$ for each example $e$. They then use either a mean square, sum of absolute values, or maximum to summarize the sensitivity of an input value for the whole dataset. They then prune based on the resulting metric, re-train, and optionally repeat the procedure. 

\subsubsection{Selection based on activity and correlation}

\parad{intro}
One simple observation is that, in many networks, some neurons are often activated together, relating to the Hebbian observation ``neurons that fire together wire together''~\cite{hebb-organization-of-behavior-1949}. Several sparsification schemes are based on this observation.  
A simple sparsification scheme could merge neurons that have very similar output activations and simply adapt their biases and rewire the network accordingly. A similar idea has been used in ``data free'' schemes described in Section~\ref{sec:data-free}.

\parad{merge similar neurons based on their activation similarity}
In a method that could be seen as a generalization of APoZ (yet, it was developed earlier), \citet{1988-sietsma,1991-sietsma} observe that some neurons are producing very similar outputs for all examples during inference. They identify such pairs of similar-output neurons across the training examples and remove redundant ones. \citet{1991-kameyama} extend the idea by fusing those neurons and accumulate their weights and biases to minimally affect the sparsified networks to reduce the re-training time.
\citet{2018-suau} perform principal component analysis of max-pooled filter and neuron outputs to select the number of filters for a layer. They use either Principal Component Analysis or KL divergence to compute the number for each layer and then remove the most correlate neurons or filters.

\parad{preferential weight drop between uncorrelated neurons}
A different method would strengthen connections between correlated neurons: we could preferentially drop weights between weakly correlated neurons and maintain connections between strongly correlated neurons. \citet{2015-sun} found that this method works particularly well to refine fully trained networks and leads to better generalization and good sparsification for pruning a convolutional network for face recognition.


While data-driven sensitivity-based schemes consider the outputs across the examples drawn from the input distribution, they purely aim at minimizing the impact on the input-output behavior of the network. Thus, if the network has a low accuracy, it will not gain from such pruning methods. We could now consider the training loss function itself in the pruning process and use it to improve the model accuracy of the pruned network as much as possible.

\subsection{Selection based on 1st order Taylor expansion of the training loss function}\label{sec:1storder}
\parad{intro}
Gradient-based first order methods are most successful for learning weights in deep neural networks. It is thus not far-fetched to also apply similar methods to the selection of weights. Since gradients of the weights are computed during the normal optimization process, one can easily re-use those for determining weight importance. Furthermore, gradient computations are generally cheap, so one could employ them together with additional, so called \emph{gating} elements to select arbitrary elements (weights, neurons, filters, etc.) for removal.

\parad{explain Taylor of L}
If we consider the loss function $L(\vec{w})$ at any time during the training process, we can write a small perturbation at $\vec{w}$ as
$$\delta L = L(\w + \delw) - L(\w)  \approx   {\gradw} \delta \w + \frac{1}{2} \delta \w^{\top} \hess \; \delta  \w,$$ 
where ${\gradw} \delta \w $ and $\frac{1}{2} \delta \w^{\top} \hess \; \delta  \w$ are the first and second order Taylor expansion of $L$, respectively. (It is usual to assume that the influence of higher order terms is negligible and thus they are ignored.) In this and the next section, we describe how to use those terms to view pruning as part of the model optimization process. 

\parad{least weight change}
A first and probably simplest approach to prune weights is to consider the \emph{total weight change} during training. Here, we store the sum of all updates during the training and prune the weights that have changed least~\cite{1990-karnin,2019-golub}. \citet{2019-molchanov} use a squared gradient-weight product as first-order approximation to a neuron's or filter's importance. The intuition is that if weights are changed little from their initial random values during the network's learning process, then they may not be too important. This method would be identical to sparsification techniques based on absolute magnitude (see Section~\ref{sec:sparse-magnitude}) if we consider the change with respect to a (contrived) starting state of all-zero weights. 

\parad{element gating}
One generic way to decide whether elements can be removed is to use a gradient based scheme with respect to a binary \emph{gating} function that regulates whether to include that element or not. Then, during training, differentiate that function at the positions $1\rightarrow 1-\delta$ to determine its importance. 
\citet{1988-mozer} uses this technique to ``trim fat'' neurons from networks in order to improve generalization.  
They define the gradient of a function $\alpha_i$ that disables (``gates'') a neuron $i$ in a fully-trained network as measure of its relevance. 
The transfer function of a fully-connected layer $l$ changes to $f_l = \sigma_R(W_l\cdot \vec{\alpha} \odot f_{l-1})$ where $\vec{\alpha}$ is a vector with the same size as $f_{l-1}$ and $\odot$ stands for the element-wise Hadamard product. 
This method requires two backprop stages---one for the weights and another one for the gate perturbation $\frac{\partial L}{\partial\alpha_i}$. The method can now prune the least important neurons iteratively and stops when it observes a large jump in $\frac{\partial L}{\partial\alpha_i}$. 
\citet{2019-lee} and \citet{2019-xiao} apply a very similar method based on the absolute value of the gradients to gate weights in the model.

\parad{tri-state relu}
The Tri-state ReLU~\cite{2015-srinivas-relu} unit is a generalization of element gating and can be used to learn neuron pruning. It is defined as:
$$
\mathrm{tsReLU}(x)=
\begin{cases}
wx, ~\hspace{1em}x\geq 0\\
wdx, ~\hspace{6px} \mathrm{otherwise.}
\end{cases}
$$
Both $w$ and $d$ are learnable binary parameters; $w$ is similar to the gating function above and $d=1$ turns the nonlinearity into the identity function. If we use a single $d$ for each layer, then we can remove the whole layer for $d=1$. We note that for $d=0$ and $w=1$ the Tri-state ReLU is identical to the traditional ReLU. Learning binary parameters is as tricky as described above and Srinivas and Babu choose the simple function $w(1-w)$ as regularizer with final rounding and constrain the values of $d$ and $w$ to the interval $[0,1]$. This can be interpreted as learning the parameters of a binomial distribution, where each Bernoulli trial indicates whether the weight is chosen or not. 
More general schemes for learning discrete parameters are described in Section~\ref{sec:approx_discrete}.
\citet{2017-srinivas} use the maximum likelihood (simple rounding as before, see Section~\ref{sec:ddsens}) of this formulation to gate weights during training. 
\citet{2019-you} use a 1st order approximation of the loss function (the gradient-weight product, see~\cite{2017-molchanov}) to select filters to prune structurally.

\parad{Jacobian rank deficiency}
One could also investigate the \emph{Jacobian matrix} after training has progressed for some iterations. \citet{1999-zhou} and \citet{2006-xu} found that the Jacobian is usually not full rank, which means that the gradients for some weights are correlated. Zhou and Si use QR factorization of the Jacobian matrix to determine which weights are redundant while Xu and Ho use QR factorization on the output of hidden nodes to determine redundant neurons. Both approaches benefit from the nonlinearity (e.g., sigmoid or ReLU) creating the rank deficiency due to saturation or cut-off. 

\parad{transfer learning}
Specifically pruning during \emph{transfer learning} can benefit from first order gradient information. \citet{2016-molchanov} use the magnitude of the gradients to prune full feature maps to improve the inference efficiency of fine-tuned CNNs. They use the absolute value of the gradient to determine whether a parameter should be removed or not. It seems intuitive to consider the change of parameters during fine-tuning. Movement pruning~\cite{2020-sanh} recognizes that the direction of the gradient plays a crucial role: if the pre-trained weights move towards zero for fine-tuning examples, then they are more likely to be less important (prunable) than if they move away from zero. Their technique accumulates the parameter movement and uses this as task-specific information for pruning. 

\parad{global sparse momentum}
\citet{2019-ding} propose global sparse momentum to change the gradient flow during backpropagation. They classify the weights into two sets based on their importance during training. The important set is updated with the gradients during backprop while the other set does not receive gradient updates but follows weight decay to be gradually zeroed out. The importance of parameters is determined by the magnitude of the gradients and the weights as $S_w=\left|\frac{\partial L}{\partial  w} w\right|=|g_w w|$ (similar to sensitivity-based approaches). The selection of the two sets is performed at each iteration such that weights may move from the unimportant into the important set during training. While the authors point out that this ``re-selection'' is important for the overall accuracy of the model, they also observe that it happens rarely and decreases during the training process following the early structure adaptation observation (see Section~\ref{sec:early_structure_adaptation}).

\subsection{Selection based on 2nd order Taylor expansion of the training loss function}

The question of selecting the ``least significant'' set of weights to remove from a fully-trained model relative to the difference in loss with respect to the current model was considered in the work of \citet{1990-lecun}, followed by \citet{1992-hassibi}. 
These references consider an ``optimization''  approach to pruning, trying to answer the question of which parameter to remove in order to minimize the corresponding loss increase, under the assumption that the  second-order Taylor approximation of the loss around the dense model is exact. Their frameworks differ in terms of assumptions, with the latter work being more general. We will present them jointly, outlining the differences at the end. 

\subsubsection{Pruning as an optimization task}
Let us again consider the Taylor expansion of the loss function at $\vec{w}$ 
$$\delta L = L(\w + \delw) - L(\w)  \approx   {\gradw} \delta \w + \frac{1}{2} \delta \w^{\top} \hess \; \delta  \w,$$ 
where the model perturbation $\delta  \w$ is chosen so that it zeroes out a single weight $\w_i$ in position $i$ and leaves the other ones unchanged, i.e., $\delta  \w = (0, \ldots, - \w_i, \ldots, 0) $.
Since we are assuming that the model $\w$ is trained to a local minimum, the (zero) gradient term can be ignored, and the problem reduces to finding the weight $\w_i$ whose pruning perturbation $\delta \w_i$ minimizes the expression 
\[
    \frac{1}{2} \delta \w_i^{\top} \hess \; \delta  \w_i.
\]
This minimization problem can be solved exactly via the method of Lagrange multipliers, to yield the following ``saliency measure'', which is associated to each weight $\w_i$ 

\begin{equation}\label{eq:prun_stat}
{\rho_i  = \frac{\w_i^2}{2 \, \lbrack\hessinv\rbrack_{ii}},}
\end{equation}

\noindent where $\lbrack\hessinv\rbrack_{ii}$ denotes the $i$th diagonal element of the \emph{inverse Hessian matrix} of the loss $L$ of the given model $\w$. 
To choose which weights to prune, one can sort the weights in decreasing order of this pruning statistic, the lowest-value weight being the best candidate for removal. 

Interestingly, this procedure suggests that the value of the remaining weights should also change, and provides the corresponding optimal perturbation ${\delw}^\ast$.  
This is as follows:
\begin{equation}\label{eq:opt_delw}
\delw^\ast  = - \frac{w_i \hessinv \e_i}{ \lbrack\hessinv\rbrack_{ii}}.
\end{equation}
 
The work of~\citet{1990-lecun, 1992-hassibi} provided the first derivations for this metric, and numerical methods for computing this metric on tiny networks, with tens or hundreds of parameters.

Optimal Cell Damage (OCD)~\cite{1996-cibas} applies a very similar technique to prune the \emph{input features} to the network. The scheme uses the sum of the saliencies $\rho_i$ of all outgoing weights of an input value to compute the saliency of that input. The authors find it to perform worse than approaches based on regularization (see Section~\ref{sec:regularization}).

\subsubsection{Magnitude pruning as a special case}
To gain some intuition, let us consider the above pruning statistic when the Hessian is the identity, possibly rescaled by a constant. 
Intuitively, this would mean that the Hessian matrix is diagonally-dominant, and that its diagonal entries are roughly uniformly distributed. 
In this case, a quick examination of the above equations will yield that following the statistic is equivalent to pruning the weight of \emph{lowest magnitude}, as the saliency measure becomes proportional to the square of each weight.  
As noted, the weight magnitude is a  popular pruning criterion in practice, e.g.,~\cite{2019-gale, 2020-singh, 2020-blalock}. 
We do note that this structural assumption on the Hessian is somewhat strong, and may not hold in practice.

\subsubsection{Discussion of assumptions and guarantees}
The OBD/OBS method offers a powerful mathematical framework for pruning. 
However, the framework comes with a few important assumptions and limitations that are worth noting:

\begin{enumerate}
	\item The original framework assumes that pruning is performed upon a well-trained model, whose loss gradient ${\gradw}$ is negligible. Follow-up work~\cite{2020-singh} has shown that the formulation can be extended to the case where the gradient is non-zero. 
	 
	\item The framework inherently assumes that (a) the Hessian matrix is \emph{invertible} at the point where pruning is performed, and that (b) the pruning perturbation is \emph{small}, and in particular that the Hessian matrix is \emph{constant} along the direction of the pruning perturbation (this is necessary in order to ignore the higher-order terms).  This constraint is addressed by practical schemes by either performing \emph{gradual pruning} of the weights, or by re-computing the Hessian along the pruning direction, as we will detail in the next section. 
	
	\item Importantly, the above derivation holds if we are willing to remove a single weight at a time, and to re-compute the Hessian upon each new removal. Clearly, this would be infeasible for modern networks, so, to apply this method at scale and remove several weights in a step, one assumes that the correlations between removed weights are negligible.\footnote{Technically, it could be that the removal of the lowest weight in the order of the pruning statistic would cause the second-lowest weight to become significantly more important towards the loss.} 
	
	\item Finally, we note that the early work of~\citet{1990-lecun} introduced the above formulation under the assumption that the Hessian matrix is \emph{diagonal}, and applied this method on small-scale networks. \citet{1992-hassibi} generalized this diagonal approximation, and presented efficient numerical methods for estimating the inverse Hessian under additional assumptions, which we will detail in the next section. 

\end{enumerate}

\subsubsection{A Simple Illustration}

\begin{figure}[h!]
	\includegraphics[width=0.5\textwidth]{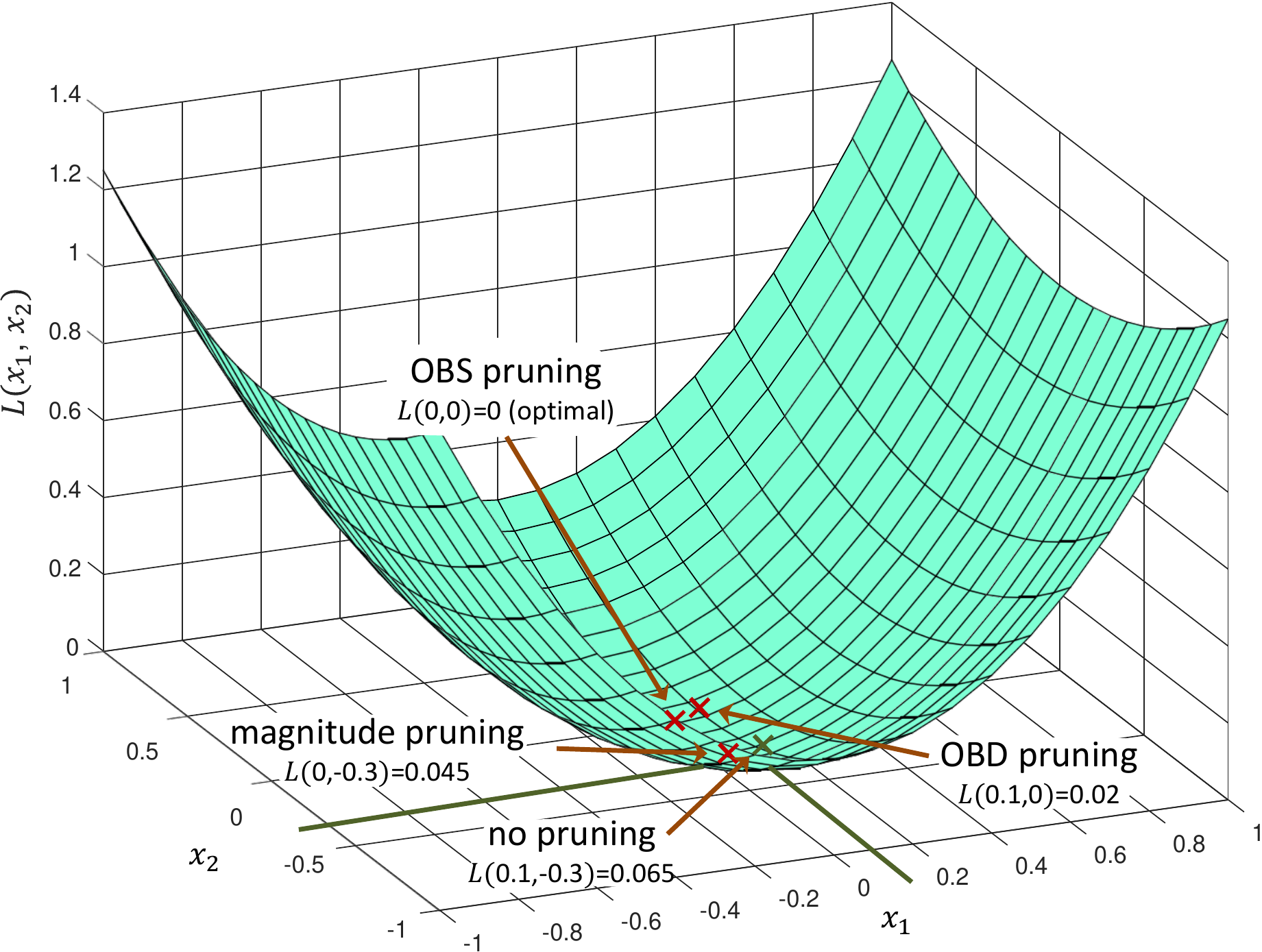}
	\caption{Example function $L(x_1,x_2)=2x_1^2+0.5x_2^2$ with estimated minimum at point $(0.1, -0.3)$.}
	\label{fig:2nd_order}
\end{figure}

We now provide an intuitive example for the workings of the different methods based on the Taylor expansion of the loss function.  Fig.~\ref{fig:2nd_order} shows the function $L(x_1,x_2)=2x_1^2+0.5x_2^2$. Let us assume that SGD found an approximation of the minimum at the point $(x_1^*, x_2^*)=(0.1, -0.3)$. (Clearly, in this example, the optimum is $(0,0)$ but we use $(0.1, -0.3)$ for illustration.) The gradient of $L$ at this point is 0.1 but it is common for second-order methods to assume that it is negligible since the model is well-optimized.

The function value is $L(0.1, -0.3)=0.065$.
Magnitude pruning would evaluate the absolute values for $x_1^*$ and $x_2^*$ and decide to prune $x_1$, getting us to a pruned function value $L^{MAG}(0, -0.3)=0.045$. 
OBD assumes that the Hessian is diagonal (which holds here but may not in general) and would dampen the absolute values of the weights by their inverse Hessian diagonals (i.e., $x_1$ is doubled and $x_2$ is halved), and would decide to remove $x_2$, achieving a better function value $L^{OBD}(0.1, 0)=0.02$. 
Relative to OBD, OBS has two main differences. 
First, OBS does not assume that the Hessian is diagonal, which is more general. In our case, this would lead to the same saliency values, so $x_2$ would be removed.
Second, OBS would also update $x_1$'s value to adjust for the fact that $x_2$ is now set to zero. 
Concretely, we can follow Equation~\eqref{eq:opt_delw} to obtain that $x_1$ should be updated by $\delta x_1 = - 0.1 \cdot \frac{0.5}{0.5} = - 0.1$. 
Thus, the updated sparse point given by OBS is $(0, 0)$, leading to $L^{OBS}(0, 0) = 0$, which in this simple case is optimal.

\subsubsection{Large-scale pruning based on second-order information}

The key question addressed by subsequent work on applying second-order methods to pruning has been how to apply such methods at the scale of deep neural networks, where the dimension parameter is in the millions or even in the billions. 
Calculating the  pruning metric above requires estimating the diagonal of the \emph{Hessian inverse}, which faces several hurdles, as the Hessian is hard to store, let alone invert, and may technically not even be invertible. 

\paragraph{Layerwise Optimal Brain Surgeon (L-OBS)} 
One extension of this classical approach to deeper networks, called L-OBS, was proposed by \citet{2017-dong}, by defining separate layer-wise objectives, and by approximating the Hessian matrix at the level of carefully-crafted blocks, which follow the neuron structure of the network. 
The paper showed superior results relative to the layer-wise magnitude pruning baseline.  

\paragraph{The Empirical Fisher Approximation to the Hessian.} 
A common approach, first proposed by~\citet{1992-hassibi} has been to leverage the \emph{empirical Fisher} approximation to the Hessian matrix. This approximation should hold under the following assumptions: (1) the task being considered is a classification task, e.g., whose output is given via a SoftMax function; (2) the model whose Hessian we wish to estimate is already well-optimized, and in particular its output distribution approximates well the true output distribution. 
Then, following our discussion of the empirical Fisher, one can approximate the Hessian matrix via

\begin{equation*}
\begin{aligned}
H {\simeq} \frac{1}{N} \sum_{j=1}^{N} \nabla \ell_j \cdot \nabla \ell_j^{\top},
\end{aligned}
\end{equation*}

\noindent where $N$ is the number of samples used for the approximation, $\nabla \ell_j$ is the gradient of the loss at sample $j$, and $\cdot$ denotes the outer product. (Recall that, for this approximation to hold, the model's output distribution should match well with the true output distribution.) 

\paragraph{Fisher pruning} \citet{2018-theis} provide an example application of this approximation. 
Specifically, they assume a diagonal approximation of the empirical Fisher matrix, i.e., only compute the diagonal elements, and invert the resulting diagonal matrix. 
They apply this technique to perform structured pruning of gaze prediction models.

\paragraph{Approaches based on low-rank inversion} One can then leverage the observation that this approximation is effectively a sum of rank one matrices to estimate its inverse, via the classic Sherman-Morrison formula. We obtain the following recurrence, which integrates the series of gradients $(\nabla \ell_j )_{j = 1}^N$ taken over individual samples into an approximation to the Fisher matrix:

\begin{equation}\label{eq:sherman-morrison-fisher}
\widehat{H}_{j+1}^{-1}=\widehat{H}_{j}^{-1}-\frac{\widehat{H}_{j}^{-1} \nabla \ell_{j+1} \nabla \ell_{j+1}^{\top} \widehat{H}_{j}^{-1}}{N+\nabla \ell_{j+1}^{\top} \widehat{H}_{j}^{-1} \nabla \ell_{j+1}}, 
\end{equation}
\noindent where initially $\widehat{H}_0^{-1} = \lambda I_{d}$, and $\lambda$ is a small dampening parameter, usually assumed to be small. 
This approach was initially proposed by~\citet{1992-hassibi}, and then re-discovered by \citet{1998-amari} in the different context of optimization via natural gradient. Both these references apply the method at small-scale, specifically on single-layer neural networks.

Recently,~\citet{2020-singh} revisited this method at the large scale of modern deep neural networks. Specifically, they proposed a block-diagonal approximation of the above approach, and showed that it leads to an accurate local prediction of the loss along the direction of pruning, relative to the magnitude, diagonal Fisher, and to the K-FAC approximations.
They then apply this method to both one-shot and gradual pruning, leading to state-of-the-art accuracy for unstructured pruned models in both cases. 
Specifically, they show that the accuracy drop at a single pruning step, when computed using their method, can be significantly lower than using other methods, which leads to higher accuracy following fine-tuning steps.
They also show that results can be further improved by taking the first-order (gradient) term into account, and by re-estimating the Hessian along the direction of pruning.

\paragraph{Extensions of OBD/OBS} 
Several non-trivial extensions of the OBD/OBS framework were presented in the early 90s. 
\citet{NIPS1995_3473decc}, for example, propose the following host of improvements.
First, they extend the method so that pruning is performed with respect to an estimate of the generalization error, rather than the loss. 
For this, they use a framework for the estimation of the generalization error given by \citet{1991-moody}\footnote{A similar approach, but using a different estimator, is given by \citet{713928}.}.  
Second, they incorporate the weight decay term into the OBS metric, following earlier work by~\citet{1994-hansen}. 
Third, they recognize and address the problem of ``nuisance parameters,'' described in brief as the issue that, if eliminating an output weight $w_o$, all the weights in the corresponding hidden unit are practically pruned as well. Thus, their method eliminates these parameters from the model as well, to avoid spurious contributions from them.

\paragraph{Other uses of the Fisher matrix} 
The relatively simple structure of the empirical Fisher matrix inspired additional approaches. For example, 
 \citet{1993-tamura} and \citet{1998-fletcher} use singular value decomposition of the Fisher matrix to determine the ideal number of neurons in each hidden layer. Assuming that outputs are linearly activated, they use the rank of the resulting covariance matrix of maximum likelihood to compute the number of neurons in the compressed network.

\paragraph{Kronecker-Factored Approximate Curvature (K-FAC)} An alternative approximation for the Fisher matrix (and thus, for the Hessian) is a family of methods based on the \emph{Kronecker-Factored Approximate Curvature {(K-FAC)}} \citep{2015-martens}. The method has been originally developed for the purposes of optimization, i.e., to determine an efficient pre-conditioner for the gradient update. 

Following~\citet{2020-singh}, we illustrate the method through a simple example. Consider a fully-connected network with $\ell$ layers.  Let us denote the pre-activations of layer $i$ by $\vs_i$. Then, they can be written as $\vs_i = W_i \va_{i-1}$, where $W_i$ is the weight matrix at the $i^\text{th}$ layer and $a_{i-1}$ denotes the activations from the previous layer, which represent the input of the $i^\text{th}$ layer. 

Following the chain rule, the gradient of the objective function $L$ with respect to the weights in layer $i$ is  
$$\nabla_{W_{i}} L = \text{vec} (\,\vg_i \va_{i-1}^\top\,).$$ 

Above, we denote by $\vg_i$ the gradient of the objective with respect to the pre-activations $s_i$ of this layer, which implies that $\vg_i = \nabla_{s_{i}} L$. 
Using the fact that $\text{vec}(\vu \vv^\top) = \vv \otimes \vu$, where $\otimes$ denotes the Kronecker product, we can simplify our expression of the gradient with respect to $W_i$ as $$\nabla_{W_{i}} L = \va_{i-1}^\top \otimes \vg_i .$$

Given the above, observe that we can now write the  block of the Fisher matrix which corresponds to layers $i$ and $j$ as follows: 

\begin{equation}
\begin{aligned}
F_{i, j}=\mathrm{E}\left[\nabla_{W_{i}} L  \, \nabla_{W_{j}} L^{\top}\right]=\mathrm{E}\left[\left(\va_{i-1} \otimes \vg_{i}\right)\left(\va_{j-1} \otimes \vg_{j}\right)^{\top}\right] &\stackrel{(a)}{=}\mathrm{E}\left[\left(\va_{i-1} \otimes \vg_{i}\right)\left(\va_{j-1}^{\top} \otimes \vg_{j}^{\top}\right)\right] \\
&\stackrel{(b)}{=}\mathrm{E}\left[\va_{i-1} \va_{j-1}^{\top} \otimes \vg_{i} \vg_{j}^{\top}\right], 
\end{aligned}
\end{equation}

\noindent where, in steps (a) and (b) we have used the transpose and mixed-product properties of Kronecker product. The expectation is taken over the model's distribution, as in the formulation of Fisher. 

Finally, the Kronecker-Factored Approximate Curvature (K-FAC) approximation for $\widetilde{F}$  can be written as 
\begin{equation}
\widetilde{F}_{i, j} = \mathrm{E}\left[\va_{i-1} \va_{j-1}^\top \right] \otimes \mathrm{E}\left[\vg_i \vg_{j}^\top\right] = \widetilde{A}_{i-1, j-1} \otimes \widetilde{G}_{i, j}.
\end{equation}

Essentially, we have moved the expectation inside the expression, and applied it prior to performing the Kronecker product. This is a significant analytical assumption, since in general the expectation of the Kronecker product would  not be equal to the Kronecker product of the expectations of its terms. 

The advantage of this approximation is that it allows one to compute the inverse of K-FAC approximated Fisher efficiently. This is because the inverse of a Kronecker product is equal to the Kronecker product of the inverses. This implies that instead of inverting one matrix of size $n_{i-1} n_{i} \times n_{j-1} n_{j}$, one only needs to invert two smaller matrices $\widetilde{A}_{i, j}$ and $\widetilde{G}_{i, j}$, of sizes $n_{i-1} \times n_{j-1}$ and $n_{i} \times n_{j}$, respectively, where we denote the number of neurons in layer $\ell$ by $n_\ell$.

One potential issue with this approach is that it is especially-crafted for fully-connected layers. 
If we wish to apply it to the case of convolutional or recurrent neural networks, the Kronecker structure needs to be further manipulated to yield an efficient approximation, as shown in \cite{2015-martens,2016-ba}.

The K-FAC approximation has found several applications in optimization \cite{2016-ba, 2019-osawa} and reinforcement learning \cite{2017-wu}. 
Specifically in the case of pruning, \citet{2019-wang, 2019-zheng} present applications to unstructured and structured pruning, respectively. 

More precisely,~\citet{2019-wang} introduces a technique called EigenDamage, which consists of (1) a novel reparameterization of the neural network in the Kronecker-factored eigenbasis (KFE), and then (2) the application of the Hessian-based structured pruning framework described above, in this basis. As an intermediate technical step, the paper provides an extension of the OBD/OBS framework to the case of \emph{structured} pruning, with the key difference that the correlations between weights inside the same structure must be taken into account. The method is validated experimentally on the CIFAR-10 and Tiny-ImageNet datasets, for pruning residual networks. 

Concurrent work by~\citet{2019-zheng} used a similar K-FAC-based approximation of the Hessian, but applied it to \emph{unstructured} pruning. Relative to layer-wise pruning schemes, their approach, called MLPrune, has the advantage that it provides an approximate \emph{global} saliency metric. Specifically, this allows the user to set a global average sparsity percentage, and the technique will automatically distribute sparsity among layers, proportionally to their sensitivity to pruning.

\subsection{Selection based on regularization of the loss during training}\label{sec:regularization}

\parad{intro}
A large class of sparsification approaches uses the well-known technique of regularization, in which we add penalty terms to the cost function, for example, $L'(\vec{x},\vec{w}) = L(\vec{x}) + P(\vec{w})$. Here, $L(\vec{x})$ is the original loss function and $P(\vec{w})$ is a penalty term defined on the weights. Penalty functions can be defined with respect to arbitrary elements in the network (e.g., gating terms for neurons~\cite{2020-zhuang}) or metrics (e.g., required floating point operations~\cite{2016-molchanov}) and are generally easy to implement. The penalty will guide the search function to the desired output (e.g., sparse weights) and reduce the complexity of the model. The former leads to a sparse, smaller, and potentially faster model and the latter may lead to improved generalization. \citet{2006-mukherjee} show a strong link between stability and generalization. The choice of penalty term is most crucial for the success of the method. The resulting problem is often non-convex and can hardly be characterized theoretically. In fact, penalty terms can introduce additional local minima~\cite{NIPS1988_1c9ac015}, which makes the optimization landscape harder to navigate. Furthermore, tuning the regularization parameters often requires a delicate balancing between the normal error term and the regularization term to guide the optimization process. Even more, regularization may require fine-tuning per layer~\cite{2006-lauret}. Yet, well-tuned regularization terms are essential to deep learning training and sparsification.

\parad{weight decay}
One of the first penalty terms that was shown to significantly improve generalization was weight decay~\cite{10.5555/2986916.2987033}, where the weight update rule adds a reduction in absolute magnitude: $w' = (1-\lambda)w - \alpha g$, with the decay factor $\lambda$ and the learning rate $\alpha$. Weight decay is similar to an $L_2$ normalization for an $\alpha$-specific parameterization of the decay factor. Weight decay is a standard techniques for improving generalization today and it can be combined with magnitude pruning for sparsification. 

\subsubsection{$L_0$ norm}
\parad{L0}
The most obvious penalty term to generate sparse weights is the $L_0$ norm of the weights:
$$
P(\vec{w}) = \alpha \lVert \vec{w}\rVert_0 = \alpha \sum_i 
\begin{cases} 
0 & w_i = 0 \\
1 & w_i \neq 0
\end{cases},
$$
which simply counts the number of non-zero elements, weighted by a penalty term $\alpha$. Unfortunately, optimizing this metric directly is hard due to the discrete nature (binary, either zero or non-zero) of the problem, which cannot be differentiated. In fact, the problem is NP-complete~\cite{ge2011note}. \citet{2018-louizos} approximate the $L_0$ norm using differentiable non-negative stochastic gating variables to determine which weights to set to zero. Their method can be used with gradient-based optimization maintaining the original learning schedules. However, as with a similar method by~\citet{2017-srinivas}, it may suffer from the stochastic nature of parameter selection (see \cite{2020-savarese}): during training, new masks (weight structures) are sampled at each iteration for the forward pass. This may introduce noise into the training process if the sampling has a high variance. Furthermore, it leads to a discrepancy in the training and inference performance if a fixed deterministic sample is used at inference time. \citet{2020-verdenius} even find that tuning hyperparameters for $L_0$-based schemes is particularly hard to an extent that they could not apply the method to a different network.

\begin{figure}[h!]
	\centering
	\begin{subfigure}[b]{.3\linewidth}
	\centering
	\includegraphics[width=\linewidth]{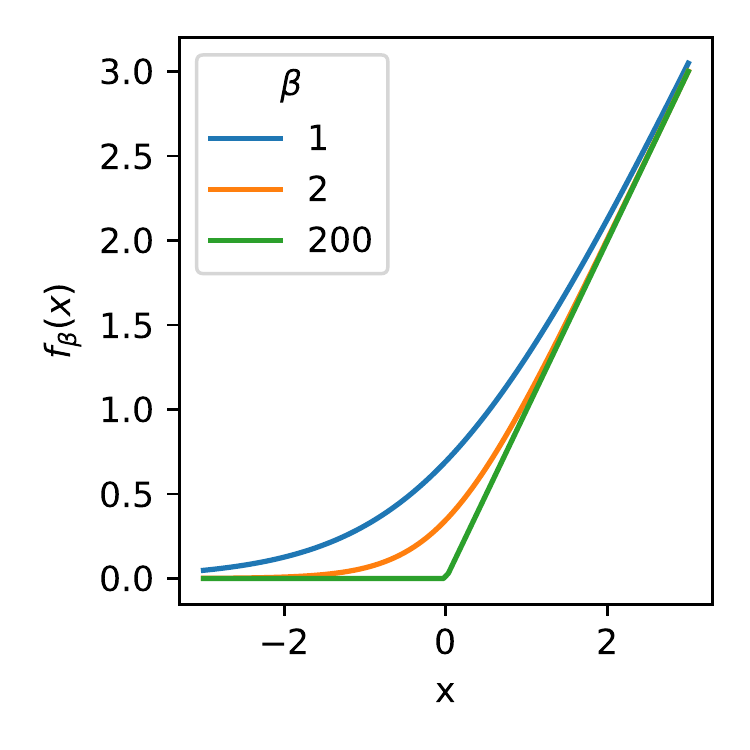}
	\caption{Softplus approximation of the ReLU function.}
	\label{fig:softplus}
	\end{subfigure}
	\quad
	\begin{subfigure}[b]{.3\linewidth}
	\centering
	\includegraphics[width=\linewidth]{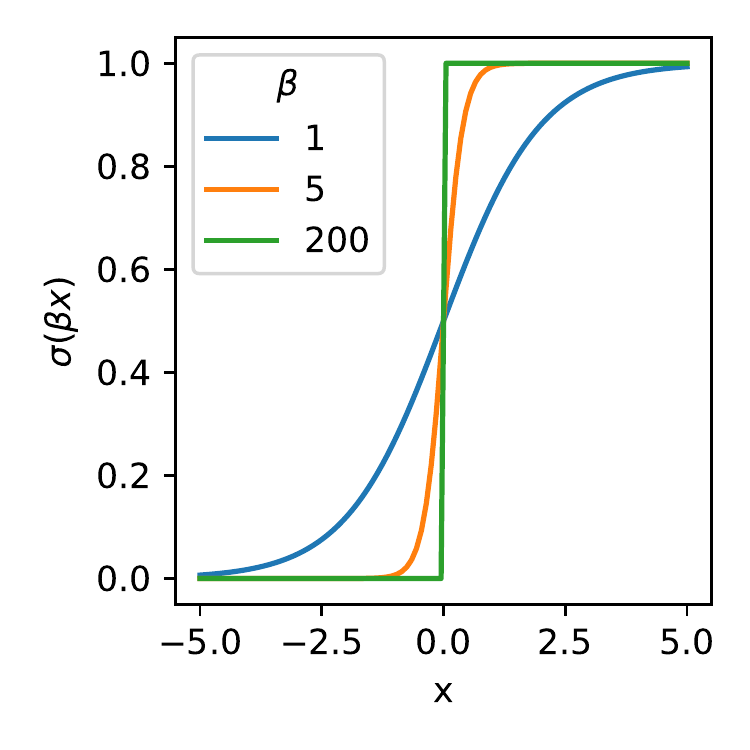}
	\caption{Sigmoid approximation of the Heaviside step function.}
	\label{fig:sigmoid_ste}
 	\end{subfigure}
	\quad
	\begin{subfigure}[b]{.3\linewidth}
	\centering
	\includegraphics[width=\linewidth]{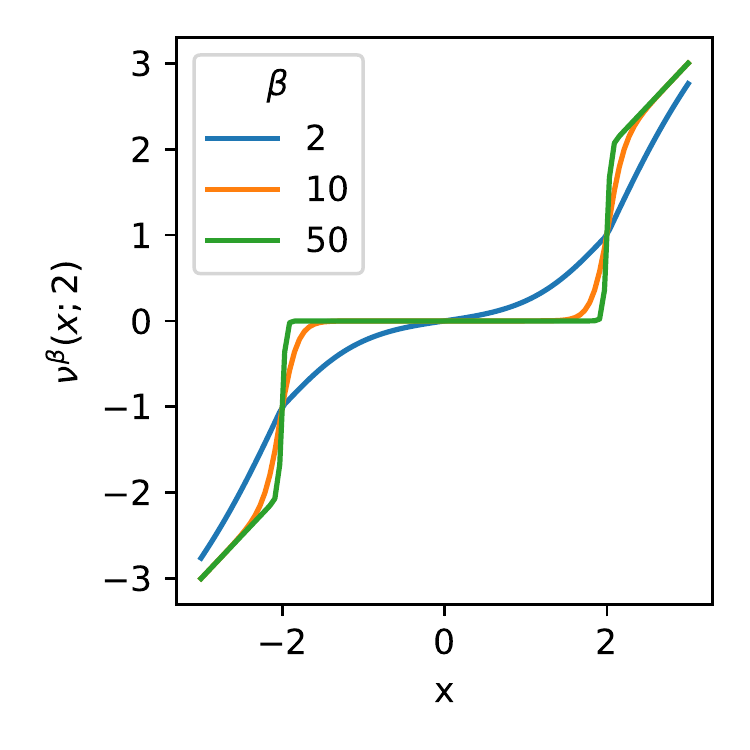}
	\caption{Approximation of magnitude pruning.}
	\label{fig:lin_thresh}
	\end{subfigure}
	\caption{Various approximations for non-differentiable step functions. The $\beta$ parameter regulates the temperature choosing between approximation quality and smoothness.}
	\label{fig:xxx}
\end{figure}

\paragraph{Estimating discrete functions}\label{sec:approx_discrete}
The main complexity lies in selecting the non-differentiable binary gating variables whose gradient is zero almost everywhere. 
The possibly simplest approach is \emph{Straight-through Estimators}~\cite{2013-bengio} that simply ignore the derivative of the non-contiguous binary function during backpropagation (treat it as if it was an identity function). Several works use this simple trick to optimize arbitrary element gating functions (\cite{2017-srinivas,2019-wortsman,2020-li-bits,2020-sanh}). Others find it to be unstable at minima and suggest variants of ReLU~\cite{yin2019understanding}.
\citet{2019-xiao} point out that hard thresholding does not support weight reanimation and they suggest ``softer'' selection functions such as the Leaky ReLU or Softplus shown in Fig.~\ref{fig:softplus}.  

A second direction to estimate discrete functions is to design parameterizable continuous approximations. 
\citet{2019-luo} and \citet{2020-savarese} choose the sigmoid function as such a continuous approximation to the Heaviside step function ($H(x)=1$ if $x>0$, and $0$ otherwise).
They introduce a varying ``temperature term'' $\beta$ to control the smoothness: $\sigma(\beta x)=\frac{1}{1+e^{-\beta x}}$. 
For high $\beta$, $\sigma(\beta x)$ approximates the Heaviside step function better but is ``harder'' to train. 
Fig.~\ref{fig:sigmoid_ste} shows the function for various values of $\beta$.
Furthermore, they continuously sparsify during deterministic training by rounding the mask to $H(x)$ in the forward pass. A key aspect of this method is the adoption of an exponential schedule for the development of $\beta$ from $1$ to an upper bound selected as a hyperparameter.
For regularization during training, they use the differentiable $L_1$ norm $||\sigma(\beta x)||_1$. 
\citet{2020-azarian} use another sigmoid-based ``soft pruning'' estimator and combine it with layer-wise threshold learning. They also observe that pruning needs to performed slowly but they use an iterative scheme with a fixed temperature but increasingly aggressive penalty parameter. 

One could also directly learn the threshold for magnitude pruning during training. \citet{2018-manessi} propose to use a soft version of the threshold linear function: $\nu^\beta(x, t) = ReLU(x-t) + t\sigma(\beta(x-t)) - ReLU(-x-t)-t\sigma(\beta(-x-t))$. Here, $t$ is the threshold parameter and $\beta$ is an approximation factor, as before. We show the varying ``sharpness'' of the curve in Fig.~\ref{fig:lin_thresh}. 
This function reduces $x$ to near-zero in the range $[-t:t]$ while $t$ can be learned through SGD. \citet{2018-manessi} then tune $\beta$ as hyperparameter and apply another fixed parameter to round the values in the learned pruning interval to zero.

\paragraph{Top-$k$}
\citet{2012-yu} and \citet{2014-collins} specify a hard limit to the number of parameters $k$ and simply prune all but the top-$k$ weights by magnitude. Both report that this scheme outperforms other ``soft'' regularization schemes. \citet{2014-collins}~define a simple greedy scheme to select layers to sparsify and thus distribute the weights. 
\citet{2019-xiao} regularize gating variables, which is essentially an $L_0$ regularizer and train it via a hard sigmoid straight-through estimator~\cite{hubara-bin}.

\paragraph{Polarization}
A related approach for pruning is polarization~\cite{2020-zhuang} where the regularizer is defined to pull some gating elements to zero and others away from zero: $$R(\vec{\alpha}) = t\lVert \vec{\alpha}\rVert_1 - \lVert \vec{\alpha}-\bar{\alpha}\mathbf{1}_n\rVert_1 = \sum_{i=1}^n t|\alpha_i| - |\alpha_i - \hat{\alpha}|,$$ where $\bar{\alpha}=\frac{1}{n}\sum_{i=1}^n \alpha_i$. The effect of the term $- \lVert \vec{\alpha}-\bar{\alpha}\mathbf{1}_n\rVert_1$ added to the $L_1$ norm is to separate small and large weights---it reaches its maximum when all $\alpha_i$ are equal and its minimum when half are equal to zero and the other half is equal~\cite{2020-zhuang}.

\subsubsection{$L_1$ norm}
\parad{L1}
The $L_1$ norm is the tightest convex relaxation of the $L_0$ norm that is almost everywhere differentiable. It has been popularized through the well-known lasso technique~\cite{tibshirani1996regression}. The left side of Fig.~\ref{fig:lasso_vs_ridge} visualizes Lasso in three dimensions. As opposed to $L_1$, the penalty is not discrete but linear, i.e., the sum of absolute magnitude of the weights:
$$
P(\vec{w}) = \alpha \lVert \vec{w}\rVert_1 = \alpha \sum_i |w_i|.
$$
While $L_1$ norms lead to very small weight values, they usually do not reduce weights to exactly zero and magnitude-based thresholding is often used to sparsify models~\cite{2014-collins}. 
\citet{1995-williams} uses a penalty term proportional to the logarithm in the $L_1$ norm to achieve better generalization through sparsification. \citet{2015-liu} use $L_1$ sparsification for convolutional networks. \citet{2020-chao} use a carefully tuned $L_1$ proximal gradient algorithm which can provably achieve directional pruning with a small learning rate after sufficient training, and show that their solution reaches similar minima ``valleys'' as SGD. 

\begin{figure}[h!]
	\centering
	\begin{subfigure}[b]{.3\linewidth}
		\centering
		\includegraphics[width=\linewidth]{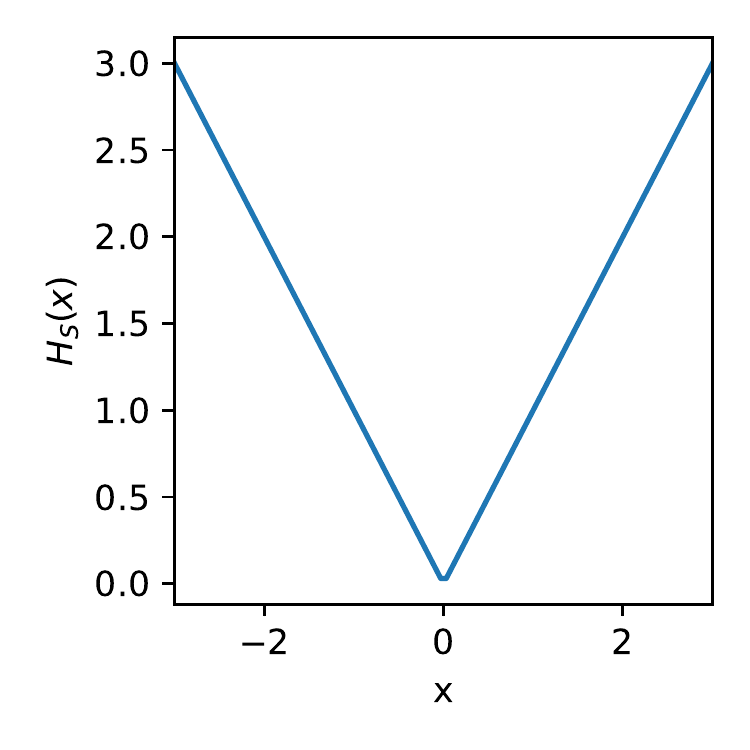}
		\caption{One-Dimensional}
		\label{fig:hoyer1d}	
	\end{subfigure}
	\qquad
	\begin{subfigure}[b]{.4\linewidth}
		\centering
		\includegraphics[width=\linewidth]{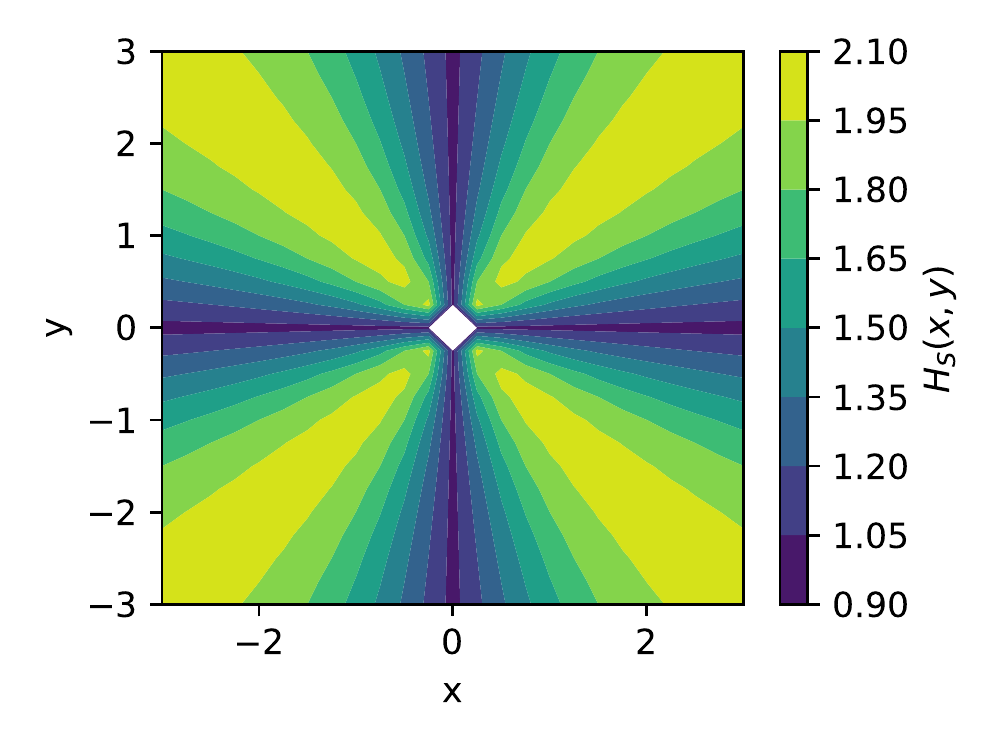}
		\caption{Two-Dimensional}
		\label{fig:hoyer2d}	
	\end{subfigure}
	\caption{Squared Hoyer regularizer for inputs with varying dimensions.}
	\label{fig:hoyer}
\end{figure}

\paragraph{Related regularization approaches}
\parad{L1 problems, Hoyer operator}
$L_1$ norm regularization has multiple shortcomings: First, it shrinks all parameters in the weight matrices with the same speed and second, it is also invariant to a scaling of the parameters, i.e., $||x \vec{w}||_1 = |x|\cdot ||\vec{w}||_1$. \citet{2019-yang} address both shortcomings by use the square of the Hoyer regularizer (Fig.~\ref{fig:hoyer}), which represents the almost anywhere differentiable scale-invariant ratio between $L_1$ and $L_2$ norms: $H_S(\vec{w}) = \frac{\left(\sum_i |w_i| \right)^2}{\sum_i w_i^2}$. This operator can also be applied in a group setting for structured pruning operations (see below).

\parad{shrinkage operator}
Another related method, the shrinkage operator~\cite{tibshirani1996regression} has significantly better empirical and theoretical properties than simple thresholding after $L_1$ regularization: $w' = (|w|-\delta)_+ sgn(w)$ with $(x)_+$ representing the positive component of $x$ and $\delta$ is acts as a weight threshold. 
This operator will zero out weights that would change sign and $\delta$ implements thresholding. 

\parad{layer-wise L1}
\paragraph{Layer-wise regularization}
While regularization as part of the overall loss function is most common, one could also imagine a layer-wise regularization to restrict the focus of the optimization problem to a smaller scope. 
\citet{2016-aghasi} use an $L_1$ norm regularizer for the weights at each layer while keeping the layer output $\epsilon$-close to the original output: $\vec{w}' = \argmin \lVert \vec{w}\rVert_1$ s.t., $\lVert \sigma_R(\vec{w}'x_{l-1})-\sigma_R(\vec{w}x_{l-1})\rVert \leq \epsilon$, where $\vec{w}'$ are the sparsified weights. For the special but very common case of ReLU ($\sigma_R(\cdot)$), they use the ``cut off'' to provide a convex relaxation to this optimization problem.

\subsubsection{Grouped regularization}
\parad{group Lasso}
The group lasso generalizes the lasso operator to a setting where variables are segmented into predefined groups, for which either all group members should be non-zero or zero together~\cite{2007-yuan}. 
We define a vector $\mathbf{y}$ of $E$ examples and a feature matrix $\mathbf{X}$ of size $E\times N$, all with mean zero. Suppose that the $N$ elements are divided into $G$ groups, and the matrix $\mathbf{X_g}$ contains only examples of group $g$ with the corresponding coefficient vector $\beta_g$ and $n_g$ is the size of group $g$.
The group lasso is defined as solving the convex optimization problem:
$$
\min_{\beta \in \mathbb{R}^p}\left(\left|\left|\mathbf{y} - \sum_{g=1}^G \mathbf{X_g}\beta_g\right|\right|_2^2 + \lambda \sum_{g=1}^G \sqrt{n_g}||\beta_g||_2\right).
$$

It is easy to see that, if all groups are of size one, the original lasso is (up to factors) recovered:
$$
\min_{\beta \in \mathbb{R}^p}\left(\frac{1}{E}||\mathbf{y}-\mathbf{X}\beta||^2_2 + \lambda ||\beta||_1 \right).
$$

\begin{figure}[h!]
	\includegraphics[width=\textwidth]{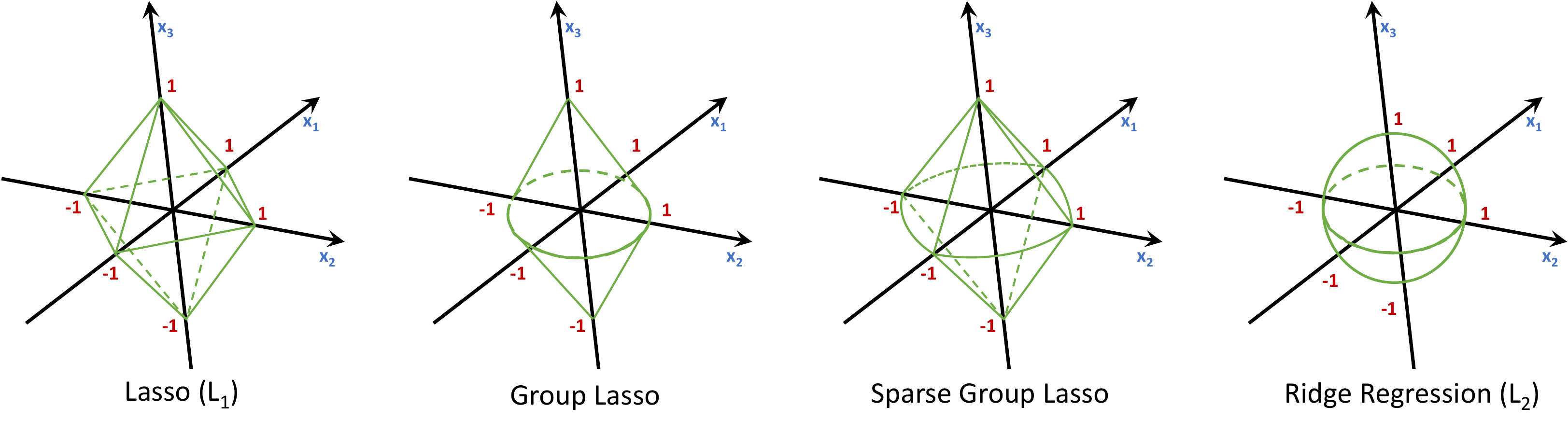}
	\caption{Lasso vs. ((Sparse) Group) Lasso with $G_1=\{x_1,x_2\}$ and $G_2=\{x_3\}$ vs. Ridge Regression.}
	\label{fig:lasso_vs_ridge}
\end{figure}

\citet{sparse_group_lasso} point out that the group lasso does not promote sparsity within groups, which can be achieved with a small tweak to the regularization term, arriving at the sparse group lasso:
$$
\min_{\beta \in \mathbb{R}^p}\left(\left|\left|\mathbf{y} - \sum_{g=1}^G \mathbf{X_g}\beta_g\right|\right|_2^2 + \lambda_1 \sum_{g=1}^G ||\beta_g||_2 + \lambda_2||\beta||_1\right).
$$
The middle two parts of Fig.~\ref{fig:lasso_vs_ridge} visualize group lasso and sparse group lasso with three dimensions and two groups. Group lasso uses a simple $L_2$ norm within each group while its sparse variant even attempts to sparsify within groups, adjustable by parameters.

\parad{uses of group Lasso}
A simple definition of such a group is to assign all outgoing weights of either input or hidden neurons to a group~\cite{2016-scardapane}. Thus, if a group is zeroed during optimization, then the corresponding neuron/input can be removed from the network. For convolutional layers, groups could be used to sparsify filters or channels~\cite{2016-wen}. At a much coarser granularity, groups could also tie whole layers together and optimize the overall model structure~\cite{2016-wen}. 
Group lasso can also be used to keep important structures of the network, such as residual connections intact~\cite{2018-gordon}.

\parad{DropNeuron}
\citet{2016-pan} use regularization on both the input and output of each neuron to facilitate neuron pruning. Their method, DropNeuron, is similar to group Lasso and they define the regularizer as the sum over all $L_2$ norms of all neuron's inputs or outputs: $L_i=\lambda_{l_i} \sum_{l=1}^L \sum_{n=1}^{n_{l-1}} ||W^l_{:,n}||_2$ and $L_o=\lambda_{l_o} \sum_{l=1}^L \sum_{n=1}^{n_{l-1}} ||W^l_{n,:}||_2$, where $W^l_{n,:}$ and $W^l_{:,n}$ are the input and output weights of neuron $n$ in layer $l$, respectively. The authors propose to use the sum of both regularization terms with carefully tuned parameters $\lambda_{l_i}$ and $\lambda_{l_o}$ as penalties to further sparsify after a magnitude pruning step on the weights. 

\parad{scaling terms}
A somewhat similar scheme adds scaling factors, which are related to gating variables, to each filter~\cite{2017-liu}. Those scaling factors can be merged into a batch normalization layer and thus do not lead to additional values. Liu et al.~then penalize the factors with an L1 norm before pruning them by magnitude globally. \citet{2018-gordon} use this scheme in a  grow/prune algorithm for neurons and \citet{2020-kang} extend the scheme to consider the effects of ReLU operations and the bias of batch normalization to also prune neurons that are mostly zero.
\citet{2018-huang} generalize this scheme and add scaling factors to neurons, groups, or whole layer blocks in various convolutional networks. They train the factors with an Accelerated Proximal Gradient method.
\citet{2018-ye} use a similar scheme by adding factors to the batch normalization of filter outputs. They use ISTA~\cite{ista} as a sparsifier for those factors, eventually pulling the output of each filter to a constant. They then remove the corresponding filter and merge the removed constant into the biases of the next-layer elements. 

\subsubsection{Other regularization techniques}
\parad{others}
Similar regularization approaches can also be used to promote low-rank matrices for the weights such that a later compression by factorization is more effective~\cite{2017-alvarez} or promote similarity of weights and filters~\cite{2019-ding-c-sgd}. Yet, such schemes are outside the scope of our work.

\parad{neuron energy term}
\citet{1988-chauvin} adds an \emph{neuron energy penalty} term $P(o) = \mu_{en} \sum_{i=1..|o|} e(o_i^2)$ over the output neurons. The positive monotonic energy function $e(\cdot)$ and the scaler $\mu_{en}$ are parameters to the method. This penalty will decrease the magnitude of the neurons and implicitly the weights, which can then be used to sparsify the network.  

\parad{output-sensitivity based penalty}
\citet{2018-tartaglione} use a penalty term that is based on the \emph{output sensitivity} of each neuron to the parameters. This sensitivity measures the relevance of the parameters to a specific output. If the sensitivity of an output neuron with respect to a specific parameter is small, then setting it to zero will change the output little. They use a regularization term to gradually decrease the absolute value of low sensitivity parameters and eventually set them to zero once they pass a certain threshold. This method can be applied during training but the authors suggest to start from a pretrained network.

\parad{quantization and pruning}
Pruning can be modeled as a special case of weight quantization by breaking the error down to the contributions of the quantization error at each bit width~\cite{2020-baalen}. They use powers-of-two bit-widths with a gating term $\alpha_i$ for each width $i$, including a general gating term for zero bits (pruned weights): $w=\alpha_2(w_2 + \alpha_4(\epsilon_4 + \alpha_8\epsilon_8)))$, where $w_i$ and $\epsilon_i$ are the weights quantized to $i$ bits and the quantization error with respect to the $i$th bit-width, respectively; $\alpha_2$ prunes the whole weight. 

\subsubsection{Potential issues}
\parad{norms and batch normalization}
\citet{2020-azarian} observe that $L_1$ (and $L_2$) regularization fails to prune networks with batch normalization layers. This is because batch normalization layers can rescale the output of previous layers arbitrarily and thus eliminate any regularization penalty. In practice, such weights would become simply very small while \emph{keeping their original relative values} (i.e., not benefiting pruning decisions), followed by an upscaling in the batch normalization layer such that model performance is not influenced. 

\subsection{Variational selection schemes}\label{sec:varbayes}

\parad{intro}
Other methods for selecting candidate elements to be removed from the network rely on a Bayesian approach or draw inspiration from the minimum description length (MDL) principle \cite{2007-grunwald}. Namely, one can assume a distribution across the elements of a neural network (e.g., over individual weights or neurons), and prune elements based on their variance. The intuition behind this approach is that elements with high variance would have little contribution to the final network performance, and therefore it might be beneficial to remove them. We now discuss methods  based on variants of this approach, and refer the reader to Section~\ref{sec:bayesian} for the relevant mathematical background.
  
\paragraph{Variational dropout}
Sparse Bayesian learning~\cite{2001-tipping} is a framework originally used for obtaining sparse models, such as the ``relevance vector machine'' (RVM), through carefully designed prior distributions, without additional manual tuning of hyperparameters. More recent advances in variational inference \cite{2013-kingma, 2014-rezende, 2015-kingma} have enabled the use of Bayesian learning techniques for large-scale models, such as neural networks. The connection between variational inference and dropout \cite{2015-kingma}, together with the idea of defining the relevance of a weight in terms of its variance during training have motivated Sparse Variational Dropout (Sparse VD) \cite{2017-molchanov} as a method for pruning neural networks. 

As described in Section \ref{sec:bayesian}, Sparse VD approximates a posterior probability distribution for each weight $\w \sim \mathcal{N}(\w | \theta, \alpha \theta^2)$, where the pair $(\theta, \alpha)$ corresponding to individual $\w$ consists of the variational parameters learned by optimizing the variational lower bound. \citet{2014-srivastava} observed empirically that Gaussian dropout has a similar performance to regular binary dropout for $\alpha = \frac{p}{1-p}$; following this observation, weights $\w$ with large values of $\alpha$, for example $\log \alpha \geq 3$, have corresponding Binary Dropout rates $p>0.95$, which suggests that these weights $\w$ can be set to zero during testing. This approach is also intuitive: large values of $\alpha$ correspond to high amounts of multiplicative noise in $\w$, which would hurt the performance of the network, unless these weights are set to $0$. The benefits of this approach are that no additional hyperparameters need to be tuned, and at the end of training the weights corresponding to large values of $\alpha$ can be dropped in one-shot, without additional fine-tuning of the sparse network. However, one disadvantage is that this new model has twice as many parameters as the original network; additionally, the authors reported difficulties in training the model from scratch and have proposed either starting from a pretrained model, or having a ``warm-up'' period in which the KL-regularizing term of the bound is gradually introduced. Although the original paper reports results only on smaller datasets such as MNIST and CIFAR-10, \citet{2019-gale} has shown that Sparse VD can also sparsify large models at ImageNet scale. We do note that in this case Sparse VD achieves high sparsity, but has high variance in the results with respect to final accuracy and average model sparsity.

One intriguing question that is not entirely resolved in the literature is whether methods such as Sparse VD applied at scale are truly ``variational''. Namely, how different are variances of the weights considered redundant, from those of the un-pruned parameters. Following the intuition presented in \cite{2017-molchanov}, for the weights $\w \sim \mathcal{N}(\theta, \sigma^2)$ corresponding to large $\alpha$ it is desirable to have $\theta=0$, which in turn favors values close to zero for $\sigma^2 = \alpha \theta^2$; this would prevent large amounts of multiplicative noise that would corrupt the model quality. 

To examine this question, we reproduced the results for CIFAR-10 presented in \cite{2017-molchanov}, focusing on the converged values of the variational parameters. Specifically, we separated the weights corresponding to large values of $\alpha$, which are eventually pruned, from the remaining weights, and studied the differences for log-variances $\log \sigma^2$. Surprisingly, all values of $\log \sigma^2$ were very close to $-15$, which was also the value used at initialization. Such a small initial value of all $\log \sigma^2$ was chosen by the authors to prevent the training process from diverging. Reproducing the same experiment at a larger scale for ResNet-50 trained on ImageNet using the implementation from \citet{2019-gale} revealed the same behavior: variances of the model's weights are all very small (close to $\e^{-15}$) and do not move during training. In this case, the threshold $\log \alpha = \log \frac{\sigma^2}{\theta^2}$ will make decisions very similar to global magnitude pruning. A distinctive behavior could be observed on Transformer networks, as implemented in \cite{2019-gale}, where the weights corresponding to large $\alpha$ generally had smaller $\log \sigma^2$ than the pruned weights, while the values of $\log \sigma^2$ moved significantly from their initial value. In spite of the intriguing observation that for CNNs, Sparse VD has a very similar behavior to global magnitude pruning, it is worth noting that for models trained using variational dropout, a large proportion of the weights can be pruned immediately after training, with a small drop in test accuracy. This is in contrast with magnitude pruning methods, which require fine-tuning to recover from the drop in performance, and suggests a powerful regularization effect in Sparse VD, which is not always reflected in the final variances of the weights.


\paragraph{Structured Bayesian pruning}
Although Sparse VD can lead to sparse neural networks, the unstructured sparsity achieved can rarely accelerate inference today. If the goal is acceleration, then structured sparsity is a more desirable outcome, and \citet{2017-neklyudov} showed how this can be achieved using the Bayesian dropout framework. The authors propose using a truncated log-normal distribution as the approximate posterior, where $\theta \sim \text{LogN}(\mu, \sigma^2) \iff \log(\theta) \sim \mathcal{N}(\mu, \sigma^2)$; here the variational parameters $(\mu, \sigma^2)$ are shared across different groups, such as neurons or convolutional filters. This has the advantage that log-normal noise does not change the sign of its input, as the noise is non-negative both during train and test. Furthermore, using truncated versions of both the log-uniform prior and log-normal posterior gives a closed form solution of the KL divergence term used in the variational lower-bound. To obtain a sparse solution, the authors propose thresholding neurons by their corresponding signal-to-noise ratio (SNR); intuitively, neurons with low SNR are mostly propagating noise and therefore should be set to zero. The authors show acceleration for their method on smaller datasets, such as MNIST and CIFAR-10.   



\paragraph{Soft weight sharing} \citet{2017-ullrich} propose combining soft weight sharing with pruning to compress neural networks. The idea of soft weight sharing \cite{1992-nowlan} is to compress a neural network by assigning its weights to different clusters. This is done using empirical Bayes methods, in which the prior over the parameters is learned during the training process. Following \citet{1992-nowlan}, \citet{2017-ullrich} define the prior over the weights of a neural network as a mixture of Gaussians. One of the mixture components has a zero mean and a chosen mixture probability close to one, which will enforce a certain sparsity level for the resulting neural network. Thus, the proposed soft weight-sharing algorithm for compression starts from a pre-trained network and after optimizing the corresponding variational lower-bound, the resulting weights are assigned to the most probable cluster from the Gaussian mixture prior.

\paragraph{Bayesian pruning with hierarchical priors} \citet{2017-louizos-bayes} use the variational inference framework and the minimum description length (MDL) principle to compress neural networks, by defining hierarchical sparsity inducing priors to prune neurons. The MDL principle \cite{2007-grunwald} states that the best hypothesis is the one that uses the smallest number of bits to communicate the sum between the model's complexity cost and the data misfit error; thus, MDL is directly related to compression. Additionally, it has been well understood that variational inference can be reinterpreted through MDL \cite{1993-hinton}. With this theoretical support, \citet{2017-louizos-bayes} define a zero-mean Gaussian prior over the weights of a neural network, where the variance is sampled from a separate distribution, for example a log-uniform or half-Cauchy. This formulation enables weights within the same neuron or feature map to share the corresponding scale variable in the joint prior, which encourages structured sparsity. Furthermore, the optimal fixed point precision for encoding the weights can be determined from the posterior uncertainties, which in turn leads to quantized networks. 

\paragraph{Bayesian pruning for recurrent neural networks} Earlier works have focused on inducing sparsity in standard feed-forward neural networks. Yet, Bayesian pruning methods have also been successfully applied to recurrent neural networks (RNNs) \cite{2018-lobacheva, 2019-kodryan}. \citet{2018-lobacheva} use Sparse VD \cite{2017-molchanov} to prune individual weights of an LSTM or follow the approach from \citet{2017-louizos-bayes} to sparsify neurons or gates and show results on text classification or language modeling problems. \citet{2019-kodryan} use instead the Automatic Relevance Determination (ARD) framework, in which a zero-mean element-wise factorized Gaussian prior distribution over the parameters is used, together with a corresponding Gaussian factorized posterior, such that a closed-form expression of the KL divergence term of the variational lower bound is obtained. Subsequently, the Doubly Stochastic Variational Inference (DSVI) method is used to maximize the variational lower bound and the weights for which the prior variances are lower than a certain threshold are set to zero.

\paragraph{Related methods} \citet{2018-dai} prune neurons based on a simple layer-wise information bottleneck, an information-theoretic measure of redundancy. For this, they penalize the ``inter-layer mutual information using a variational approximation'' to sparsify. Their Variational Information Bottleneck Networks modify the loss function to contain a term that compares the mutual information from layer $i$ to layer $i+1$ with the mutual information between layer $i$ and the final result. With the optimization goal to minimize the former and maximize the latter, they prune based on their KL-divergence. 
\citet{2001-engelbrecht} prunes based on the variance of sensitivity of inputs and neurons, and could therefore be seen as variational. Specifically, their method dictates that if the sensitivity of a parameter varies very little across the training set, then it can be pruned. 
 

\subsection{Other selection schemes}

\subsubsection{Genetic algorithms}\label{sec:genetic-algorithms}

Like any optimization problem, pruning can also be modeled using genetic algorithms~\cite{whitley:ijcnn90,10.5555/646365.691221}. The population is created from multiple pruned versions of the neural network and each is trained separately. New networks are created using mutation, reproduction, and cross-over parameter selection. These populations are then rewarded for smaller numbers of parameters and for improved generalization. However, this approach is not practical for modern large compute-intensive training due to the high complexity of training ensembles of models.

\subsubsection{Sampling-based pruning with guarantees} 
Another method for selecting candidate elements for pruning relies on an approach different from the Bayesian framework. Namely, \citet{2018-baykal}, propose using a subset of the data to estimate the relative importance, or ``empirical sensitivity'' of incoming edges to a neuron; this allows the definition of an importance sampling distribution over the incoming edges, which in turn leads to sparse weight matrices. The proposed algorithm has theoretical guarantees in terms of the sparsity level obtained, as well as generalization guarantees for the sparse network. Furthermore, the framework can be improved to allow for structured pruning of neurons. Following work \cite{2019-lieberwein} has extended the idea of sampling-based pruning to removing filters from CNNs, while also providing guarantees on the size and final output of the pruned network.  

\subsubsection{Diversity and quantized networks}

\parad{divnet}
Diversity networks~\cite{2015-mariet} employ Determinantal Point Processes to select a subset of ``diverse neurons'' in each layer while fusing other similar neurons. It starts from fully-trained networks and does not require fine-tuning.

\parad{binary and quantized networks}
Quantized neural networks already employ an approximation function that could also be used to guide pruning decisions. 
\citet{2020-guerra} use a metric related to the distance between quantized and full-precision weights (i.e., the rounding error) in binary or quantized networks for selecting filters to prune. 

\parad{scientific control}
Some neurons or filters may learn properties of the training set distribution that are not relevant to distinguish between classes within that distribution. \citet{2020-tang-scop} propose to generate ``knockoff'' features that draw from the same distribution but are independent of the example's label. They feed the example and the knock-off into the same network and compare scaling factors for filters (cf.~filter sensitivity). Then they prune the features that have a large sensitivity for knockoff inputs and a relatively small sensitivity for real inputs. 

\subsection{Parameter budgets between different layers}

\parad{intro}
All of these schemes define several hyperparameters to adjust sparsity --- be it based on the value of the elements themselves, or be it based on a target sparsity level (\emph{top-$k$}). One remaining question is about whether or not these parameters should be chosen per layer/operator or globally for the whole model. 

\parad{global sparsity and manually tuned per layer}
Earlier works implicitly choose the sparsity level globally, such as ``drop the bottom 90\% of all parameters $\vec{w}=\vec{w_1} \cup \vec{w_2} \cup \cdots \cup \vec{w_{\ell}}$''. \citet{2016-see} found that global selection without differentiating layers performs best for pruning of RNNs. It was recognized soon that, especially for networks with very different layer types, e.g., convolutional and densely connected, different layers should be treated differently. Furthermore, empirical evidence suggests that even the same layer types should be sparsified differently depending on their position in the network (earlier vs. later layers). One can now consider introducing different sparsities for each layer separately~\cite{2018-mocanu}, requiring to tune potentially complex hyperparameters. 

\parad{automatic per layer sparsity}
Later schemes automatically determine a good parameter budget per layer to reduce the hyperparameter complexity. A simple option would be to 
link the sparsity to properties of the layer, such as the ratio of weights to neurons or kernel dimensionality~\cite{2020-evci}.
Parameter budgets can also be redistributed during training depending on various saliency metrics. 
For example, \citet{2019-mostafa} drop small magnitude parameters during training and preferentially re-add parameters in layers with larger loss gradients (i.e., layers that have been pruned less). 

\begin{figure}[h!]
	\includegraphics[width=\textwidth]{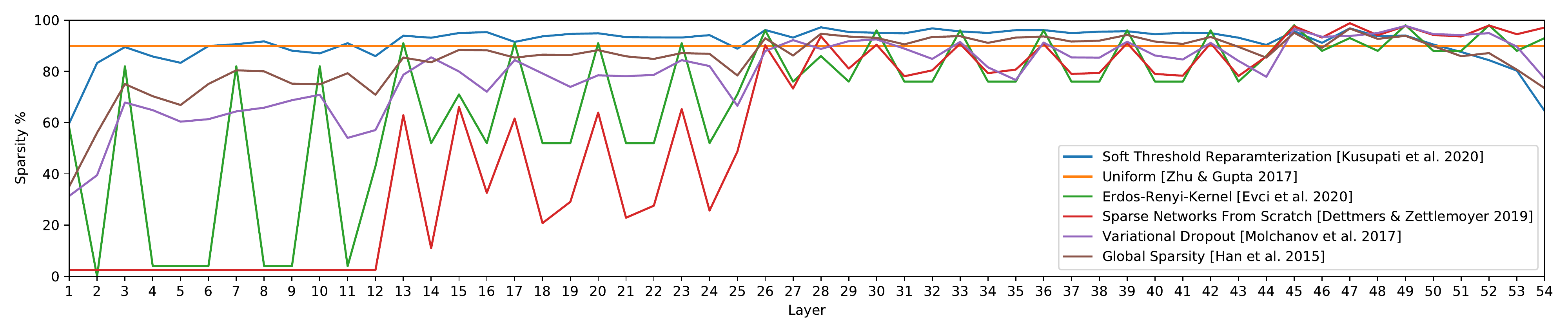}
	\caption{Distribution of sparsity across layers for ResNet-50 and various sparsification methods.}
	\label{fig:layer_sparsity}
\end{figure}

Figure~\ref{fig:layer_sparsity} shows the distribution of sparsity across the various layers of a ResNet-50 network for different methods (see Sections~\ref{sec:removal} and~\ref{sec:convarch} for details).
\parad{general observations}
An interesting and seemingly general observation is that many global schemes that can balance the parameters across layers automatically tend to assign more parameters to earlier layers than later ones~\cite{2020-sanh}. Many practitioners even disable pruning of the first layers because they empirically found this to yield higher accuracy. Tuning sparsity across layers is an important consideration for practical sparsification.

\subsection{Literature overview}

\begin{figure}[h!]
	\centering
	\begin{subfigure}[b]{.32\linewidth}
		\centering
		\includegraphics[width=\linewidth]{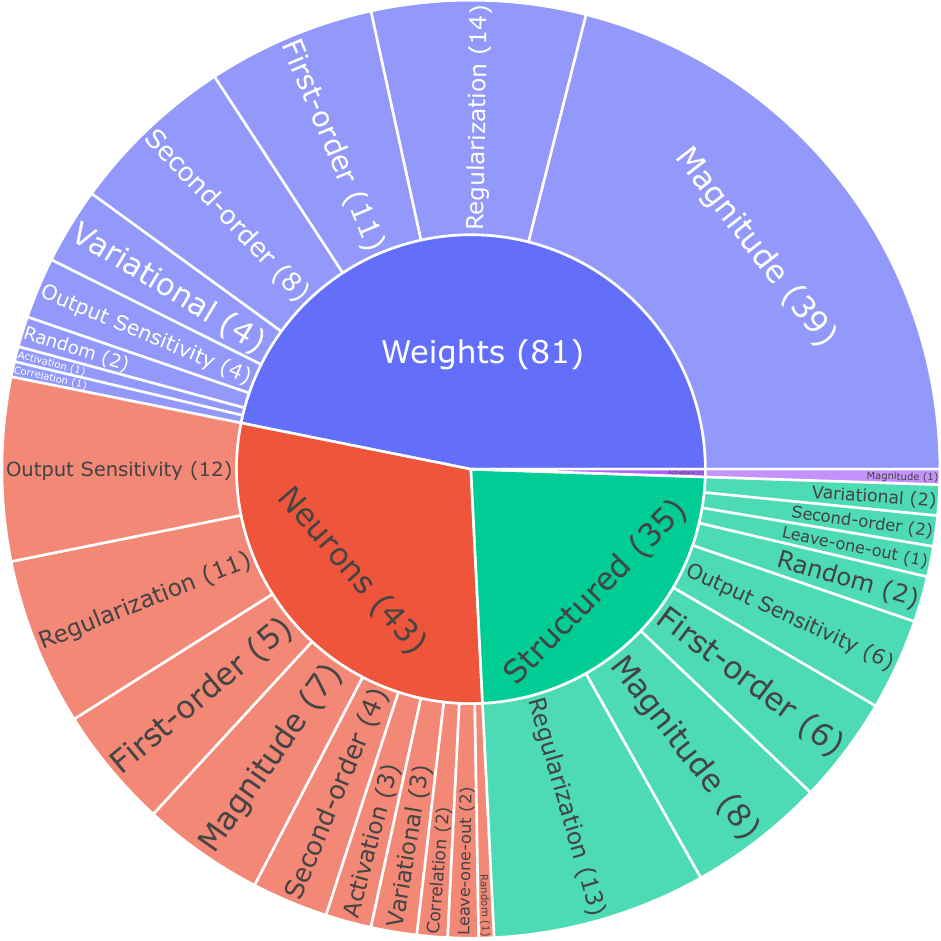}
		\caption{Selection method by candidate.}
		\label{fig:sbef}	
	\end{subfigure}
	\begin{subfigure}[b]{.32\linewidth}
		\centering
		\includegraphics[width=\linewidth]{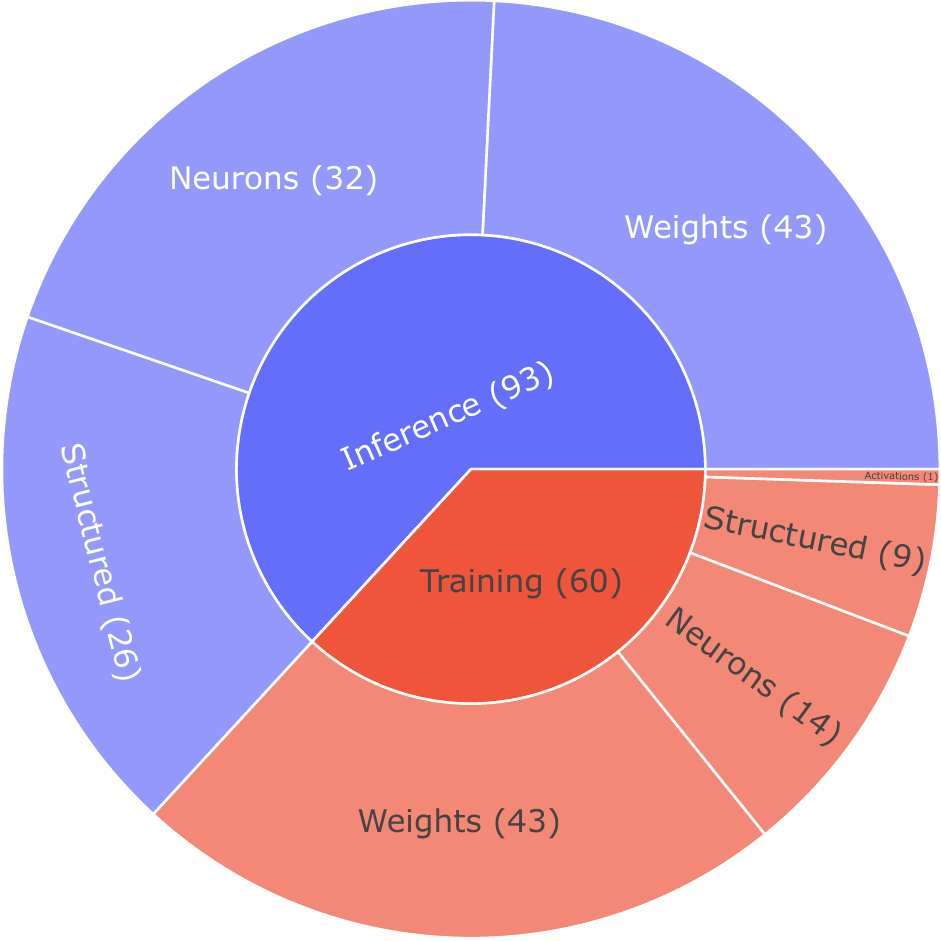}
		\caption{Candidates by type.}
		\label{fig:sbde}	
	\end{subfigure}
	\begin{subfigure}[b]{.32\linewidth}
		\centering
		\includegraphics[width=\linewidth]{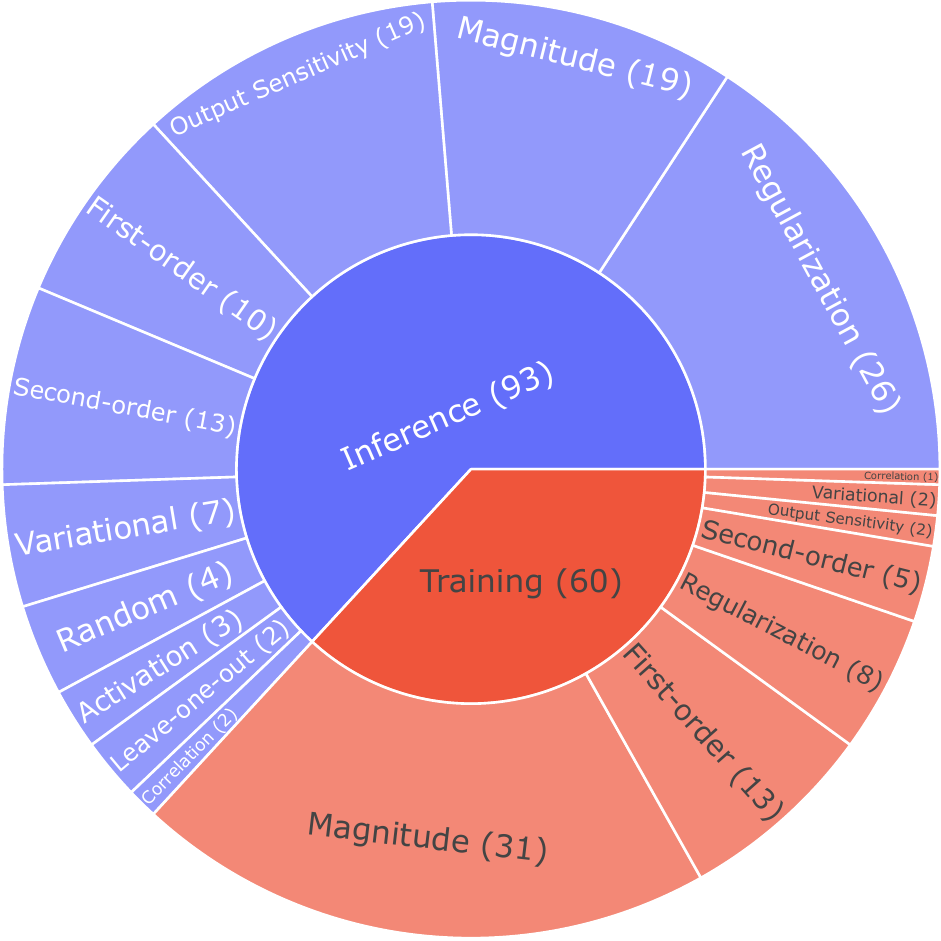}
		\caption{Methods by type.}
		\label{fig:sbdf}	
	\end{subfigure}
	\caption{Statistics of how many papers combine a specific selection method, to prune a specific candidate element for training or inference (type).}
	\label{fig:pruning_literature1}
\end{figure}

\begin{figure}[h!]
	\centering
	\begin{subfigure}[b]{.4\linewidth}
		\centering
		\includegraphics[width=\linewidth]{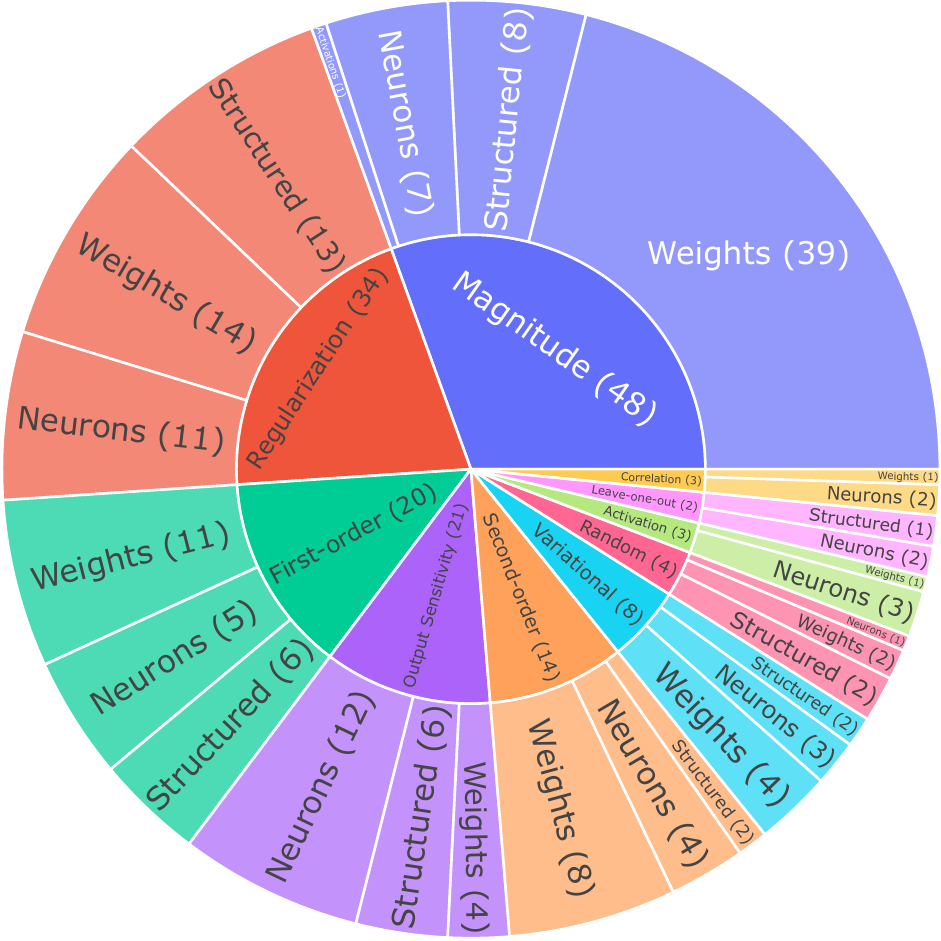}
		\caption{Candidates by selection method.}
		\label{fig:sbfe}	
	\end{subfigure}
	\quad\quad
	\begin{subfigure}[b]{.4\linewidth}
		\centering
		\includegraphics[width=\linewidth]{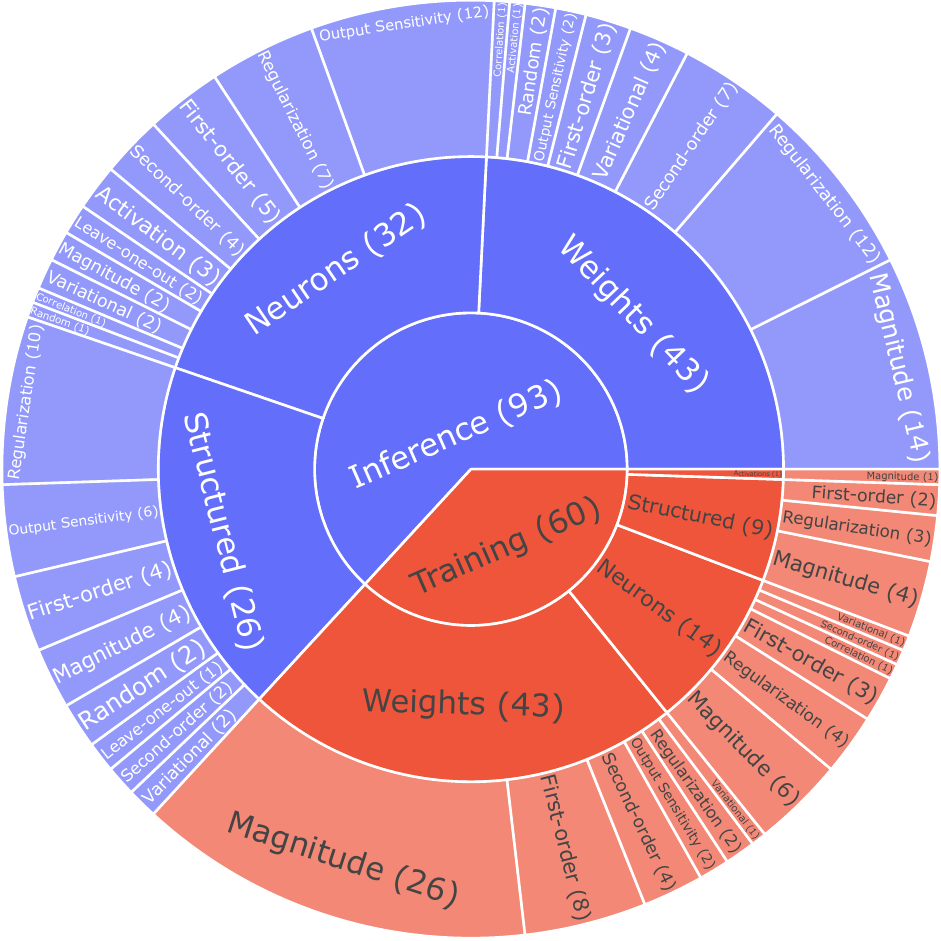}
		\caption{Method by candidates by type.}
		\label{fig:sbdef}	
	\end{subfigure}
	\quad
	\caption{Statistics of how many papers combine a specific selection method, to prune a specific candidate element for training or inference (type).}
	\label{fig:pruning_literature2}
\end{figure}

After describing the flurry of different approaches, we attempt to overview the landscape of the literature to provide some information about the popularity of the various techniques.
Figures~\ref{fig:pruning_literature1} and~\ref{fig:pruning_literature2} show various different views of the same data summarizing all surveyed papers from 1988 to 2020. 
We classified each paper in three different categories: (1) the candidate element to be removed, (2) the method to choose elements for removal, and (3) whether the authors discuss optimizing inference or improving training (type). 
The different candidate elements are, as described in Section~\ref{sec:candidates}, neurons, weights, convolutional filters, transformer heads, transformer hidden dimensions, and inputs. 
The different methods follow the structure of this section. 

Fig.~\ref{fig:sbef} shows that nearly 50\% of all papers focus on weight sparsification, closely followed by neuron sparsification. Other structured schemes and inputs form a minority. Of the weight sparsification schemes, the vast majority uses simple magnitude pruning followed by first and second order schemes. 
Fig.~\ref{fig:sbde} shows that more than 60\% of the papers focus on inference while training is recently gaining popularity. Most inference works focus on pruning either neurons, weights, or filters while pruning to improve training largely focuses on weights. 
Fig.~\ref{fig:sbdf} allows us to compare popular pruning methods for inference and training. Inference is interestingly dominated by regularization approaches, closely followed by magnitude pruning while training focuses on magnitude.

Fig.~\ref{fig:sbfe} shows that 50\% of the works focus on either magnitude pruning or regularization. Magnitude pruning is most often used for weights while regularization is equally applied to all element types. Here, we summarize filters, blocks, and heads into a single ``structured'' category.
Fig.~\ref{fig:sbdef} shows an overview including all three classification dimensions. It illustrates once more the dominance of pruning weights by magnitude, followed by sensitivity-based neuron pruning. 

\section{Dynamic Pruning: Network regrowth during training}\label{sec:growth}

\parad{intro}
Fully-sparse training schedules remove elements during training but also need to re-add other elements in order to ensure that the model remains of approximately the same size. The process is very similar to architecture search in that it traverses the space of possible model architectures. If we prune and re-add neurons, then relatively simple schemes to add new neurons perform  well~\cite{2008-narasimha,2012-han} because the order of neurons in a layer is insignificant.
However, weights are more complex because re-adding the best weights is as crucial as removing the right weights. Yet, it is often much harder because the information for all non-existent weights is the same: they are zero. Additional hints, such as the gradient or Hessian magnitude could be used but cause additional overheads in terms of memory and compute and invalidate some of the benefits of sparsity. We will now describe the various schemes put forward to select weights to add to a sparse model during training.


\subsection{Random or uniform regrowth}

\parad{intro}
The simplest weight addition scheme is to activate a random new weight during training, which essentially leads to a random walk in the model space~\cite{2018-bellec}. \citet{2018-mocanu} show that this scheme leads eventually to power-law graphs following a preferential attachment rule. They also draw parallels to biological brains and argue that weight removal can be seen as natural selection, similar to synaptic shrinking during sleep~\cite{de2017ultrastructural,diering2017homer1a}, and weight addition can be seen as natural mutation.

\parad{layer-wise}
Similar to layer-wise pruning, layer-wise addition can also lead to improved accuracy. The main idea is to add parameters preferentially in layers that would be sparsified less. \citet{2019-mostafa} initially distribute all parameters according to a fixed fraction to all layers. After magnitude pruning, they add new parameters proportionally to the number of parameters retained in each layer to strengthen the significant layers. 

\parad{uniform addition of filters/neurons}
Uniformly adding filters or neurons via a ``width multiplier'' to layers as part of an iterative grow/prune methodology has also been shown to be effective~\cite{2018-gordon}.

\parad{use initial weights}
Based on the observation that the optimization process benefits from large dense models (see Section~\ref{sec:opt}), one could argue that learning in a dense space should be beneficial. \citet{2019-golub} realize that the initial (random) weight values that were pruned influence the non-pruned weights during the optimization process (not at inference, where they are removed). Since those weights have been generated with a pseudo-random number generator, the authors propose to simply recompute them on demand for training. 

\subsection{Based on gradient information}

\parad{intro}
One simple way to determine which weights should be added is to observe the gradients during the backwards pass, including those gradients for zero weights. While this immensely increases memory and computation overheads and removes some of the benefits of sparse computations, it provides good information about the importance of specific weights: if the gradients are large, then those weights would be important. 
\parad{biological relation}
\citet{2019-dai} show that adding weights by largest gradient is biologically plausible and related to Hebbian learning. They show that this scheme is mathematically identical to adding a new synapse between two highly stimulated neurons in adjacent layers. 

\parad{full matrix}
The simplest version of this scheme to to compute gradients for all parameters. 
\citet{2020-lin} keep a full copy of all weights up to date during training and only deactivate them with a mask in the forward pass. This enables an efficient search through different architectures with various pruning methods for the dense model. They show good results using simple magnitude pruning every $k$ iterations. 
\citet{2019-wortsman} uses a similar scheme and but restricts the flow of gradient information through pruned weights. In this scheme, gradients flow \emph{to} pruned weights but their contribution is not forwarded to other weights.
\citet{2019-dettmers} compute a momentum term that captures gradients over multiple iterations as a criterion for growing weights. While the memory and compute overheads are significant, these methods still reduce the number of arithmetic operations substantially compared to dense training. They can be combined with layer-wise redistribution strategies to focus the addition of new neurons to more efficient layers. 
\citet{2019-dettmers} find in an ablation study that updating pruned weights during training is critical for final model accuracy.

\parad{neuron addition across different layers}
One way to reduce gradient storage is to compute it only layer-by-layer and discard it after  layer-wise regrowth decisions~\cite{2020-evci}. This reduces the memory overheads but potentially decreases the accuracy due to noise in the instantaneous gradients. They use three different schemes to determine the number of parameters per layer: (1) the same uniform fraction for each layer, (2) scaling the number of weights with the number of neuron's (``Erd\H{o}s R\`enyi''), and (3) incorporating the kernel dimension into the scaling factor. 
\parad{top-k gradient matrix}
Another way to reduce gradient storage is to only compute the top-($k+d$) gradients~\cite{2020-jayakumar} for $k$ non-zero weights. In this way, the additional $d$ gradients can be seen as a ``halo zone'' of the most relevant gradients to be added. 

\subsection{Locality-based and greedy regrowth}

\parad{intro}
Biological brains are sparse structures with hierarchical sparsity distributions that are locally dense and globally sparse~\cite{betzel2017modular}. It is now perceivable that local connectivity could also benefit deep neural networks. \citet{1997-stroem} decays the probability for adding a new weight exponentially with the distance between neurons, leading to a hierarchically sparse structure. 

\parad{greedy}
Simple greedy schemes that start from a trained network, remove all neurons and add the most beneficial neurons provide theoretical guarantees albeit with limited sparsification. \citet{2020-ye} show a scheme that adds neurons based on maximum loss reduction and \citet{2019-zhuang} add filters based on minimizing a gradient-based sensitivity.

\section{Ephemeral Sparsification Approaches}\label{sec:dynamic}

In biological brains, model sparsity is one important component. However, activity sparsity is at least as important: the connections among neurons are fixed on a longer time-scale ranging from hours to days while the electrical signals appear and disappear on a millisecond time-scale. Not all 86 billion neurons of the human brain are active at any moment and are controlled by complex activation and inhibition signals. While it is hard to estimate the exact activity factor of this asynchronous system, several works suggest that only around 10\% of the neurons are active at any moment~\cite{Kerr14063}. This is necessary to keep the human brain's energy budget around 20W ($\approx$20\% of a typical human's operating budget, as the most expensive organ).

Deep neural networks use ephemeral sparsification to mimic that behavior: activation functions such as ReLU inhibit certain signals by shutting down whole paths through the network, implicitly selecting the information-rich paths specific to each input problem. 
We can also extend ephemeral sparsity to the backpropagation learning process where we can sparsify gradients and errors during training. 
Ephemeral sparsity has initially been used as a regularizer but it is increasingly seen as another opportunity to save memory and energy during processing of neural networks. 

We start by describing inference sparsification where neural activations are set to zero during inference and the forward pass of training. 
Then we consider sparsification during training. We start with the various forms of dropout, a set of techniques to sparsify networks during the forward pass of training to improve generalization. Gradient sparsification has received special attention to reduce the communication overheads in distributed data parallelism~\cite{bennun2018demystifying}. We then discuss less common options to sparsify back-propagated errors between layers and the optimizer state.

\subsection{Sparsifying neuron activations}\label{sec:activations}

The output activations of any ReLU-based neural network layer are naturally sparse~\cite{2011-glorot} since, intuitively, on random inputs, half of the output values of such a layer would be zero. 
In practice, it appears that the fraction of sparse activations is significantly higher than 50\%. 
This phenomenon does not currently have an analytical explanation, but it has been leveraged by several hardware architecture proposals (see Section~\ref{sec:hwacc}). 
Specifically,~\citet{2018-rhu} were among the first to perform an in-depth analysis of activation sparsity on a range of large-scale convolutional models with ReLU activations, showing high sparsities of up to $90\%$ in some layers, well in excess of the $50\%$ predicted by the structure of the ReLU activation. 

This phenomenon has inspired a line of work on compressing the activation maps in a neural network for memory and computational gains, and potentially augmenting this natural sparsity. 
The standard technique for reducing the memory footprint of activation maps is \emph{quantization}, see e.g.,~\citet{2017-mishra}. 
Since quantization is not the main focus of this work, we do not detail this approach here. 
For sparsifying activations,~\citet{2016-alwani} suggested to \emph{stochastically} prune activations, although the objective is not to gain performance, but to design a defense to adversarial attacks. To further reduce the size of activations, \citet{2018-gudovskiy} suggested converting fixed-point activations into vectors over the smallest  finite field $GF(2)$ followed by nonlinear dimensionality reduction (NDR) layers embedded into the structure of the neural network. The technique results in a factor of two decrease in memory requirements with only minor degradation in accuracy, while adding only bitwise computations. At the same time, we note that the technique requires modifying the network structure, and additional retraining. Both these techniques incur low, but persistent, accuracy loss. Activation sparsity can also be used to significantly reduce memory consumption during the training process~\cite{2019-liu-dynamic}.

More recently, \citet{2019-georgiadis} proposed and investigated the use of $L_1$-regularization applied to the activation maps, and showed that it can result in a significant increase in sparsity---up to 60\% relative to naturally-occurring activation sparsity on a range of CNNs for image classification on ImageNet. 
Further, he investigated a range of encoding techniques for the activations, and evaluated them in terms of their resulting compression factors. 
\citet{2020-kurtz} followed up on this idea, and showed that Hoyer regularization~\cite{hoyer2004non}, a popular regularizer in the context of sparse recovery, is superior to $L_1$ regularization, in the sense that it provides higher activation sparsity with lower accuracy loss. 
The paper goes on to introduce a series of thresholding methods that are complementary to regularization, in the sense that they zero out activation values that are close to, but not exactly, zero. 
In addition, this paper describes a complete set of algorithms for leveraging activation sparsity for fast inference on CPUs, showing end-to-end inference speedup for activation-sparsified models. 
Concurrent work by~\citet{2019-dong} also introduced an algorithmic framework for obtaining computational speedups on models where layers have extremely high input sparsity. 
Their method is different from~\citet{2020-kurtz}, but appears to require higher input sparsity to ensure speedup. In particular, it is applied to tasks such as LiDAR-based detection, or character recognition, in which inputs (and therefore further activations) are naturally extremely sparse.

Other operators such as GELU or SoftMax may also sparsify to some degree, be it through rounding towards zero with limited precision.  Since those two operators are often used in transformers, see Section~\ref{sec:tformers}.

\subsection{Dropout techniques for training} \label{sec:dropout}

\parad{Description}
Dropout~\cite{2014-srivastava,2012-hinton} is a regularizing operator in DNNs that forces the network to ``prevent co-adaptation'' of neurons during training. 
Specifically, dropout is a data-free sparsifier that uses Bernoulli random sampling (with $p$ typically ranging from $0.01$ to $0.5$) to zero out neurons and nullify their contributions. During training, the neuron-masking vector, which is randomly sampled at every step, is kept stored in memory in order to mark the neurons to be ignored during backpropagation. At inference-time, no dropout masks are applied, i.e., the entire set of neurons is considered.
The operator is applied mostly on the activations of fully connected layers, and is widely used to increase generalization in MLPs, CNNs, and Transformers.
\parad{Dropout and sparsity}
An interesting property of dropout is that it induces sparsity in activations~\cite{2014-srivastava}, likely due to the repeated ephemeral sparsification. The sparsity factor was observed to increase with the dropout probability $p$. 

\parad{Interpretations}
There are several interpretations to dropout's generalization effect. 
The initial line of research claims that neuron ``co-adaptation'' (a concept borrowed from genetics) harms generalization, and dropout prevents it by ``making the presence of other hidden units unreliable''~\cite{2014-srivastava}. 
\citet{2013-baldi} characterize dropout in neural networks as simultaneously training an ensemble of an exponentially large set of networks, each one generated by the different masked versions, and that at inference-time their sum is taken (similarly to ensembles).
Another interpretation originates from Bayesian statistics~\cite{2016-gal,2017-molchanov}. The claim is that dropout is an approximating distribution to the posterior in a Bayesian neural network with a set of random weights. It is shown that dropout's minimization objective reduces the epistemic uncertainty of a DNN, or more specifically the KL-divergence with a Gaussian process~\cite{2016-gal}.

\parad{Other variants of dropout}
Over the years, several successful extensions and generalizations of dropout were proposed. 
DropConnect~\cite{2013-wan} drops out weights instead of activations.
\citet{2014-srivastava} proposed to replace the Bernoulli distribution with a normal $\mathcal{N}(1,1)$ distribution in order to add multiplicative noise. 
Other variants of dropout specialize to certain operators: For convolutions, instead of random activation subsets, SpatialDropout~\cite{2015-tompson} drops entire feature maps, and DropBlock~\cite{2018-ghiasi} drops contiguous spatial regions.
For recurrent neural network units, ZoneOut~\cite{2017-krueger} modifies information propagation through sequences by randomly selecting between the old hidden state and the new hidden state of the RNN unit, dropping the hidden state update.
Stochastic Depth~\cite{2016-huang}, Drop-Path~\cite{2017-larsson}, and LayerDrop~\cite{2020-fan} are more coarse-grained versions of dropout, dropping layer weights and outputs of entire subgraphs of DNNs to prevent co-adaptation of paths and increase regularization.

The variational interpretation has been used to generalize the dropout operator in various ways.
Concrete Dropout~\cite{2017-gal} uses the Concrete distribution~\cite{2017-maddison} instead of Bernoulli sampling, which results in increased generalization as well as the ability to evaluate epistemic uncertainty of the results.
Variational dropout~\cite{2015-kingma} uses Bayesian principles to define a variational dropout probability specific to each neuron based on measured noise during training, foregoing the data-free property of dropout to reduce the gradient variance. \citet{2017-molchanov} makes use of variational dropout to select weights to prune (see Section~\ref{sec:varbayes}).

\citet{2019-gomez} also propose a modification to the original dropout procedure to ``prepare'' the learned network structure for pruning. Their targeted dropout stochastically selects a set of weights or neurons to drop that may be pruned later. Specifically, they rank weights and neurons (activation outputs) by their magnitude and apply dropout only to a fraction of those deemed less important. For this, they select the $\gamma|W|$ elements with lowest magnitude and drop each of those with probability $\alpha$. This scheme allows lower-valued elements to emerge from the set of unimportant values during training.

\subsection{Gradients}\label{sec:gradients}

\parad{Intro/overview; discuss broader area.}
Gradient sparsification aims to introduce sparsity in the gradients of parameters during training.
While there are exceptions, this is primarily done in order to compress the gradients communicated as part of distributed data-parallel training (see \citet{bennun2018demystifying} for an overview).
In this context, gradient sparsification is a subset of the more general area of communication compression, which also includes quantization and low-rank approximations (see \citet{2020-tang} for a broad overview of this area).
The key intuition is that the gradients produced by SGD are noisy, and therefore identifying and removing unimportant sub-components should not have a significant impact on convergence or may even have a regularizing effect, while enabling compression.

\begin{figure}[h!]
	\centering
	\includegraphics[width=\linewidth]{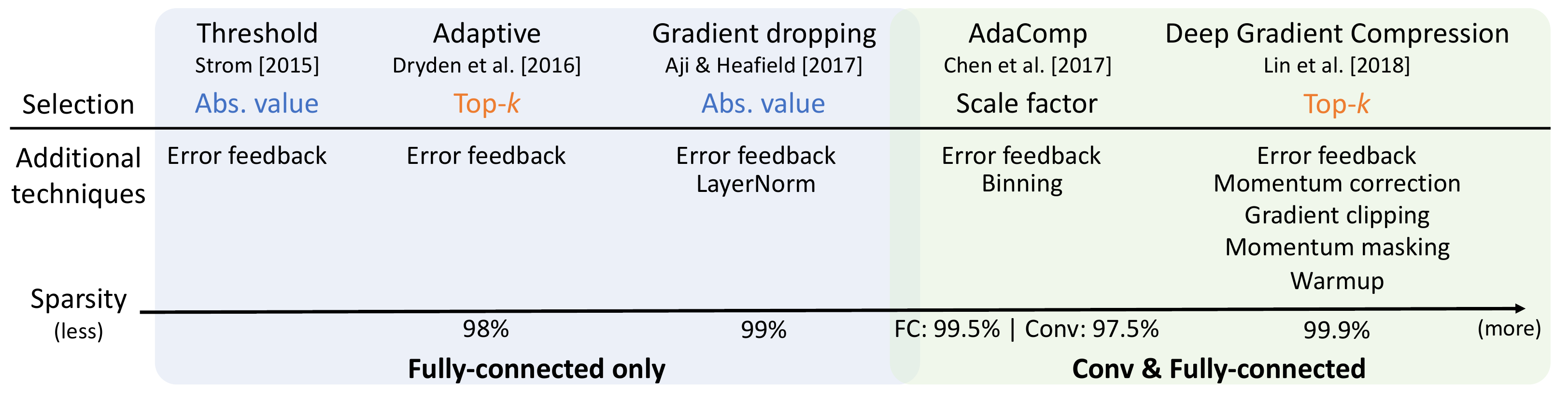}
	\caption{Overview methods for magnitude-based gradient sparsification.}
	\label{fig:gradient_overview}
\end{figure}

\subsubsection{Magnitude-based gradient sparsification}
\parad{It's mostly magnitude-based pruning and tricks.}
Most methods for gradient sparsification select gradients to remove based on magnitude, on the assumption that smaller gradients are relatively less important.
The first work on gradient sparsification, \citet{2015-strom}, is prototypical.
A fixed threshold $\tau$ is introduced as a hyperparameter and only gradient components of absolute magnitude larger $\tau$ are applied directly to the model.
The remaining values are quantized to a single bit per component based on their sign, and each is packed into a single 32-bit integer representing the index and quantized value.
The other key feature is \emph{error feedback}~\cite{2014-seide}, where each worker locally accumulates the error introduced by its compression and incorporates the residual into the next iteration, by simply adding it to the gradient.
Using this method, \citet{2015-strom} showed that communication bandwidth was reduced by three orders of magnitude for training a DNN for acoustic modeling, with no reduction in accuracy.

Absolute cut-off magnitudes are hard to pick because different networks or layers within a network may have gradients of different magnitudes, and the magnitude may change during training.
\citet{2016-dryden} use a form of top-$k$ selection, whereby a fixed proportion of the positive and negative gradients are retained.
They use sampling to find an absolute threshold for top-$k$ selection in linear time.
They also quantize those top-$k$ gradients to a single bit~\cite{2014-seide}, compress them based on entropy, and utilize all rounding errors through error feedback.

Subsequent works improved upon these by refining the methods for selecting gradients or incorporating other tricks.
\citet{2017-aji} use a single proportion for all gradients, and select it globally for all layers, finding that layer normalization~\cite{2016-ba-layernorm} is sufficient to keep gradients on a similar scale.
\citet{2017-sun} performs top-$k$ sparsification of gradients as part of sparsifying all computation in backpropagation (see Section~\ref{subsec:ephem-errs}).
\citet{2018-chen} study sparsification for CNNs in addition to fully-connected networks.
They formalize \emph{binning}, where compression is applied separately to subsets of a layer's gradients.
This ensures that sampling windows are small enough to effectively capture different gradient dynamics within a single layer.
They also use a self-adjusting threshold based on a scale factor, rather than a fixed top-$k$ threshold.
\citet{2018-lin} introduces a number of tricks to improve the convergence of top-$k$ sparsification, including incorporating momentum into error feedback, gradient clipping, stopping momentum on excluded gradients, and a warmup phase with less sparsity. 
This can result in orders-of-magnitude communication-compression; however, their results appear to be quite sensitive to hyper-parameterization. 
\citet{2019-sun} further approximate the gradient momentum and incorporate local update steps.

Fig.~\ref{fig:gradient_overview} provides an overview of these methods, their key components, and the sparsity they are able to achieve (we omit the sparsity for \citet{2015-strom}, as they focus on very different applications and the sparsity results are not comparable).
Gradient sparsification has steadily improved in the amount of sparsity it can introduce, with \citet{2018-lin} achieving up to 99.9\% sparsity.
Compared to pruning for weights or activations, gradients seem to be significantly more amenable to sparsity.

\subsubsection{Variance-based gradient sparsification}
\parad{Variance-based sparsification.}
The convergence of SGD is significantly impacted by the variance of the stochastic gradients used.
However, sparsification can increase the variance in the resulting sparse gradients, and hence slow convergence.
\citet{alistarh2017qsgd} noticed that, when stochastically quantizing gradient vectors normalized by their $L_2$-norm to only three quantization levels: 0, 1, and -1, in expectation all but a $\Theta(\sqrt n)$ fraction of the $n$ gradient values will be set to zero. This results in non-trivial compression, but also induces high additional variance, which hurts convergence. 
To alleviate this issue, \citet{2018-wangni} first propose rand-$k$ sparsification, where $k$ gradients are retained at random, biased by their absolute value, and the rest zeroed; the remaining gradients are then rescaled to ensure the gradient is unbiased. They then develop algorithms to select the optimal sparsification strategy given a variance budget. In practice this turns out to be similar to choosing an appropriate $k$ for top-$k$ sparsification.
Similarly, \citet{2018-wang} considers the problem of minimizing variance subject to a sparsity budget.
They also consider the more general problem of sparsifying arbitrary atomic decompositions, rather than just element-wise sparsification.
Concurrently, \citet{2018-tsuzaku} also identify variance as a key metric, and use the variance of gradients within a mini-batch, rather than their magnitude, as a criterion for sparsification.
The variance can be computed for relatively little extra cost during standard backpropagation.
Using variance as a sparsification metric thus has attractive theoretical properties, and \citet{2018-tsuzaku} show that it matches or outperforms \citet{2015-strom}'s threshold sparsification on CIFAR-10 and ImageNet.

\subsubsection{Other methods for gradient sparsification}
A variety of other approaches to sparsification have also been studied.
\citet{2019-ivkin} use count sketches on each worker to approximate large gradients, and the sketches are communicated.
\citet{2019-lim} combine sparsification with ternary quantization, and use a tunable sparsity factor to control how many values are rounded to zero.
\citet{2020-basu} studies the convergence of the combination of sparsity, quantization, and local updates, showing this converges at the same rate as SGD in certain settings.
\citet{2020-wang} apply top-$k$ sparsification in the frequency domain after applying an FFT to gradients. 

\subsubsection{Convergence of sparsified gradient methods}
\parad{Theoretical convergence results.}
There have been several theoretical analyses of the convergence of sparsified gradient methods.
Concurrently, \citet{2018-stich, 2018-alistarh, 2018-jiang} show that sparsified gradient methods converge at roughly the same rate as standard SGD, provided error feedback is used. 
These works differ in terms of the assumptions made and guarantees provided: for instance, \citet{2018-stich} consider the case where a single node compresses its gradient via sparsification with error correction (``memory''), assuming a convex objective function, and provides very strong convergence guarantees, similar to those of regular SGD.
By way of comparison,~\citet{2018-alistarh} consider non-convex objectives, and the multi-node case, but require an additional analytic assumption for their convergence bounds. 
Overall, these works provide a strong theoretical justification to the previous empirical results, in particular highlighting the importance of error feedback for convergence. 
\citet{2019-karimireddy} extend these results to more general settings.
However, these results are for sparsifying an entire model's gradients, as opposed to layer-wise operation.
\citet{2020-dutta} further extend these convergence results and show that layer-wise compression is theoretically better.
They also experiments and show that, while this usually holds in practice, there do exist cases in practice where sparsifying an entire model out-performs layer-wise sparsification.
\citet{2019-tang} provide a convergence analysis for the case where, in addition to workers sparsifiying their individual gradients before communication, the aggregated gradient is also sparsified before being communicated back to the workers.
This situation is common in practice, but was neglected in previous analyses.

\subsubsection{Runtime support for sparse gradient summation}
\parad{Sparse allreduces.}
Sparse communication was first implemented in the parameter server setting, where all workers communicate with a single central parameter server.
However, many scalable high-performance distributed training systems perform communication without a central store using allreduces.
Extending sparse communication to this case is challenging.
\citet{2016-dryden} implements a ring-based allreduce that includes custom reduction operators that uncompress vectors, sum them, and recompress them with the same hyperparameters.
\citet{2019-shi-gtopk} proposes a similar mechanism, global top-$k$, where instead of using the top $k$ gradients from each worker, only the top $k$ gradients among all workers are used.
\citet{2019-shi} provides convergence results for this approach.
\citet{2019-renggli} propose SparCML, a framework for efficiently performing distributed aggregations of sparse vectors.
They combine sparse vectors and retain all non-sparse coordinates; as this may eventually result in dense vectors, SparCML includes a mechanism to switch from sparse to dense or even dense-quantized representations.

\subsubsection{Gradient sparsification for better accuracy}
The prior approaches have primarily focused on sparsification in order to reduce communication bandwidth.
\citet{2015-shokri} investigate such methods in the context of privacy, while \citet{2020-sinha} study top-$k$ sparsification to improve the training quality for GANs~\cite{2014-goodfellow}.
When training a GAN, a critic network is used to identify whether samples produced by the generator are ``bad''.
This work uses the critic to select the best $k$ samples in each mini-batch to perform updates with.

\parad{Outside the DL field.}
Note that the core idea behind parallelizing mini-batch SGD consists essentially of computing an average of the gradients of the samples within a mini-batch, which functions as a lower-variance estimate of the full gradient.
Computing this average is a special case of the more general distributed mean estimation problem.
Several works have tried to achieve optimal communication bounds for this and related problems~\cite{2018-konecny, 2017-suresh, 2019-huang, 2020-davies}.
We note however that the above gradient sparsification approaches do not solve exact distributed mean estimation, since the approximation to the true mean is inherently lower-dimensional; instead, they use error feedback to correct for the inherent error.

\subsection{Errors and optimizer state}
\label{subsec:ephem-errs}

In addition to the gradients of parameters, the gradients of a layer's input, or the ``errors'', can also be sparsified.
\citet{2017-sun} introduces meProp (``minimal effort backpropagation'') which applies top-$k$ sparsification to the errors to reduce flops.
This also necessarily leads to sparse gradient updates, as only $k$ rows (or columns) of the resulting gradient matrix are non-zero.
The top-$k$ sparsification is first applied to the gradient of the loss initially computed in backpropagation, and then reapplied after every fully-connected layer to keep the errors sparse. \citet{wei2017minimal} demonstrate that this scheme can lead to 95\% gradient sparsity. 

Whether the optimizer state can be sparsified and the benefits of sparse optimizer states have yet to be explored. We expect it to lead to more memory efficient training algorithms.

\subsection{Dynamic networks with conditional computation}

Dynamic networks where outputs of previous layers determine a path through the network increase model capacity without increasing the computational cost. Conditional computation achieves this by \emph{routing} the computation through the network without touching all weights. Many practical approaches use various trained gating techniques (binary or continuous, deterministic or stochastic)~\cite{bengio2013estimating,bengio2016conditional,almahairi2016dynamic} or use switching methods that explicitly select the next ``expert''~\cite{2017-shazeer,jacobs1991adaptive,jordan1994hierarchical}. 
Both approaches lead to ephemeral sparsity during the execution. 

Recently, mixture of experts models have achieved impressive success in natural language processing. \citet{2017-shazeer} define a Mixture of Experts (MoE) layer to contain $n$ expert subnetworks $E_1, \ldots, E_n$ and a gating network $G$ that outputs a sparse $n$ dimensional vector. The function of this layer can be written as $y = \sum_{i=1}^n G(x)_i E_i(x)$, where $G(x)$ selects (gates) the relevant experts. 
One way to implement a $k$-sparse gating function is to use a top-$k$ method. \citet{2017-shazeer} use a noisy top-$k$ gating where they add tunable Gaussian noise to the selection function to improve load balancing the experts. A typical basic gating function is $G(x) = \mathrm{softmax}(W_g x)$ with learned weights $W_g$.
\citet{lepikhin2020gshard} apply this idea to transformer networks to train a model with 600 billion parameters by using a similar gating function for $k=2$ and stochastic load balancing across the experts to enable large-scale parallel training. 
Switch transformers~\cite{2021-fedus} evolve the model further and show that MoE sparsity can improve pretraining speed by up to 7x compared to a dense model and supports models with extreme capacity of up to a trillion parameters. They show that $k=1$ (a single expert) performs best and they design a load balancing loss term for the gating function. 

\parad{runtime neural pruning}
Conditional computation during inference requires quick decision making at low overhead. 
Runtime Neural Pruning~\cite{2017-lin}, uses a Markov decision process to determine the path through the network. Its parameters learned by a reinforcement learner, the path through the network is determined at inference time. Here specifically, the agent determines which channels are important to be considered for a specific input. During training, two networks are trained in tandem: the original (``backbone'') convolutional network and the decision network that guides filter selection at runtime. 
\citet{2020-chen-rl} show a reinforcement learner used at runtime to select convolutional channels during runtime with low storage. Several similar approaches use gating modules~\cite{liu2018dynamic} or routing~\cite{rosenbaum2017routing}.

Other similar approaches, such as product key networks~\cite{2019-lample} increase model capacity without increasing the computation using key-value store-like memory layers. This vast topic of dynamic memory networks is outside the scope of this overview.

\section{Sparse deep learning architectures}\label{sec:architectures}

After describing the building blocks of sparsification methods, we continue to highlight specific applications and results that were achieved applying these methods. Many works prune for a specific goal such as performance/inference latency~\cite{2018-gordon}, memory consumption~\cite{2020-li}, or energy efficiency~\cite{2017-yang}. There, several sparsifying techniques are often combined, for example, regularization and magnitude pruning~\cite{2017-yang,2017-he}. Layer-wise sensitivity schemes or data-free methods can then be used to improve the performance further. Many schemes iterate over a mix of such techniques and their carefully engineered combinations with pruning schedules can result in impressive gains for specific purposes~\cite{2015-han,2017-yang}.

Each methodology represents a combination of specific elements to sparsify, a sparsification schedule, a removal method, and (optionally), a re-addition method. 
Each result is measured by the authors of the original work and can be reproduced through the description in the original paper. 
As pointed out by \citet{2020-blalock}, these works do not always follow a consistent experimental discipline and thus many results are unclear, and may not be fully interpretable from both an accuracy and performance perspective~\cite{benchmarking}. 
Thus, when comparing works, we rely on the author's results and only do so to provide a rough overview of the relative performance of these methods. 
Different setups and problem statements are likely to shift this balance---however, we believe that we can derive several important observation from the quantitative results.

Here, we focus on more recent results after 2015 when broader interest in deep neural networks emerged and works solved large-scale data-driven problems from computer vision, natural language processing, and related domains, that are still relevant today.

\subsection{Sparsifying convolutional neural networks}\label{sec:convarch}

\parad{mini-intro}
CNNs with diverse structures have recently become the primary target for sparsification, and diverse architectures were successfully pruned. As opposed to MLPs, CNNs contain combinations of convolutional operators (Section \ref{sec:conv}), fully connected layers, skip connections, and other statistical normalization operators such as batch normalization. The composition of these operators determine which sparsification strategies would be effective. Convolutions are used from the inputs onwards to compute feature maps, whereas the fully connected layers are used as classifiers. The convolutional operators can be pruned in a structured or unstructured manner, but typically less than fully connected layers, due to the fewer and structured connections between the inputs and the output.

\begin{figure}[h!]
	\centering
	\begin{subfigure}{.45\textwidth}
		\centering
		\includegraphics[width=\linewidth]{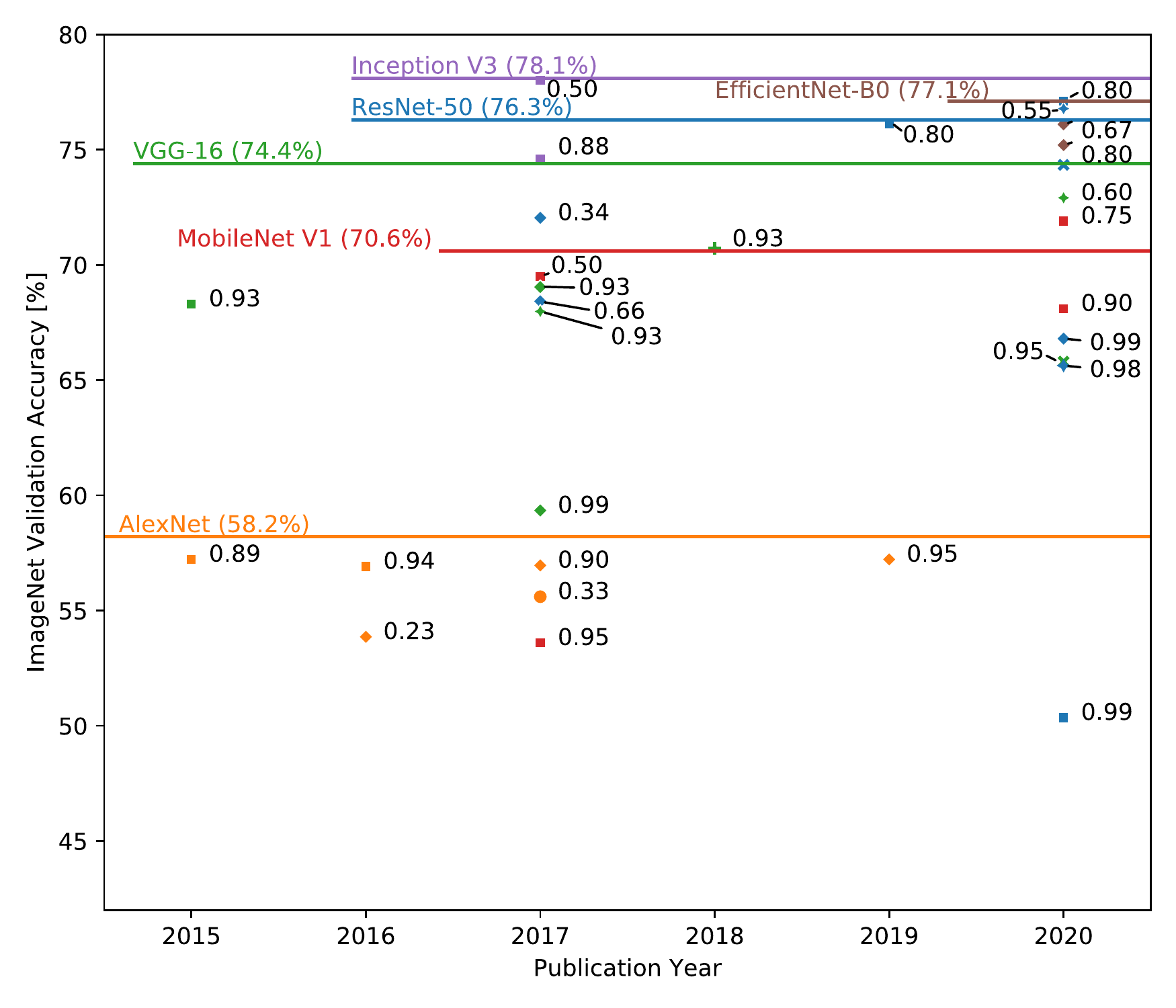}
		\caption{Best accuracy and sparsity over time}
		\label{fig:cnn_acc_over_time}
	\end{subfigure}
	\qquad
	\begin{subfigure}{.45\textwidth}
		\centering
		\includegraphics[width=\linewidth]{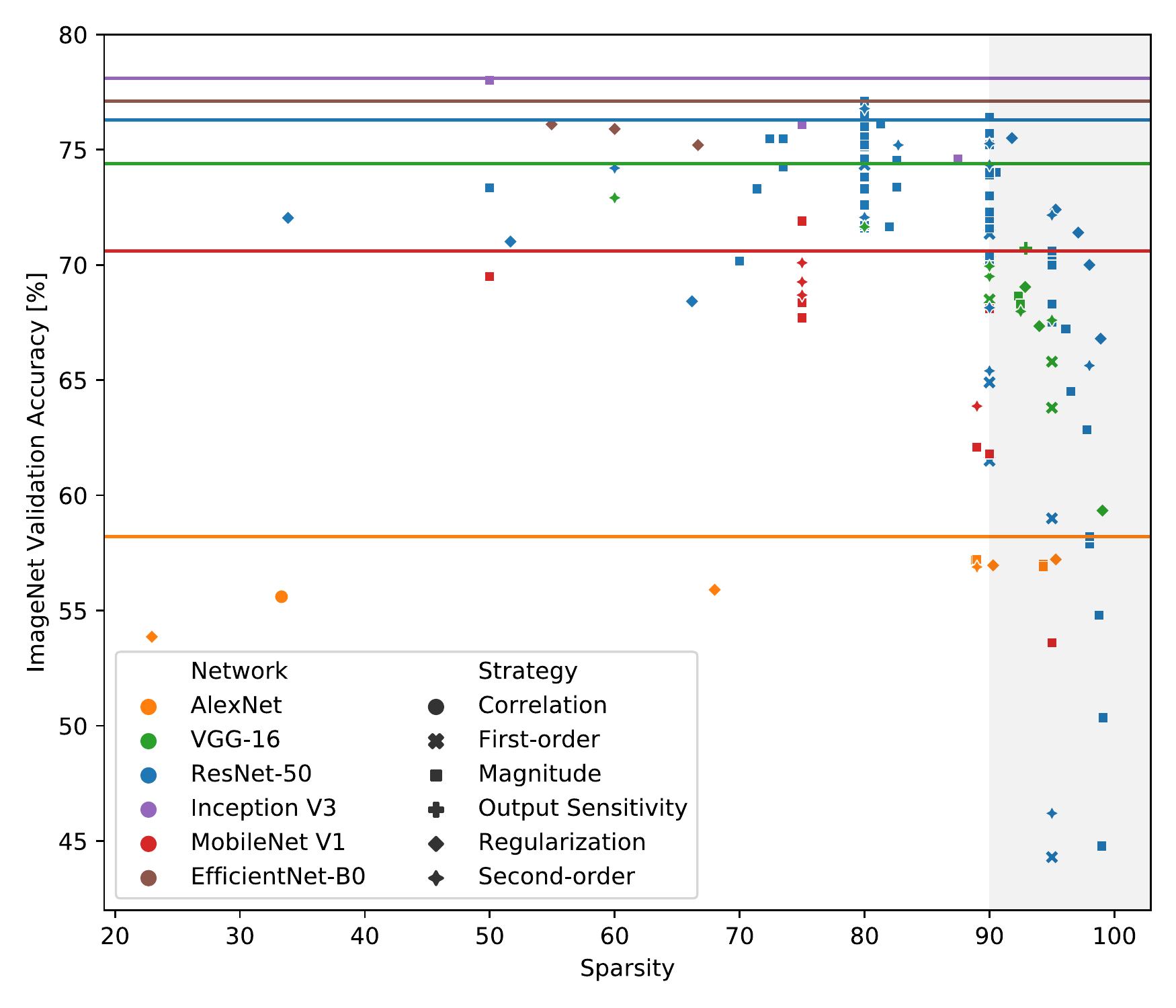}
		\caption{Sparsity vs. accuracy}
		\label{fig:acc_over_sparsity}
	\end{subfigure}
	\caption{Accuracy of pruned CNNs. Marker shape indicates pruning strategy and labels indicate sparsity.}
	\label{fig:cnn_accuracy}
\end{figure}

In Fig.~\ref{fig:cnn_accuracy} we see the development of accuracy in CNNs over time (Fig.~\ref{fig:cnn_acc_over_time}) and sparsity (Fig.~\ref{fig:acc_over_sparsity}), where in the former we highlight the two extremes on the Pareto front of sparsity --- best validation accuracy and highest compression ratio --- for every year and every strategy. 
We see that over the years research was able to increase compression and accuracy at the same time, and the composition of pruning strategies (see Section \ref{sec:removal} for details) changed, but magnitude constitutes the majority of the reviewed works. 
From Fig.~\ref{fig:acc_over_sparsity}, we see that two regions emerge: dense to moderate sparsity (0--90\%), and moderate to high sparsity (marked in darker background). In the lower compression ratios, magnitude-based pruning works relatively well (especially when iterative pruning is applied), achieving state of the art accuracy for all studied networks. However, when >90\% sparsity is desired, regularization, first-order, and second-order sparsification yield the best networks in the sparsity-accuracy tradeoff.
Below we review the history and methodology behind the papers shown in the figures. First, we focus on the convolutional operator and modifications to the CNN architecture. Then we discuss approaches for pruning CNNs and the derived training schemes.

\subsubsection{CNN architecture evolution}
\parad{modifications to cnn architectures}
Over-parameterization in convolutional operators was already noted by \citet{2014-szegedy}. In order to reduce the computational requirements and memory footprint of CNNs, the authors proposed the GoogLeNet architecture, using ``Inception'' modules that trade large convolution kernels with 1$\times$1 convolutions and smaller convolutions following dimensionality reduction. This was later improved to chaining separable 1D convolutions instead of 2D in the ``Inception V3'' CNN~\cite{2016-szegedy}, and with depth-wise separable convolution~\cite{sifre2014rigid} in the parameter-efficient MobileNets~\cite{howard2017mobilenets}, both of which can be seen as handmade sparse formulations of convolutions.
\citet{2019-kuzmin} provides a survey about structured compression of convolutions, including tensor decompositions, channel pruning, and probabilistic compression.
A recent popular technique to reduce the size of CNNs and increase their parameter efficiency is Neural Architecture Search (NAS)~\cite{2019-tan}, formulating the process as a meta-optimization problem. EfficientNet~\cite{2020-tan}, the current state-of-the-art CNN for image classification, uses NAS to construct their base EfficientNet-B0 network, and defines a compound method to scale it up while retaining parameter efficiency.

\subsubsection{CNN sparsification}
\parad{magnitude}
In unstructured pruning, the popular paper on model compression by \citet{2015-han} combines magnitude-based sparsification, quantization, weight sharing, and Huffman coding into a compression scheme able to reduce AlexNet and VGG-16 on the ImageNet dataset by 35$\times$ and 49$\times$, respectively, without loss of accuracy. They were able to sparsify those models by more than 90\% when manually tuning the sparsification level per layer. The authors show that convolutional layers should be sparsified less (15--65\%) than fully connected layers (91--96\%). Their compression scheme starts from a fully-trained baseline model, performing re-training to reach the original accuracy. 

\parad{initialization and training scheme}
Using a scheme based on neuron output correlation, \citet{2015-sun} demonstrate 33\% improved accuracy for the DeepID2+ face recognition CNN when sparsified by 74\%, and retaining the same accuracy with sparsification of up to 88\%. The authors also discuss the sparsification re-training scheme, showing that a fully-sparse training approach could not match the performance of the dense-trained, then sparsified network. They conjecture that, with a sparser model, the randomized initial values of the weights play a more significant role than in a dense model. Thus, training a sparse model is more prone to converge to suboptimal local minima than a dense network, 
agreeing with later proposed theory~\cite{2019-frankle}.

Some works advocate for training the sparse network in tandem with the dense network, or by modifying the training process to promote sparsity. One example is the effect of dropout on sparse networks (see Section \ref{sec:dropout}). 
\citet{2016-zhou} propose a forward-backward splitting method to enforce sparsity as regularization, pruning 61.3\% of VGG-13 and 65.4\% of AlexNet parameters with 1.7\% and 0.53\% accuracy degradation respectively.
\citet{2018-tartaglione} use sensitivity-driven pruning until the network drops below the required accuracy. They achieve higher sparsity than earlier magnitude-based mechanisms. 
\citet{2017-molchanov} use variational dropout (see Section~\ref{sec:varbayes}) to prune weights starting from relatively small pre-trained networks. For those networks, they show record sparsity levels for small networks: 98.5\% for LeNet-300-100 and 99.996\% for LeNet-5 with 98.1\% and 99.3\% accuracy on MNIST, respectively. Training takes twice the number of operations for forward and backward but converges equally fast on LeNet and MNIST. VGG-style networks on CIFAR-10 and CIFAR-100 could be sparsified by more than 97\% at similar accuracy.

\parad{Retraining schemes}
\citet{2017-dong} use 2nd order OBS pruning per layer to limit its computational complexity. They approximate the inverse of the Hessian matrix with the Woodbury matrix identity. The achieved compression ratios are similar or slightly better than magnitude-based pruning. However, they show that after applying 2nd order pruning, the resulting network (before retraining) has a much higher quality than the ones obtained with magnitude pruning ($<$5\% vs. $>$80\% error for LeNet and $<$50\% vs. $>$73\% error for VGG and AlexNet). This reduces the number of iterations needed to re-train the model to near-original accuracy ($>$200$\times$ less for LeNet, $>$40$\times$ less for AlexNet, and $>$12$\times$ less for VGG-16). 

\citet{2016-guo} observe that the process of sparsification can benefit from re-adding weights during training that were erroneously pruned before. 
For this, they maintain the set of all weights during the whole training process, including the pruned ones, and mask them during the forward pass. This allows them to later re-add pruned weights if they reach a certain magnitude. 
Furthermore, they specify a pruning schedule to decrease the sparsification probability over time. They demonstrate that this method significantly improves upon earlier methods~\cite{2015-han} that use iterative retraining --- specifically, they show that the number of iterations to prune AlexNet can be reduced from 4.8M to 0.7M (6.9$\times$) while improving the sparsity from 89\% to 94\% (2$\times$). Using the same method, they compress LeNet-5 and LeNet-300-100 by 99\% and 98\%, respectively. They again show that convolutional layers that already share weights compress less (46-97\%) than fully-connected layers (93-99\%).
%

\parad{fully-sparse retraining schemes}
Despite prior claims against fully-sparse training schemes, and due to growing CNN memory footprints, several recent works attempt to improve such schemes to produce usable networks.
\citet{2018-bellec} use a fully-sparse training schedule to enable training higher-dimensional models that would not fit in a dense configuration. They use a variant of magnitude based pruning and random weight addition and show that this method outperforms densely-trained methods if the target sparsity is very high ($>$95\%). They showed that longer training leads to improving generalization. 
The paper also studies aspects of transfer learning and pre-training, in that the sparsified architecture quickly adapts to similar learning tasks. \citet{2018-mocanu} use a similar training schedule and show that it can improve accuracy while pruning by more than 96\%. They also show that the degree distribution of sparsely learned connections follows a power law.
\citet{2019-mostafa} refines fully-sparse training for CNNs by automatically adjusting the parameter budget across layers. Their method may require more operations to converge than hand-tuned schedules, as sparsity may only slowly be redistributed to the later fully-connected layers. Their sparsely-trained models achieved significantly better performance than dense models of the same size.
\citet{2019-dettmers} perform fully-sparse training and point out that parameter redistribution is especially important for larger layers. They use a cosine decay schedule for the pruning rate across iterations and achieve similar performance with a 95\% sparse VGG on CIFAR-10, and slightly outperform 
prior approaches with a 90\% sparse ResNet-50 on ImageNet, achieving 72.3\% accuracy, while reducing the required computations between 2.7$\times$ and 5.6$\times$.

\parad{cnn sota}
%
In more recent, parameter-efficient networks, state-of-the-art pruning techniques become more adaptive to the  training process. \citet{2020-azarian} propose soft pruning, where sparsifying a weight is a continuously differentiable function, and $L_0$ regularization. The authors prune ResNet and EfficientNet-B0, where the latter attains 76.1\% accuracy, compared with a 77.1\% accuracy for the dense counterpart. 
\citet{2019-he} use a reinforcement learning approach combined with a CNN embedding scheme to prune the network. Their first-order approach to sparsification is able to sparsify ResNet-50 to 80\%, keeping the same baseline top-1 validation accuracy of 76.1\%.
\citet{2020-evci} train networks fully-sparse with pruning based on magnitude and re-addition based on instantaneous gradients. They decay the sparsification probability using a cosine schedule and stop sparsification before the end of training. The method attains good generalization for ResNet-50, with 76.6\% at 80\% sparsity and 75.7\% accuracy at 90\% sparsity, while reducing the computations compared with dense training. 
\citet{2020-singh} use second-order information (specifically, inverse Hessian-vector products using an approximation based on the empirical Fisher Information Matrix) to estimate which weights to prune. The authors report that with gradual sparsification, ResNet-50 can be pruned with no extra epochs to higher accuracies than existing approaches that do the same (76.8\% at 80\% sparsity).
\citet{2019-gale} provide a systematic study of various pruning strategies: random (baseline), magnitude pruning, $L_0$~\cite{2018-louizos}, and variational Bayes~\cite{2017-molchanov} applied to ResNet-50 and transformer networks, ranging from 50--98\% sparsity. Their main result is that simple magnitude pruning is competitive, if the pruning schedule and per-layer distribution is tuned (76.52\% accuracy for 80\% sparse ResNet-50 and 75.2\% for 90\% sparsity after 100 epochs). 
They report that variational dropout performs best only for very high sparsity ($>$95\%), but tuned magnitude pruning remains close. 
They also show that both variational dropout and $L_0$-based pruning can be up to 3$\times$ slower and uses 2$\times$ more memory than magnitude pruning. Their general conclusion is that well-tuned magnitude pruning is probably the most practical pruning method. 

\begin{figure}[h!]
	\centering
	\includegraphics[width=.6\linewidth]{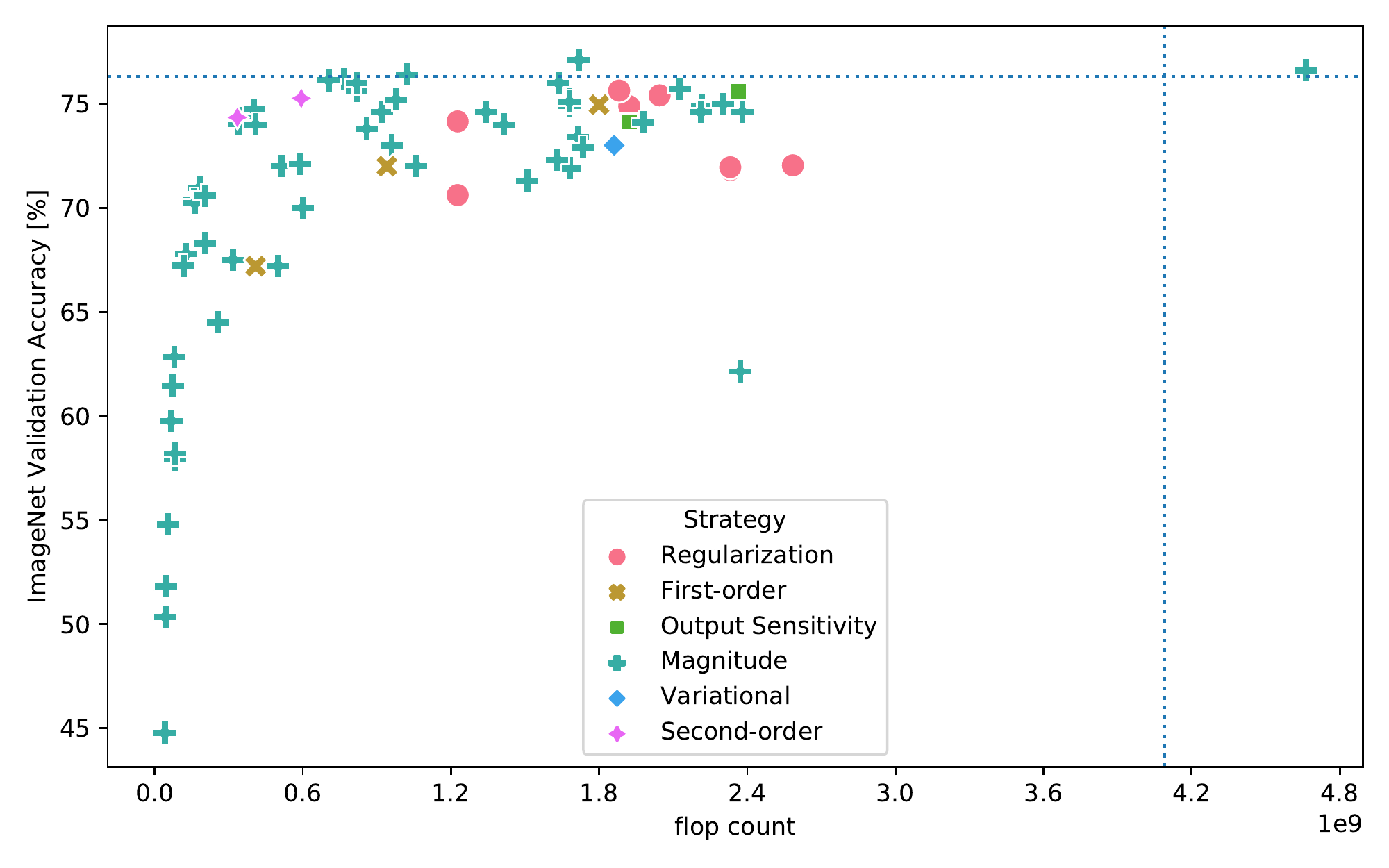}
	\caption{Flop count and resulting accuracy of state-of-the-art pruning methods for ResNet-50 over ImageNet. Dotted lines represent dense baselines.}
	\label{fig:flop}
\end{figure}

Fig.~\ref{fig:flop} shows an overview of the computational intensity (flop count) for inference of sparsified ReNet-50 models. It shows that one can save between 50-80\% of the operations without significant loss in accuracy, leading to a potential speedup of up to 5x. 

\subsection{Sparsifying transformer networks}\label{sec:tformers}

\begin{figure}[h!]
	\centering
	\begin{subfigure}{0.45\textwidth}
		\centering
		\includegraphics[height=5cm]{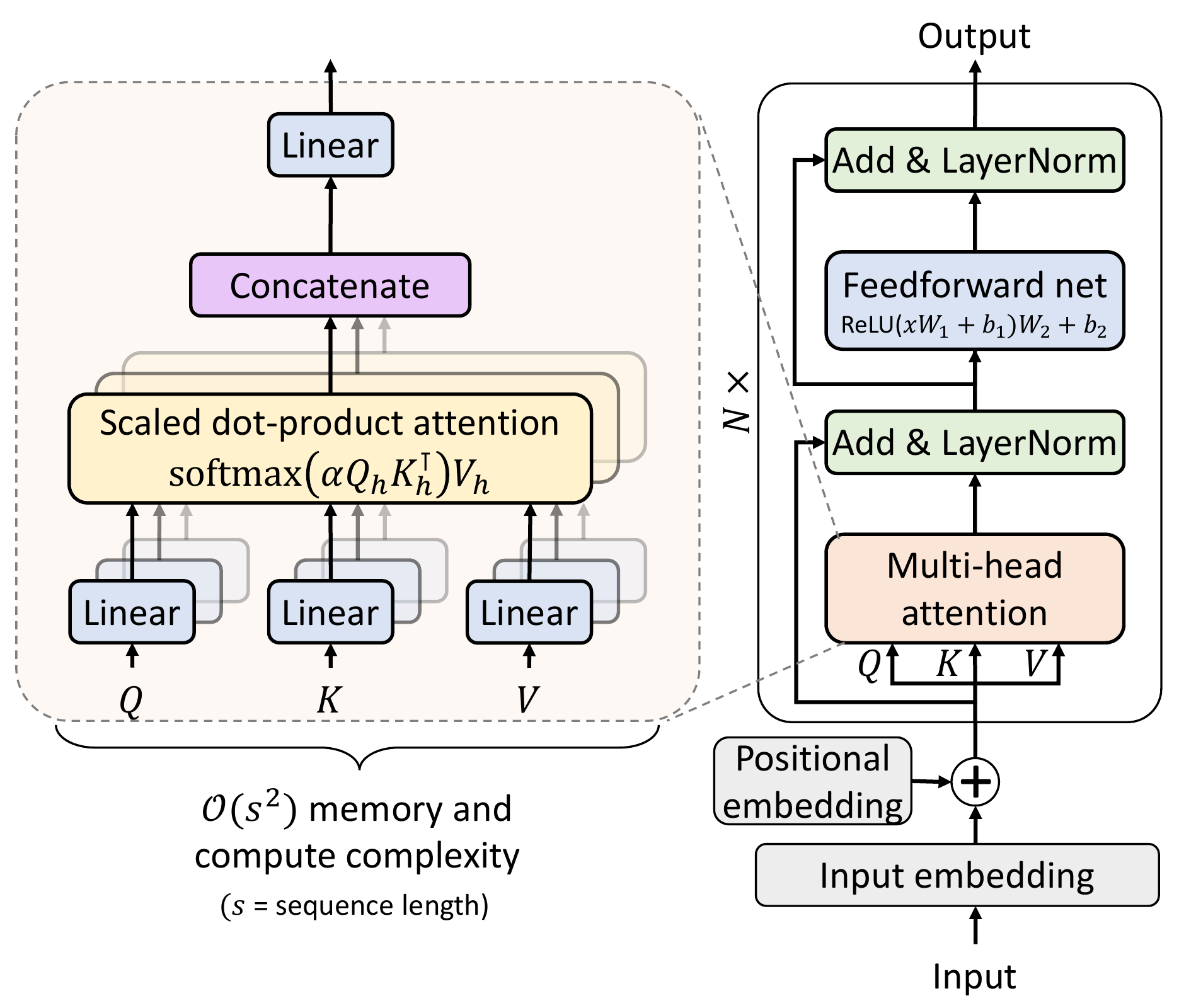}
		\caption{Transformer architecture~\cite{vaswani2017attention}.}
		\label{fig:transformer_arch}
	\end{subfigure}
	\begin{subfigure}{0.5\textwidth}
		\centering
		\includegraphics[height=5cm]{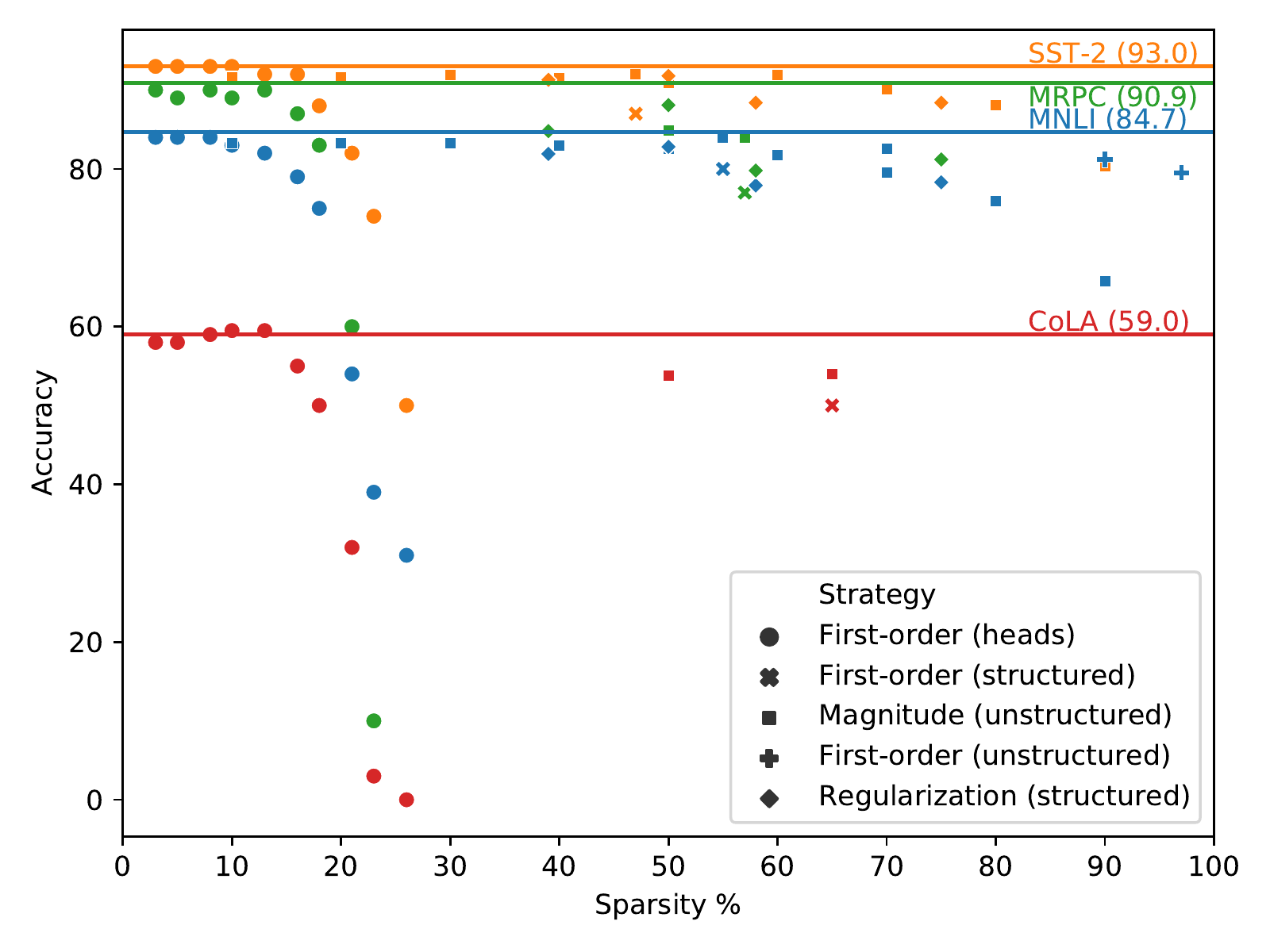}
		\caption{Accuracy of pruned BERT-base on selected tasks.}
		\label{fig:transformer_accuracy}
	\end{subfigure}
	\caption{Overview of transformers.}
	\label{fig:transformer_overview}
\end{figure}

\parad{Intro: Why care and what they are.}
Transformers~\cite{vaswani2017attention} are a class of sequence transduction models that have led to breakthroughs in natural language processing and are recently expanding into other fields such as computer vision~\cite{2020-dosovitskiy}.
Widely used transformer models include the original Transformer~\cite{vaswani2017attention} for language translation as well as language models such as BERT~\cite{2019-devlin} and GPT-3~\cite{gpt-3}.
The key idea behind transformers is to generalize prior work on shallow language embeddings to deep, multi-layer embeddings, while being more parallelizable in training than RNNs.
Like CNNs, transformer architectures are a combination of a variety of operators, which we illustrate in Fig.~\ref{fig:transformer_arch}.
The key primitive is multi-head attention, introduced by~\citet{vaswani2017attention}.
Each attention ``head'' performs scaled dot-product attention to identify how elements of one sequence should relate to elements of another sequence.
A transformer layer is then composed of a multi-head attention layer followed by a feedforward network (sometimes called ``expert layers''), with layer normalization~\cite{2016-ba-layernorm} and residual connections.
The full transformer network consists of one or more stacks of transformer layers, with embedding layers at the beginning.

\parad{Big models make good targets for compression.}
Transformers are often very large, ranging from about 110 M parameters in BERT-base to hundreds of billions~\cite{gpt-3} or trillions~\cite{lepikhin2020gshard, 2021-fedus} in the largest, best-performing models.
As it is infeasible to deploy such large models in production situations, compression is critical.
Indeed, \citet{2020-li} show that training a large, over-parameterized transformer and then compressing it results in better accuracy than training a smaller model (e.g., a 75\% sparse 24-layer RoBERTa model~\cite{2019-liu} outperforms a 3-layer model on MNLI, while being the same size).
Complementary to pruning, many other approaches to compressing transformers have been developed; \citet{2020-ganesh} provides an overview for BERT specifically, and \citet{2020-gupta} for deep learning models for text in general.

\parad{Overview of sparsity figure.}
Fig.~\ref{fig:transformer_accuracy} presents an overview of sparsity results for pruning BERT-base~\cite{2019-devlin} for four downstream natural language understanding tasks from the General Language Understanding Evaluation (GLUE)~\cite{2019-wang-glue} benchmark: the Stanford Sentiment Treebank (SST-2)~\cite{2013-socher}, the Microsoft Research Paraphrase Corpus (MRPC)~\cite{2005-dolan}, the Multi-Genre Natural Language Inference corpus (MNLI)~\cite{2018-williams}, and the Corpus of Linguistic Acceptability (CoLA)~\cite{2019-warstadt}.
BERT-base consists of 110M parameters (including embedding layers), with twelve transformer layers, each with twelve attention heads.

Compared to results on pruning CNNs (Fig.~\ref{fig:cnn_accuracy}), there are two qualitative differences: There are relatively few results with large accuracy degradation, and there are relatively few results with very high sparsity levels.
This stems from much of the work on pruning BERT being focused either on understanding what the model has learned or on the Lottery Ticket Hypothesis (see Section~\ref{sec:lottery}).
In many of these works, iterative pruning only continues while the pruned model remains close to the original accuracy.

We can observe several qualitative trends among methods, which generally agree with the results on CNNs.
For very low sparsity levels, structured head pruning performs very well, but it rapidly degrades as important heads are pruned.
At moderate sparsity levels (40--80\%), unstructured magnitude pruning performs very well, and outperforms structured pruning.
When >90\% sparsity is desired, however, only the first-order movement pruning method~\cite{2020-sanh} reports results, and achieves high accuracy on MNLI.

\subsubsection{Structured sparsification}
\parad{Pruning heads}
There has been much study of the importance of different components of transformers; for BERT, this is referred to as ``BERTology''~\cite{2021-rogers}.
For example, while attention heads are important for training, several works showed that most of the heads can be pruned after training with only minor accuracy loss.
\citet{2019-michel} and \citet{2019-voita} study the importance of heads in two concurrent and complementary works.

\citet{2019-voita} analyze the linguistic properties and importance of each head and conclude that specific heads take on specific roles, such as representing ``positional'', ``syntactic'', and ``rare words'' functions.
Using a simple stochastic gating scheme~\cite{2018-louizos} to prune heads, they can remove 80\% of heads and lose only 0.15 BLEU on a English-Russian translation task~\cite{jan2019iwslt} and 92\% of heads at a loss of 0.25 BLEU on OpenSubtitles~\cite{lison2019open}.

\citet{2019-michel} show similar results with a first-order head importance score for pruning.
Using an iterative greedy process to test model quality with each head removed, they are able to prune 20--40\% of attention heads with an insignificant decrease in quality.
They also find that the importance of heads is transferable across tasks and that the importance of heads is determined early in the training process, hinting that early structure adaptation may also apply to heads.

\parad{Pruning heads plus slices of things.}
However, \emph{multi-head attention layers account for only about a third of the parameters in BERT}, which limits the overall compression level, and for some tasks, \citet{2019-michel} show that pruning too many heads is detrimental to accuracy.
\citet{2020-prasanna} extended the importance metric of \citet{2019-michel} to also prune entire feedforward networks in a layer, using a similar iterative pruning process that continues for as long as the model retains over 90\% of the original's accuracy.
With this, they show that BERT can be pruned to 40--65\% sparsity on a variety of GLUE benchmark tasks.
They also show that, for low sparsity, even random structured removal achieves good performance.

\citet{2019-mccarley} evaluate a larger set of pruning approaches that can remove attention heads and slices of feedforward and embedding layers and use a gating mechanism $\alpha_j \in \{0, 1\}$ to select components for removal.
They compare four techniques for pruning: (1) random pruning as a baseline; (2) a first-order ``gain'' metric that computes $g_i = |\partial L/ \partial \alpha_i|_{\alpha_i=0}$ for each example; (3) a leave-one-out score, where the loss for each element removed is computed separately, and elements that cause a small loss on removal are retained; and (4) a sampled $L_0$ regularization.
Finally, they apply distillation using the unpruned model as a teacher for the pruned model.
The main finding is that $L_0$ regularization performs best and can prune 40--75\% of the elements in BERT and RoBERTa models while loosing about 5 points F1 score on the Natural Question benchmark task~\cite{kwiatkowski2019natural}.

\citet{2020-wang-tformer} use a modified $L_0$ regularization to prune all weight matrices in a transformer.
For each weight matrix $W$, they first reparameterize it as a low-rank factorization $W = PQ$, and then introduce a diagonal pruning matrix $G$, so that $W = PGQ$.
The pruning matrix allows the model to learn to keep the best rank-1 components of the weight matrix.
They use $L_0$ regularization to promote sparsity, with an additional term added to allow for control of the desired sparsity level.

\parad{Pruning larger blocks.}
Building on the idea that layers in transformers learn disparate tasks~\cite{2021-rogers} and that some layers may be less important than others~\cite{2019-tenney}, two works have pruned larger-scale structures.
\citet{2020-lin-tformer} prune entire residual blocks (i.e., either the entire multi-head attention layer or feedforward network) by identifying blocks whose nonlinear part has small activations.
These blocks are then pruned and replaced by a simple identity map.
To do this, they adapt $\epsilon$-ResNets~\cite{2018-yu-epsresnets} and augment each residual block with a gating function if the non-linear component is less than $\epsilon$.
Once a layer's activations fall below $\epsilon$, it will cease to contribute to the output, and its gradients will no longer be updated, leading to weight collapse.
In a similar vein, \citet{2020-fan} introduce LayerDrop, a form of structured dropout that stochastically drops entire transformer layers during training.
To reduce model size for inference, they also explore different ways to completely remove layers, and find that the simple approach of dropping the layers at depth $d$ such that $d \equiv 0 \left(\textrm{mod} \left\lfloor \frac{1}{p} \right\rfloor\right)$, where $p$ is the dropout probability, performs best.

\subsubsection{Unstructured sparsification}
\parad{Magnitude pruning.}
Simple iterative magnitude pruning has also been applied for unstructured sparsification, with several conclusions.
\citet{2020-prasanna} compared it with structured pruning using \citet{2019-michel}'s first-order importance metric and found that unstructured magnitude pruning typically results in networks that are both smaller and retain better accuracy.
\citet{2020-gordon} showed that a pretrained BERT model can be pruned to up to 40\% sparsity without affecting the performance of downstream tasks, but beyond that performance begins to degrade.
Surprisingly, they also show that fine-tuning a pretrained model and then pruning it does not result in better performance, and conclude that one need only prune BERT once after pretraining instead of for each downstream task.
\citet{2020-chen} show a similar result in the context of the Lottery Ticket Hypothesis (see Section~\ref{sec:lottery}).
They find that magnitude pruning can prune a pretrained BERT model to up to 70\% sparsity without compromising performance on the pretraining objective, and that such networks transfer universally to downstream tasks.
In contrast, they find that while pruning for a particular downstream task may result in higher sparsity levels without compromising performance on that task, such networks do not transfer as well.

\parad{Proximal pruning.}
\citet{2019-guo} conduct experiments showing that using $L_1$ or $L_2$ regularization can cause divergence during training, and that the regularization should be decoupled from the gradient update, in line with prior work on optimization~\cite{2019-loshchilov}.
To prune, they instead develop Reweighted Proximal Pruning, which uses reweighted $L_1$ minimization instead of regularization, and use a proximal algorithm to find the sparsity pattern, rather than backpropagation.

\parad{First-order movement pruning.}
\citet{2020-sanh} argue that for transfer learning, what matters is not the magnitude of a parameter, but whether it is important for the downstream task.
They introduce movement pruning (see Section~\ref{sec:1storder}), a first-order method which prunes parameters that shrink during fine-tuning, regardless of their magnitude.
Movement pruning is able to achieve significantly higher performance than magnitude- or $L_0$-based pruning for very high levels of sparsity (e.g., 97\% sparse), and can be combined with distillation to further improve performance.

\subsubsection{Sparse attention}
\parad{Overview of attention.}
Scaled dot-product attention requires a dot-product between two sequences of length $N$ ($QK^\top$), which produces an alignment matrix for the two sequences.
Producing this matrix requires both $\mathcal{O}(N^2)$ time and memory; as sequence lengths in transformers range from 128 to 2,048, this can be a large bottleneck.
This, combined with the intuition that one does not need to compute full attention to get good model performance, has resulted in a large body of work on so-called efficient transformers.
\citet{2020-tay} provide a survey of this field; we focus here on sparsity.
Recent work has also started to develop benchmarks focused specifically on efficient transformers~\cite{2021-tay}.

\parad{Theory.}
\citet{2020-yun} provide broad theoretical results showing that $\mathcal{O}(n)$ connections in an attention layer is sufficient to universally approximate any sequence-to-sequence function if the following properties are met: (1) every token attends to itself; (2) a chain of connections covers all tokens; and (3) each token connects to all other tokens after a fixed number of transformer layers.
This provides a rigorous basis for the intuition that each input token need only be able to route to each other token through successive layers.

\parad{Attention patterns.}
Many approaches to sparse attention satisfy these requirements by sparsifying the $QK^\top$ computation, including restricting attention to local neighborhoods~\cite{2018-parmar},
star topologies~\cite{guo2019startransformer},
combinations of strided and fixed sparsity patterns~\cite{child2019generating},
sliding windows~\cite{beltagy2020longformer},
and local attention plus a fixed number of tokens that attend globally~\cite{2020-zaheer}. SAC~\cite{2020-li-sac} learns a \emph{task-specific} sparsity structure using an LSTM edge predictor.

\parad{Softmax sparsity.}
The SoftMax computation in each attention head can also be modified to maintain its ranking while inducing sparsity and satisfying the above requirements.
\citet{2019-zhao} take a direct route and apply top-$k$ sparsification to the attention weights.
The sparsity patterns can also be learned directly using generalizations of SoftMax, such as $\alpha$-entmax~\cite{2019-correia} or sparsegen-lin~\cite{2019-cui}.
These build on earlier work, predating transformers, that aimed to induce sparsity in attention mechanisms to either improve performance or interpretability, including sparsemax~\cite{2016-martins}, constrained sparsemax~\cite{2018-malaviya}, and fusedmax~\cite{2017-niculae}.

\section{Speeding up Sparse Models}\label{sec:speed}

Sparse networks do not always execute faster than dense networks using current machine learning frameworks on today's hardware.
\citet{2020-sanh} demonstrate that small dense models often perform faster on current hardware than sparse models of the same and even smaller size despite the generally higher accuracy and parameter efficiency of sparse models. \citet{2016-han-ese} show that even 90\% sparse workloads execute slower on a GPU than computing 90\% zeros densely and \citet{2017-yu} show that an 89\% sparse AlexNet executes 25\% slower on CPU than its dense version.
In general, unstructured sparsity is not well supported on today's architectures. 
Some cases of structured sparsity can be mapped to dense matrix operations (e.g., neuron, filter, or head sparsity) and can thus trivially use existing optimized frameworks or libraries such as cuDNN~\cite{2014-chetlur}.
Other structured sparsity approaches such as blocks of weights would require support from the frameworks to be executed efficiently.
We will discuss algorithmic and hardware solutions to support sparsity on practical systems in this section.

\parad{be careful with training}
Training for sparsity can be especially expensive on some architectures. For example, regularization methods (e.g., \citet{2018-louizos}), schemes using gating variables (e.g., \citet{1988-mozer}), and various other techniques~\cite{2016-molchanov,2020-sanh} double the number of trainable parameters. Furthermore, variational methods are often expensive in both memory and compute during training~\cite{2019-gale}.
Those techniques may be even slower than fully dense training in a sparsified training schedule. Thus, we recommend a careful analysis of memory, data movement, and computational overheads when considering the performance of a method (see \citet{ivanov2020data}). 

\subsection{Algorithmic and software support for sparse models}

Sparse computations have a long history in the context of linear algebra and scientific computing. However, sparsities in those fields are often two orders of magnitude higher than in today's deep learning ($>99.9\%$ vs. $50-99\%$~\cite{2020-gale}) and it was long considered not beneficial to attempt sparse computations on less than 99\% sparse matrices. 
Furthermore, many scientific computing workloads have close-to banded non-zero patterns that can often be compressed as hierarchical matrices. Those structures lead to high temporal locality and many libraries such as Intel's MKL are tuned for those patterns~\cite{2016-park}.
As we will see below, sparsified neural networks have different characteristics in their non-zero distributions. 
Thus, scientific computing kernels such as the sparse BLAS or cuSPARSE are only optimized for scientific computing workloads and supported formats aimed at high sparsities such as compressed sparse row. 
We do not cover the many elegant approaches developed for very high sparsity here---albeit they may become very relevant to sparse deep learning if the trend to higher sparsity continues. 
Instead, we focus on approaches developed for sparsity levels observed in today's deep learning workloads.

\parad{unstructured storage formats}
We describe basics of storing unstructured sparse matrices in Section~\ref{sec:perf}. Many practical schemes use run-length or delta-encoding with padding for offsets~\cite{2016-han-eie}. Furthermore, it is common to combine quantization with index storage to achieve aligned number formats. For example, \citet{2016-han-ese} pack a 12-bit integer value with a 4-bit index value into a 16-bit element that is naturally aligned to DRAM page and PCIe transaction boundaries. This format would store the sparse vector $v=[0,0,1,0,0,0,0,0,0,0,0,0,0,0,0,0,0,0,0,3,2,0,0,0]$ as $v'=[2|1,15|0,0|3,0|2]$, where, for example, the padding entry ``$15|0$'' decodes to $15$ (first 4 bits) zeros followed by the value $0$ (last 12 bits). Sparse weights are often stored column-wise for inference using a compressed sparse column (CSC) format to facilitate the sparse matrix-vector multiplication. 
\citet{2020-gale} show various techniques to tune such sparse computations to GPU architectures. 

\citet{2016-park} tune unstructured sparsity for convolutional layers by implementing an optimized sparse-with-dense matrix multiplication. They only consider sparsity in the convolutional kernels and not in the activations, which, despite of up to 85\% sparsity was slower than sparse-dense in their experiments. 
Using a simple but effective performance model, they guide the sparsification such that the resulting model achieves highest performance. While they demonstrate their approach in conjunction with dynamic network surgery~\cite{2016-guo}, it is applicable as a regularization or direct performance metric to many, if not most of the sparsification approaches discussed in Section~\ref{sec:removal}.
A general observation is that there is a range of sparsity where workloads can efficiently utilize CPUs: too low sparsity leads to high overheads managing it but also too high sparsity leads to a performance reduction on CPUs. This is due to the fact that higher sparsity increases the relative storage overhead of the index structure and decreases the relative compute load. Since CPUs have a fixed ratio of memory bandwidth per compute operation, too high sparsity will underutilize the compute units and be bound by the well-known data locality and movement bottlenecks~\cite{unat-locality,ivanov2020data}. This implies that an accelerator needs to be carefully tuned to the expected workload, making a detailed co-design between the data science aspects of sparsification and the engineering aspects of representation and dataflow mandatory.

\subsubsection{Structured sparsity}

\parad{sparse structures}
Various sparsity structures have been used in the deep learning context to manage the storage overhead. They vary from completely unstructured storage where the offset for each single element needs to be encoded to structured storage formats that only store offsets of blocks or other elements arranged with a fixed structure. 
In Section~\ref{sec:perf}, we analyze the storage complexity in terms of the number of parameters needed to describe the structure of an irregular matrix. A blocked format with block size $B$ would reduce the storage overhead by a factor of $B$. 
\begin{figure}[h!]
	\includegraphics[width=0.9\textwidth]{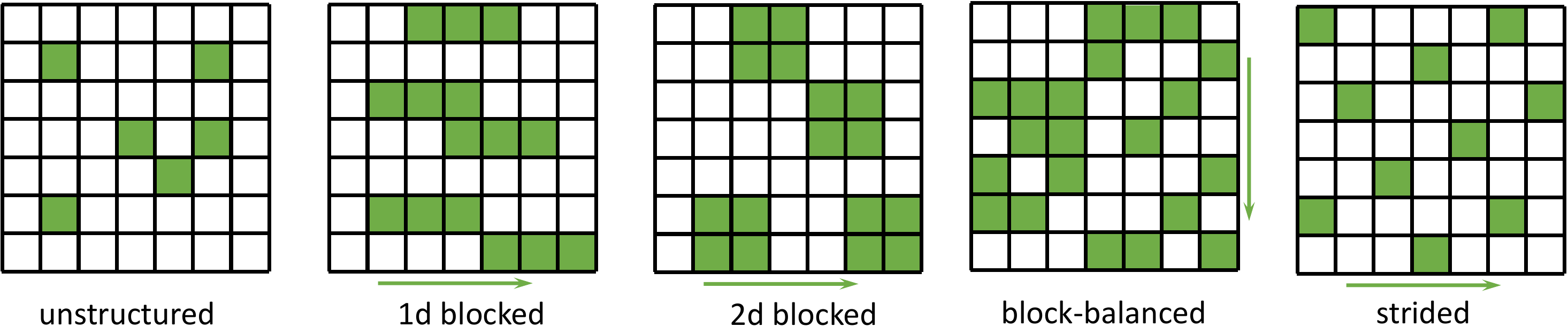}
	\caption{Overview of sparse structures in deep learning---non-zero elements are colored.}
	\label{fig:sparse_structures}
\end{figure} 
Fig.~\ref{fig:sparse_structures} shows an overview. Blocked formats can be defined for any set of dimensions, the figure shows a one-dimensional format with blocks of size three and a two dimensional format with blocks of size 4 ($2\times 2$) as used in Cambricon-S~\cite{2018-zhou}. Here, the offsets are only stored once for each block of non-zeros.
Another promising format is block-balanced. This format specifies a fixed block size and a fixed number of non-zeros per block. Here, one would only need to encode the offsets of the non-zeros for each fixed-size block. Nvidia's Ampere microarchitecture~\cite{a100} uses a bitmap to store the non-zeros in blocks of size four with 50\% sparsity. The figure above shows blocks of size seven with exactly three non-zero elements each. 
The strided format~\cite{2015-anwar} in Fig.~\ref{fig:sparse_structures} shows the most compact but also most constrained format. It fixes each 5th element of the matrix to have a non-zero value and all others zero, leading to a constant-size representation.
In general, sparse matrix storage formats can use arbitrary encodings to minimize the representational overhead. MPI datatypes~\cite{UsingAdvancedMPI} form an elegant hierarchy with clear performance properties~\cite{gropp-datatype-performance} and can provide further inspiration for specific designs. 

\subsubsection{Tuned block sparsity in practice}
\parad{blocked sparse inference}
\citet{2020-elsen} demonstrate speedups with sparse representation for inference on mobile devices. They focus on single-weight and weight-block sparsity and optimized implementations of CNNs on ARM CPUs and WebAssembly and release the XNNPACK library. They primarily focus on a medium sparsity range between 70--95\% and they optimize for caches and data movement, which has been shown to be a major bottleneck in deep learning systems~\cite{ivanov2020data}.
They investigate the influence of various block-sizes on model accuracy and show that the shape of blocks (e.g., $1 \times 4$, $2 \times 2$, or $4 \times 1$) is irrelevant and just the size matters. They also show that larger models suffer less from large block sizes. 

Scalpel~\cite{2017-yu} combines weight-blocking and neuron pruning into a scheme to support SIMD platforms. They sparsify weights in blocks the same size as the width of SIMD units and use a modified CSR format for storing the sparse weights. They prune by root mean square magnitude of weight blocks. For ARM microprocessors, their pruning scheme reduces the necessary sparsity required to achieve a speedup from 70\% to 50\% and on Intel CPUs, their scheme reduces the necessary sparsity from 50\% to less than 10\%. 
DeftNN~\cite{2017-hill} optimizes a whole row or column pruning scheme based on similarity for inference on GPUs achieving a 1.5x speedup. 

\parad{block-balanced sparsity}
\citet{2016-han-ese} introduce block-balanced pruning that restricts blocks (``sub-matrices'') to the same sparsity ratio. Thus, when loading blocks in parallel, the accelerator can process them in approximately the same time avoiding load imbalance. They find such pruning does not reduce the model quality significantly for large enough blocks. In an even simpler approach, \citet{2018-dey} fix the degree of each neuron in an MLP, leading to balanced row and column sparsity of the weight matrix. 

\parad{sparse training}
PruneTrain~\cite{2019-lym} focuses on accelerating training using a group-lasso regularization method and pruning during training. The authors mention that the freed memory from pruning can be reinvested during the training to increase the minibatch size. This fits well into existing training schedules~\cite{smith2018dont}.

\citet{2017-mao} specifically analyze the impact of structured sparsity on the accuracy of CNNs. They consider four levels of increasing structure in convolutional layers: (0) unstructured weight sparsity, (1) dimension-wise (blocked) weight sparsity, (2) kernel-level sparsity, and (3) filter-level sparsity. When considering storing the weights array as $W[C, K, H, W]$ (Channel, Kernel, Height, Width), then each of the four levels would require the following addressing per element: (0) $W[C,K,H,W]$, (1) $W[C,K,H,:]$ or $W[C,K,:,W]$, (2) $W[C,K,:,:]$, and $W[C,:,:,:]$. They show that, for a simple magnitude-based (sum per block)  pruning scheme, the top-5 accuracy degrades with increasing block size at sparsity levels of 60--77\%.


\subsection{Hardware acceleration for sparse models}\label{sec:hwacc}

\parad{intro}
Numerous hardware accelerators have been designed to accelerate deep neural networks, see \cite{mlhwsurvey,reuther2020survey} for an overview. Here, we focus on a summary of important techniques implemented in hardware accelerators that have \emph{explicit support for sparse computations} in deep learning. \citet{dave2020hardware} provide a comprehensive and generic survey including more architectures, techniques, and technical details on this topic. 
Accelerator designs are based on the observation that typical workloads have 50--90\% ephemeral activation sparsity and up to 99\% weight sparsity. 
Activation sparsity is either induced by ReLU operations or autoencoders~\cite{noh2015learning} and generative adversarial networks~\cite{goodfellow2014generative} that insert zeros in the upsampling phase of the decoder. 
Furthermore, as we outline in the previous sections, weights can often be structurally sparsified to 95\% (or more) without significant loss in accuracy. 

\subsubsection{Inference accelerators}
\parad{intro}
We start with an overview of sparse inference accelerators that typically aim at latency-sensitive batch sizes of one where the central operation is sparse matrix-vector (SpMV, or weight-activation) multiplication. 
Similarly to dense DNN accelerators, sparse accelerators can achieve record performance up to nearly 22 TOp/W~\cite{2019-zhang-snap}.
Different layer types can be expressed in terms of a small number of primitives. For example, fully-connected layers can be expressed as (sparse) matrix-vector multiplication. Convolutional layers can similarly be expressed as sparse matrix-vector or matrix-matrix multiplication~\cite{7780804}. Conversely, a $1 \times 1$ convolution can be expressed as a fully-connected operator. Similarly, recurrent layers can be unrolled into a series of matrix-vector multiplications. So any device that can process a convolution \emph{or} fully-connected layer can process all layers. Yet, accelerators are tuned for the specifics of layer types and network architectures. Thus, we structure our overview by different architectures.

\paragraph{Sparse CNN inference accelerators}
\parad{CNN}
We start with an overview of sparse CNN accelerators (including fully-connected layers). 
Minerva~\cite{2016-reagen} uses a hand-tuned threshold to prune small activation values in MLPs that save weight-fetch bandwidth and arithmetic operations. This saves 50\% of the power consumption on top of other optimizations such as quantization. Eyeriss~\cite{2016-chen} clock-gates PEs that would process zero activation values of convolutional layers to save energy. 

\citet{2016-han-eie} show Efficient Inference Engine (EIE), an inference architecture optimized for sparse models with parameter sharing. Their architecture supports both sparse matrices as well as sparse activation vectors and aims at fully-connected layers in CNNs. To enable fine-grained parallelism, they distribute the columns of the weight matrix to the processing elements (PEs). At the input, they scan the activations for non-zero entries and broadcast them to all PEs, where they are accumulated into a local partial sum. They balance the load through queues at the PEs that buffer non-zero activation values to avoid synchronization. The authors showed empirically that a queue depth of four values is sufficient to achieve good load balance. Finally, the output activations are summed and compressed through a hierarchical non-zero detection tree. 
Their silicon implementation is 13x faster and 3,400x more energy efficient than an Nvidia Pascal Titan X GPU. 

Zena~\cite{2017-kim} introduces a scheme that uses both weight and activation sparsity for convolutional layers.
Other sparse DNN accelerators such as Cambricon-X~\cite{2016-zhang},  SCNN~\cite{2017-parashar}, Eyeriss v2~\cite{2019-chen}, and Cnvlutin~\cite{2016-albericio} use a combination of similar ideas to achieve between 2--15x speedups and 2--10x lower energy consumption. 
\citet{2020-niu} design an FPGA-based accelerator for the spectral processing (based on FFT and Hadamard products in the frequency domain) of sparse convolutional layers. They keep the input activations in SRAM and stream the sparse kernels. A similar design~\cite{2019-niu} streams activations with stationary weights. Both have limited reuse due to the limited BRAM (on-chip SRAM) on FPGAs. Both store weights (kernels) in COO format arranged in device DRAM and \citet{2020-niu} uses a covering algorithm to optimize locality. 

\paragraph{Sparse RNN inference accelerators}
\parad{sparse RNN inference}
A second class of accelerators aims at sparse recurrent (RNN, LSTM) inference accelerators. 
\citet{2016-han-ese} later show Efficient Speech Recognition Engine (ESE), an FPGA accelerator design for LSTM models using block-balanced sparsity for load balancing. ESE stores (sparse) activations in fast memory with the (sparse) weights being streamed, while the (dense) output is accumulated into a fast output memory.
Their overall design achieves 3x performance and 11.5x energy efficiency improvements on a Xilinx Ultrascale (XCKU60) FPGA compared to an Nvidia Pascal Titan X GPU. 
Those systems are designed for the typical cases of 50--70\% activation sparsity as well as 90\% weight sparsity.
MASR~\cite{2019-gupta} proposes an ASIC design for sparse RNNs as used in speech recognition. They exploit sparsity in weights and activations and different from EIE, they use a bitmask scheme to store indices using a relatively moderate sparsity of 66\%. 

\paragraph{Predicting sparsity in the results}
\parad{output activation sparsity}
Most accelerators utilize either sparsity in the input activations, in the weights, or both. However, one could also aim to predict sparsity in the output activations (i.e., the result of the computation). SparseNN~\cite{2018-zhu} show that such a prediction scheme can improve performance by up to 70\% while halving power consumption. 
The key technique is a light-weight prediction of the non-zero pattern in the output. LRADNN~\cite{2016-zhu} use Singular Value Decomposition (SVD) of the weight matrix: $W=U V$, where $W\in \R^{m\times n}$, $U\in\R^{m\times r}$, and $V\in\R^{r\times n}$, where $U$ and $V$ are the first left- and right-singular vectors, respectively.
The prediction is then simply performed by computing a mask $m = sgn(U V x)$ for the input activations $x$. For small enough $r$, the computation can be 20x faster than evaluating the full layer and the generated sparse output mask can be used to avoid computing zero elements. 
SparseNN~\cite{2018-zhu} improves upon this scheme by learning $U$ and $V$ through back propagation. They estimate the derivation of the sign function with a well-known straight-through estimator~(see Section~\ref{sec:approx_discrete}). 

\paragraph{Fixed block sparsity and systolic arrays}
\parad{block-sparse accelerators}
Block sparsity (either blocks of weights or full neurons) reduces the overhead for indices and control logic and thus can efficiently be used to optimize software for any hardware type (see \cite{2017-yu}). Cambricon-S~\cite{2018-zhou} adds explicit support for block sparsity to the Cambricon series of accelerators. They skip both zero neurons and blocks of weights for arbitrary layer types. They observe that large weights tend to cluster in 2D and 4D in fully-connected and convolutional layers, respectively. Based on this observation, they define 2D and 4D block-sparse weight formats. The block sizes are tuned as hyperparameters and the authors observed that permissible block sizes for ResNets are particularly small where other (``fatter'') networks allow bigger blocks. They show 1.7x better performance and 1.37x better energy efficiency than the fine-grained Cambricon-X accelerator. 

\parad{systolic arrays with packing}
Most of the sparse accelerator architectures define logic that feeds each unit separately. However, most dense accelerators use systolic arrays to perform the matrix operations. Yet, sparsity would not use those fixed structures of systolic arrays efficiently. \citet{2018-kung} pack sparse filters into a dense structure to use efficient systolic arrays for sparse computations. They first select columns of sparse filters/weights that have minimal overlap between the non-zero elements. Then, they combine those into a single column retaining the largest values. For example, consider the following four columns: $c_1=[0,2,0,3,0]$, $c_2=[1,0,2,1,0]$, $c_3=[1,2,0,1,0]$, and $c_4=[0,0,0,0,4]$. If we were to pack three columns, we would select $c_1$, $c_2$, and $c_4$ with the minimal overlap and pack them into the single dense column $c_{p} = [1,2,2,3,4]$---note that only the 4th index conflicts in $c_1$ and $c_2$ and the large value is chosen. The group size and number of allowed conflicts are hyperparameters. The authors show that this scheme, combined with a moderate amount of retraining is efficient for small networks.
Squeezeflow~\cite{2019-li} use a conceptually similar scheme to compress sparse filters. Instead of combining different filters, they decompose sparse convolutions into effective and ineffective and map the effective ones to a dense representation for efficient processing.  
Compact~\cite{2019-zhang-compact} regularizes sparse runlength-encoded activations to be processed in a systolic array. 

\subsubsection{Training accelerators}
\parad{training accelerators}
Since the (inference) forward-pass is a part of training, one could assume that accelerators designed for inference can also be used in the forward pass of training. While this is partially true, it comes with additional complications. For example, during training, one needs to store the activations. Furthermore, specialized formats such as EIE's CSC format cannot easily be accessed (in transposition) during the backward pass. Thus, specific accelerators are designed for sparse training.
\citet{2020-yang} show a co-design approach of a sparse training algorithm and hardware to design Procrustes, an accelerator specific to the Dropback pruning algorithm~\cite{2019-golub}. 
They observe that batch normalization layers ``shift'' values away from zero and essentially eliminate sparsity in the gradients. Procrustes thus exploits only structural weight sparsity by storing weights in a compressed block format. Their design is up to 4x faster and 3.36x more energy efficient than traditional dense training accelerators. 
\citet{2019-zhang-eager} use the observation of early structure adaptation together with an iterative pruning schedule with magnitude pruning to accelerate training by up to 40\%.

\parad{more general accelerators}
More generic accelerators such as 
SparTen~\cite{2020-gondimalla} and SIGMA~\cite{2020-qin} are not specialized to particular layer types and focuses on general sparse matrix-vector products. Both architectures can support arbitrary reuse of matrices or their elements and both are using (blocked) bitmap storage to implement sparse vector products. Thus, their architecture is not specific to any layer type. 
Yet, the sparse matrix storage format determines ranges of sparsity where it performs most efficiently (see Section~\ref{sec:perf}). The used bitmap format performs best for relatively dense operations. 
Similarly, Nvidia's Ampere micro-architecture supports ``structured sparsity'' to accelerate the processing of blocks of four values with up to 50\% sparsity~\cite{a100}.

\parad{accelerators for higher sparsity}
All those architectures are designed for the relatively modest sparsity in today's deep neural networks. One could expect new breakthroughs to enable higher sparsity closer to those in scientific computing (>99.9\%). Then, another class of accelerators, such as SpArch~\cite{2020-zhang}, Indirection Stream Semantic Registers~\cite{issr}, or Extensor~\cite{2020-hegde} would play a bigger role.

\subsubsection{Overview of accelerators for sparse deep learning}

Table~\ref{tab:accelerators} shows an overview of existing accelerator architectures with sparsity support. Most accelerators are designed for inference and most can also be used for the feed-forward pass during training---albeit not always efficiently. We underline accelerators that are specifically designed for training. 
\begin{table}[h!]
\begin{center}
	\begin{footnotesize}
	\begin{tabular}{ p{2.5cm}|p{0.8cm}|p{.94cm}|p{.94cm}|p{.94cm}|p{2.5cm}|p{1.3cm} } 
		\hline\hline
		\textbf{Accelerator} & \textbf{Ops} & $\vec{w}$\textbf{ mem} & $\vec{y}$ \textbf{mem}  & $\vec{y}$ \textbf{comp} & \textbf{Load Balancing} & \textbf{Reuse} \\
		\hline
Cnvlutin [\citeyear{2016-albericio}]  & conv  & -   & COO$^*$  & x & group neurons & output\\
EIE [\citeyear{2016-han-eie}] 		  & FC    & CSC & - & x & activation queuing & output\\
Minerva [\citeyear{2016-reagen}]      & FC    & -   & - & x & N/A & - \\
Cambricon-X [\citeyear{2016-zhang}]   & SpMM  & CSC & - & x & N/A & output\\
Eyeriss [\citeyear{2016-chen}]        & conv  & -   & - & x & - & row \\
ESE [\citeyear{2016-han-ese}]         & LSTM  & CSC & - & x & block-balanced & output\\
SCNN [\citeyear{2017-parashar}]       & Conv  & CSC & CSC & x & N/A & input act.\\
SparseNN [\citeyear{2018-zhu}]        & FC    & -   & - & x & N/A & N/A\\
Cambricon-S [\citeyear{2018-zhou}]    & SpMM  & COO & - & x & group output neurons & output\\
Zena [\citeyear{2017-kim}]            & Conv  & BM  & BM & x & dynamic group alloc. & N/A\\
Eyeriss v2 [\citeyear{2019-chen}]     & SpMM  & CSC & CSC & x & activation queuing & row \\
SparTen [\citeyear{2020-gondimalla}]  & SpMM  & BM  & BM & x & precomputed greedy & output\\
MASR [\citeyear{2019-gupta}]          & RNN   & BM  & BM & x & dyn. act. assignment & N/A\\
SPEC$^2$ [\citeyear{2019-niu}]        & Conv  & COO & - & - & - & weight/kernel\\
\underline{Eager Pruning [\citeyear{2019-zhang-eager}]} & SpMM & BM  & - & x & dynamic output & weight\\
Spectral CNN [\citeyear{2020-niu}]                      & Conv & COO & - & - & - & input act.\\
\underline{Sigma [\citeyear{2020-qin}]}                 & SpMM & BM  & BM & x & - & input/weight\\
\underline{Procrustes [\citeyear{2020-yang}]}           & SpMM & CB  & - & - & split minibatch for LB & minibatch \\

		\hline
	\end{tabular}
	\end{footnotesize}
\end{center}
\caption{Overview of accelerators for sparse deep learning; those with explicit training support are underlined.}
\label{tab:accelerators}
\end{table}

Some accelerators aim at either sparse matrix vector (SpMV) or sparse matrix matrix (SpMM) multiplications that can be used for several layer types (e.g., fully connected, convolutional using the Winograd scheme, or RNNs). Others are optimized specifically for convolutional or recurrent layers. 
The second column of the table (\textbf{Ops}) shows the operation (layer) types that the accelerators were optimized for explicitly (FC = fully connected, LSTM = Long Short Term Memory, RNN = Recurrent Layers - all SpMV and Conv = Convolutional Layer - all SpMM via im2col). If an accelerator aims at both FC and Conv, we mark it with SpMM as a superset. As mentioned above, most accelerators can process all layer types at varying efficiency. 

The column ``\textbf{w mem}'' lists the storage scheme for weights. A ``-'' means that weights are stored densely and zeros are computed explicitly.
The columns ``\textbf{y mem}'' and ``\textbf{y comp}'' list whether activations are stored compressed and whether they are computed. Some accelerators store zeros but filter them before the computation engine.  
When we write CSC (Compressed Sparse Column), we include runlength encoding even though, in some special cases, the column offsets are managed outside the format. Cnvlutin uses a blocked COO format and Procrustes uses a compressed block (CB) format.
The last two columns show specific techniques for \textbf{load balancing} and \textbf{reuse}. They are described in the section above and listed as a summary.




\section{Discussion}\label{sec:discussion}

We do not understand all details of the inner workings of deep neural networks and how pruning influences them. Specifically, why can networks be pruned and what is the best pruning methodology remain as open questions. 
In this section, we provide a set of hypotheses, intuitive explanations, and possible assumptions to foster our understanding of the landscape and the characteristics of this gap in understanding.
All those are speculation and intended to help readers to develop a better feeling for the area as well as inspire new research directions that could shed more light onto aspects of sparse neural network science and methodology. 

\parad{sparse vs. dense models in general}
A general observation in most works is that sparse models outperform dense models given the same parameter budget. Some works even show that sparse models outperform dense models with larger number of parameters~\cite{2019-lee-init,2020-elsen}.
A similar set of observations seems obvious but is worth stating: pruning is most efficient for architectures that are overparameterized. Some authors state that switching to a better architecture may be more efficient than pruning~\cite{2020-blalock}. This implies that, when showing relative pruning rates (e.g., 99\%), one should always consider the degree of over-parameterization or what we call the ``parameter efficiency'' (see Section~\ref{sec:parameff} and ``Rule 1'' in \cite{benchmarking}).

\subsection{Relation to Biological Brains}

\parad{careful with analogy}
Throughout the document, we have used many metaphors linking approaches to biological brains, whose structure inspired the general idea of all neural networks.
While such metaphors can be very useful to build an intuition and provide a possible direction, they have to be considered carefully. 
Biological brains and computers work with fundamentally different compute substrates. For example, the three-dimensional arrangement of the brain encodes structural information nearly for free and learns through neuroplasticity. 
Silicon devices cannot adapt their wiring structure easily, and thus the simulation of structural properties leads to overheads in terms of memory (encoding the structure) as well as compute (controlling the execution). 
It is thus possible to design mechanisms that are not common in biological systems but outperform biologically more plausible mechanisms in silicon-based compute substrates and architectures. After all, not many animals have wheels and airplanes do not flap their wings. Nevertheless, Leonardo da Vinci discovered the principle of dynamic soaring by studying birds.   

A successful method to guide innovation is to be inspired by biological phenomena and engineer systems in a refinement and optimization step given our technical understanding of the problem. 
For example, the visual cortex does not utilize weight sharing like convolutional networks do, however, in silicon, it seems to be the most efficient technique given that weight sharing reduces redundancy during feature detection and enables reuse for performance~\cite{unat-locality}. 
A second example could be the optimization process. While we currently use SGD to train networks, it remains unclear whether biological brains use similar methods. Recent discoveries have shown a possible relationship to Hebbian learning~\cite{millidge2020predictive} and argue that SGD may be biologically plausible, albeit some gaps remain~\cite{lillicrap2020backpropagation}.

Various pruning approaches have been directly inspired by biological brains~\cite{ahmad2019dense} but have not demonstrated state of the art results for large-scale networks and complex tasks. They advocate sparse high-dimensional representation spaces. Biological brains have very large sparse layers in a relatively shallow architecture with less than ten layers. We believe that this is a very interesting direction for further exploration and inspiration if it is augmented with theoretical reasoning and solid engineering.

\subsection{Permutation Groups and Information Loss}

\parad{equivalences}
One interesting observation is that every parameterized dense network is an element in an exponentially large equivalence class, which will generate the same output for each input. Specifically, \citet{2017-changpinyo} prove the following lemma: ``any dense convolutional neural network with no cross-channel nonlinearities, distinct weights and biases, and with $l$ hidden layers of sizes $n_1$, $n_2$, $\ldots$, $n_l$, has at least $i=\Pi_{i=1}^l n_i!$ distinct equivalent networks which produce the same output.''
This suggests that the information content of sparsified networks may be exponentially larger. 
\citet{ahmad2019dense} show a similar result with respect to noise robustness in high-dimensional vector spaces. 

\subsection{Sparse subnetworks for training and lottery tickets}\label{sec:lottery}

\parad{lottery ticket intro}
Some works hinted at specific subnetworks that may exist during training which could lead to a good sparse structure~\cite{2015-sun,2016-cohen}. \citet{2016-see} demonstrated that re-training a sparse RNN with the same structure results in networks that perform well but not as well as the pruned-from-dense variants.
\citet{2019-frankle} analyze the relation between initialization and statically sparse training. They state the ``Lottery Ticket Hypothesis'': ``dense, randomly-initialized, feed-forward networks contain subnetworks (winning tickets) that—when trained in isolation—reach test accuracy comparable to the original network in a similar number of iterations.''
For shallow vision networks, they find winning tickets by magnitude pruning and show that re-training them with static sparsity starting from the initial weights, they reach similar or higher accuracy in the same number of iterations. They also demonstrate that random initialization, with the same structure, does not suffice.  
\citet{2019-zhou} empirically show that one may not need the exact weights at initialization to train lottery tickets but the signs may be sufficient. 

\subsubsection{Pruning is all you need - networks without weight training}
Several researchers argue that initial subnetworks with their random weights can perform well~\cite{ramanujan2020whats,2019-zhou}. 
Furthermore, winning tickets already identify sub-networks with non-random accuracies even without training. In fact, training to find such a ``supermask'' can produce a network that achieves 93.5\%  accuracy in MNIST and 65.4\% accuracy on CIFAR-10 at around 50\% sparsity without changing the random initial weights. 
In ``pruning is all you need'', \citet{2020-malach} prove that, with high probability, any network can be approximated with $\epsilon$ accuracy by pruning a polynomially larger network.
This means that pruning could be used to train a network without changing the weights at all. 
\citet{2020-orseau} and \citet{2020-pensia} later prove that a logarithmically larger network (except depth) suffices. Specifically, any ReLU network of width $n$ and depth $d$ can be $\epsilon$-approximated by sparsifying a $\mathcal{O}(\log(nd))$ wider and two times deeper random network, with high probability~\cite{2020-pensia}.

\subsubsection{Lottery tickets in large networks}
\parad{lottery ticket discussion}
Analysis of the original lottery ticket hypothesis already indicated problems with larger CNNs, which could be fixed with a decreased learning rate. 
\citet{2018-liu} showed that with the best learning rate for larger networks, keeping the original initialization does not improve the final accuracy over random initialization. \citet{2019-gale} also could not reproduce the hypothesis for larger networks. 
\citet{2019-frankle-b} later argue that the hypothesis (``with rewinding'') applies also to larger networks if one uses the values after some initial optimization steps at iteration $r$. They demonstrated that 0.1--7\% of the total iterations are sufficient for 50--99\% sparse networks. In line with early structure adaptation  (see Section~\ref{sec:sparsify-during-training}), they conclude that early pruning could be a promising approach.
However, finding the right $r$ remains tricky and the authors investigate the influence of ``noise'' through the ordering of batches on the training process and result~\cite{2020-frankle-linear}. Specifically, they investigate the difference in test accuracy for a model that is a smooth interpolation between two models trained with different orders. They consider networks with small such error and allow the orders to only diverge after iteration $r$. The empirical results show that $r$ relates to the iteration for which a working lottery ticket can be derived by rewinding.
\citet{2020-frankle-early} empirically analyze the early iterations of training large networks.

\citet{2020-renda} compare the standard single-shot fine-tuning after pruning to ``weight rewinding''. Rewinding resets the weights after pruning to the values of a previous SGD iteration $i$. Then, they retrain (fine-tune) with the same learning rate schedule (from the original iteration $i$) in a process called ``Iterative Magnitude Pruning''. A modification to the scheme simply uses the same learning rate schedule but without resetting the weights. However, both \citet{2020-savarese} and \citet{2020-chen} find rewinding to be less efficient than fine-tuning from the most recent weights for image recognition and natural language processing tasks. They show for a variety of medium-sized ResNets and GNMT as well as BERT that weight rewinding outperforms fine-tuning but is itself outperformed by just rewinding the learning rate to the first iteration. \citet{2019-ding} found that a simple selection based on 1st order information outperforms the simple magnitude-based scheme. 
\citet{2020-morcos} show that lottery tickets can transfer across different datasets and optimizers.
A general conclusion could be that fully sparse training is possible (see Section~\ref{sec:sparse_train}), especially if applied iteratively (see Section~\ref{sec:schedules}) but rewinding has not been proven effective.

\subsection{Structured vs. unstructured pruning}

\parad{intro}
Several works found that unstructured/fine-grained (e.g., weight) pruning maintains a better accuracy per element than structured/coarse-grained (e.g., filter, neuron) pruning~\cite{2019-gomez,2015-han,2017-ullrich,2019-lee-init}. 
However, structured pruning approaches achieve much higher computational performance on modern devices~\cite{2019-lym,2016-wen}. 
Thus, structured sparse models could afford a higher number of iterations to train and more floating point operations during inference to achieve the same overall efficiency/cost. 
Furthermore, unstructured sparsity has a higher relative representational overhead of indices for each fine-grained element as discussed in Section~\ref{sec:perf}.
It remains to be seen what level of granularity will be most efficient for the coming computer architectures.

\parad{random initialization}
We also observe that random pruning at network initialization works significantly better for neurons and filters than for weights~\cite{2019-gomez}. For neurons and filters, most works that nearly reproduce the state of the art are achieved with post-training sparsification, indicating that this form of architecture search is efficient. This is also intuitive because the specific location of neurons or filters no standard fully-connected and convolutional layers is irrelevant. 
For weights, this very structure matters and thus random pruning at initialization performs generally worse. Thus, we recommend different schemes for structured vs. unstructured pruning in order to utilize training resources best.

\subsection{Optimization algorithms during model training}\label{sec:opt}

\begin{figure}[h!]
	\includegraphics[width=0.9\textwidth]{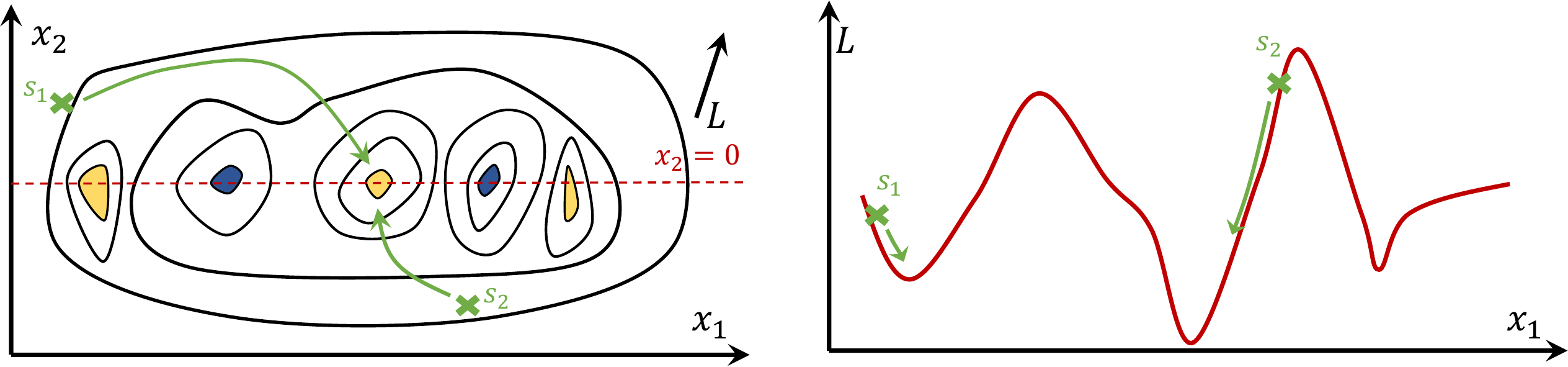}
	\caption{SGD in a 1D loss landscape.}
	\label{fig:disc_sgd}
\end{figure}

Stochastic gradient descent (SGD) is the de-facto standard algorithm in training deep neural networks. 
Most of the works investigating sparse training suggest that SGD is sensitive to the parameters as well as the network structure.
Several show empirically that training larger models is more compute-efficient than training smaller models~\cite{glorot2011deep,mhaskar2016deep,2020-li,kaplan2020scaling}. 
We conjecture that this may be explained by the iterative optimization process and the ability to use additional dimensions to ``route around'' hills in the loss landscape. 
Thus, high-dimensional dense spaces help to elude local minima in the loss landscape as illustrated in Fig.~\ref{fig:disc_sgd}: the left side shows a two dimensional function $f(x_1,x_2)$ and the loss function $L$ as contour lines. Yellow areas are valleys and blue areas are hilltops. 
The red dashed line shows the value $x_2=0$, emulating a sparsified model, which is shown in the right plot. Here, we plot the (same) loss function on the x axis. 
We show two possible starting points $s_1$ and $s_2$ and SGD trajectories in green on both sides.
We see that the $x_2$ dimension can be used to circumvent the leftmost hill when starting from $s_1$ in the two-dimensional model and proceed to the lowest minimum in the middle. 
However, when we sparsify $x_2$ in the right model, SGD will work in the projected subspace with less degrees of freedom and converge to the suboptimal minimum. 

Furthermore, when sparsified, the Lipschitz constant of the loss function increases~\cite{2020-evci,2020-lee} and complicates the optimization further. Modern techniques such as momentum can improve the optimizer but then may require more iterations~\cite{2020-lee}.

We may attribute this ``weakness'' of SGD to its fundamental property of linear first-order descent. \citet{domingos2020model} further hardens this claim by showing that models trained with SGD are approximately kernel machines.

As we have seen, iteratively applying pruning improves the quality of pruned models significantly. If we now see this overall optimization process as a series of (linear) SGD iterations mixed with (nonlinear) pruning steps, this new optimization process implements a guided nonlinear search. At each pruning step, the function is perturbed in a guided way (depending on the pruning methodology, see Section~\ref{sec:removal}) and then again minimized with SGD. At each pruning step, the model may evade a local minimum that SGD alone may not be able to overcome. 
For well tuned schedules, this scheme seems to approximate an efficient learning algorithm. 

\parad{pruning as noise}
\citet{2019-bartoldson} model pruning as ``noise injection'' to explain improved generalization capabilities, which would fit this mental framework. They specifically consider the drop of test accuracy right after pruning in iterative pruning schemes. They show empirically that a higher drop relates to better generalization of the final model. They suggest that smaller models may not be the only reason for improved generalization and carefully tuned magnitude pruning schedules can improve generalization by ``flattening'' the loss landscape.

\subsection{Emerging Benchmarks} 

\parad{intro}
Interpreting pruning results and comparing different methods is difficult due to the wide range of experimental setups, tasks, techniques, and hyperparameters used. This issue has already been identified by \citet{2020-blalock} who propose a standard methodology together with a set of benchmarks to solve this issue. One could imagine standard setups such as MLPerf~\cite{2019-mattson} or the Deep500 infrastructure~\cite{bennun2019modular} for performance measurements.
We note that even before such a benchmark is widely accepted by the community, some datasets, tasks, and network architectures are emerging as \emph{de-facto} benchmarks for pruning. We recommend researchers to use those as comparison points. As we point out above, ResNet-50 on ImageNet and BERT on the GLUE tasks seem excellent candidates for such standard benchmark sets for both model sparsity and performance. 

We observe that the achieved sparsity at high accuracies strongly correlates with the attention that certain models received in the literature. For example, ResNet-50 is well tuned and thus shows higher achieved parameter efficiencies relative to other models. Thus, they effectively define the state of the art---however, this observation also means that one cannot easily reason about the ``prunability'' of a certain architecture without extensive experiments on a level playing field. 

For toy examples, the MNIST dataset with the LeNet-300-100 and LeNet-5 networks can act as a good calibration. The state of the art is above 98\% accuracy with less than 1\% of the original parameters. However, we insist that this task alone is not indicative of good performance of a method. 
More meaningful tasks are larger convolutional networks on more complex tasks such as CIFAR-100 and ImageNet. 
In order to track progress, we recommend that those should always be reported when analyzing new pruning methodologies even though better architectures for these tasks (or better tasks) may exist. Additionally, in our experience global magnitude pruning is a good baseline method for a wide range of scenarios, see e.g.,~\citet{2020-singh} for results. 

\subsection{Parameter Efficiency}\label{sec:parameff}

\parad{parameter efficiency}
One could define the general concept of parameter efficiency as ``How much does the average parameter contribute to the overall quality of the model?''. We observe that, when pruned, the parameter efficiency often increases while the overall model quality decreases. 
\citet{2018-bianco} propose \textit{accuracy density} as a measure of parameter efficiency. It is defined as the validation accuracy (in percentage) divided by the number of parameters (in millions). With the metric, the authors show clear benefits for MobileNet (both versions) over ResNet-50, but also a benefit of AlexNet and SqueezeNet, both under 60\% top-1 accuracy, over VGG-16 (with 71.6\% accuracy). 
When extended to pruned DNNs, accuracy density increases disproportionally, with sparse but inaccurate models ranked highest and orders of magnitude of difference.
It is thus apparent that not every validation sample is as easy to predict as the others, and the measure should \emph{not} be linear with the count of correct predictions.

\begin{figure}[h!]
	\centering
	\begin{subfigure}[b]{.45\linewidth}
		\centering
		\includegraphics[height=1.45in]{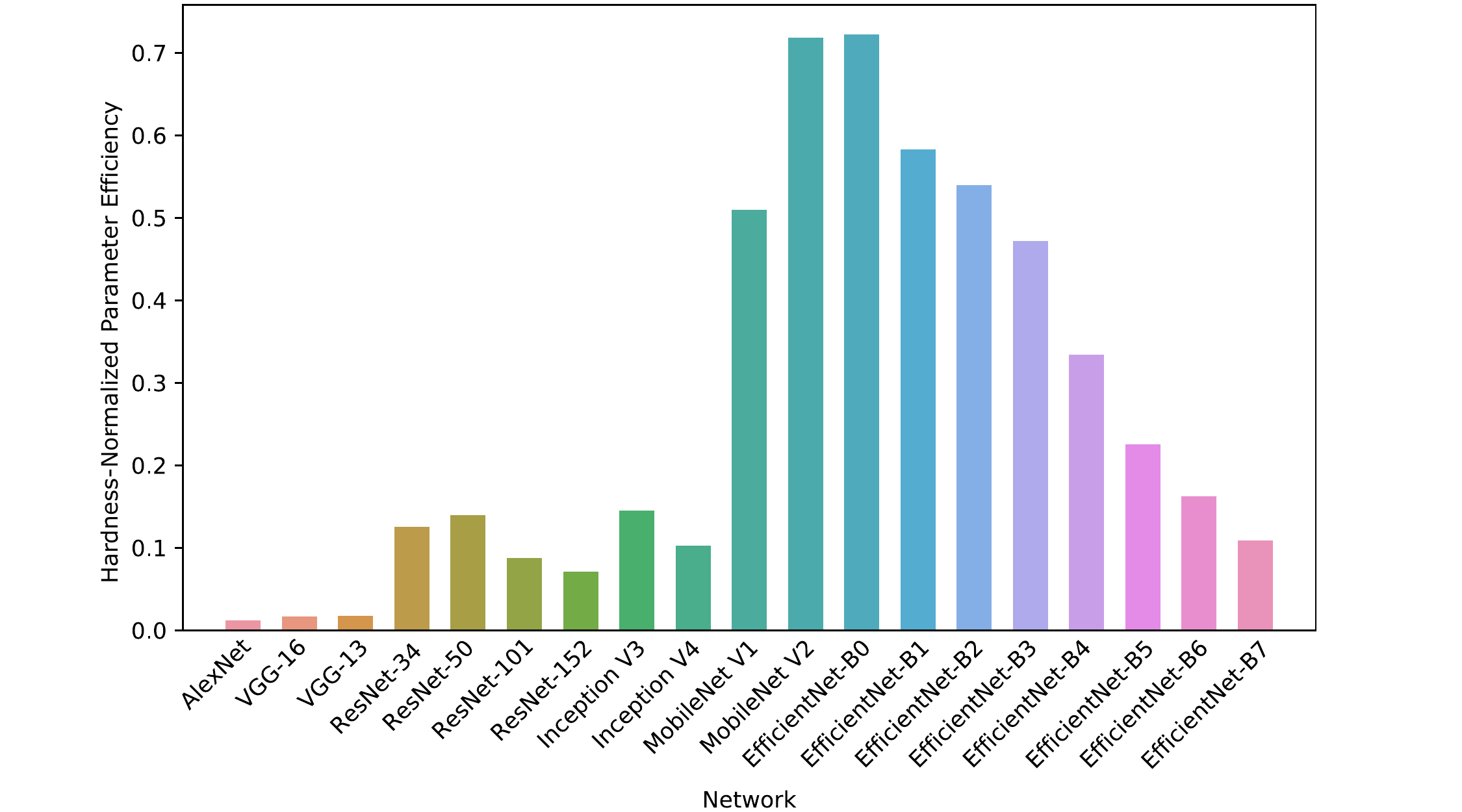}
		\caption{Dense Networks}
		\label{fig:parameff:dense}	
	\end{subfigure}
	\qquad
	\begin{subfigure}[b]{.45\linewidth}
		\centering
		\includegraphics[height=1.5in]{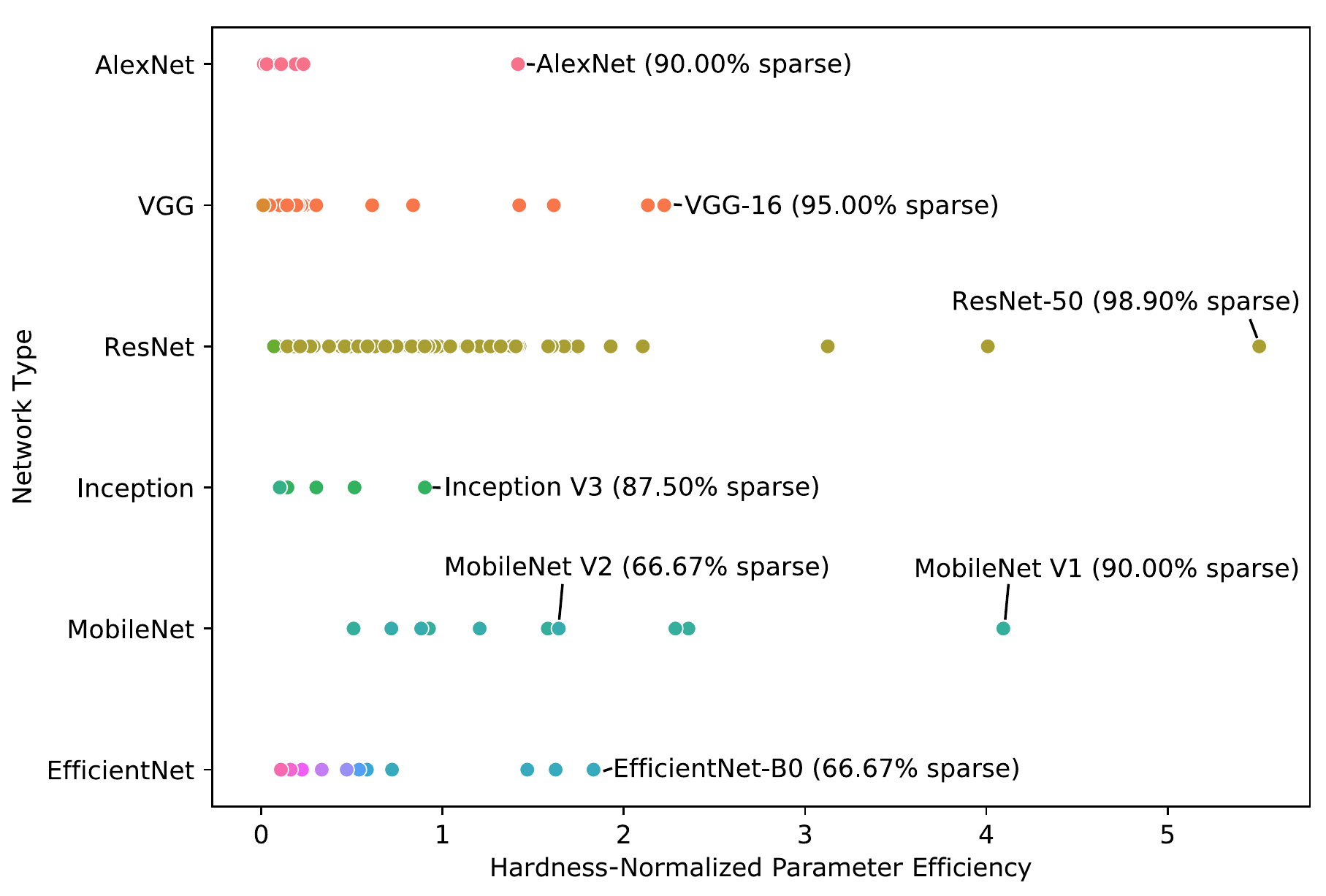}
		\caption{Sparse Networks}
		\label{fig:parameff:sparse}	
	\end{subfigure}
	\caption{Parameter efficiency of state-of-the-art DNNs on the ImageNet dataset, with color indicating DNN type. Hardness-normalized parameter efficiency is normalized based on a logarithmic fit of the top-1 validation accuracy  of DNNs over the years 2012--2020: $f(x)=5704.7\cdot\ln{x}+30908$ (as correctly predicted images).}
	\label{fig:parameff}
\end{figure}

\parad{a renormalized parameter efficiency metric}
To deal with parameter efficiency in the face of varying classification difficulty, we define a slightly modified measure called \textit{hardness-normalized parameter efficiency}. Instead of computing the ratio of accuracy to parameters, we normalize the number of correct predictions by their relative difficulty. To estimate classification difficulty for ImageNet, we fit a function through the state-of-the-art DNN-based image classifiers over the years (Fig.~\ref{fig:parameff}), and then evaluate the number of correct classifications by the inverse function to obtain the hardness-normalized correct predictions, and divide by the number of parameters (in millions).

\parad{figure highlights}
The hardness-normalized parameter efficiencies of popular dense and corresponding sparse CNNs are presented in Fig.~\ref{fig:parameff:dense} and \ref{fig:parameff:sparse}, respectively.
For dense networks, we can see that parameter efficiency similarly increases for ResNets and MobileNets over AlexNet and VGG, but that VGG variants are actually more parameter efficient than AlexNet, despite being twice larger. EfficientNet-B0 is roughly on the same parameter efficiency as MobileNet (v2), which is reasonable given that the former network is a mobile-sized baseline, albeit produced via Neural Architecture Search.
For the sparsified networks, most of the pruned networks are more parameter efficient than the best dense networks. We see that the top ranked CNN is a pruned ResNet-50~\cite{2020-savarese}, which can achieve 66\% validation accuracy with only $\approx$281,000 parameters. The second best network is a pruned MobileNet (v1) with 68\% accuracy for $\approx$423,000 parameters.
It may be interesting to investigate this metric in more depth (e.g., with different normalization scales) to understand whether the efficiency per parameter increases monotonically with smaller networks or whether the decrease in model quality leads to a decrease in parameter efficiency as well. This could provide some insight into optimal sparsity levels.

\begin{figure}[h!]
	\centering
	\begin{subfigure}[b]{.45\linewidth}
		\centering
		\includegraphics[height=1.45in]{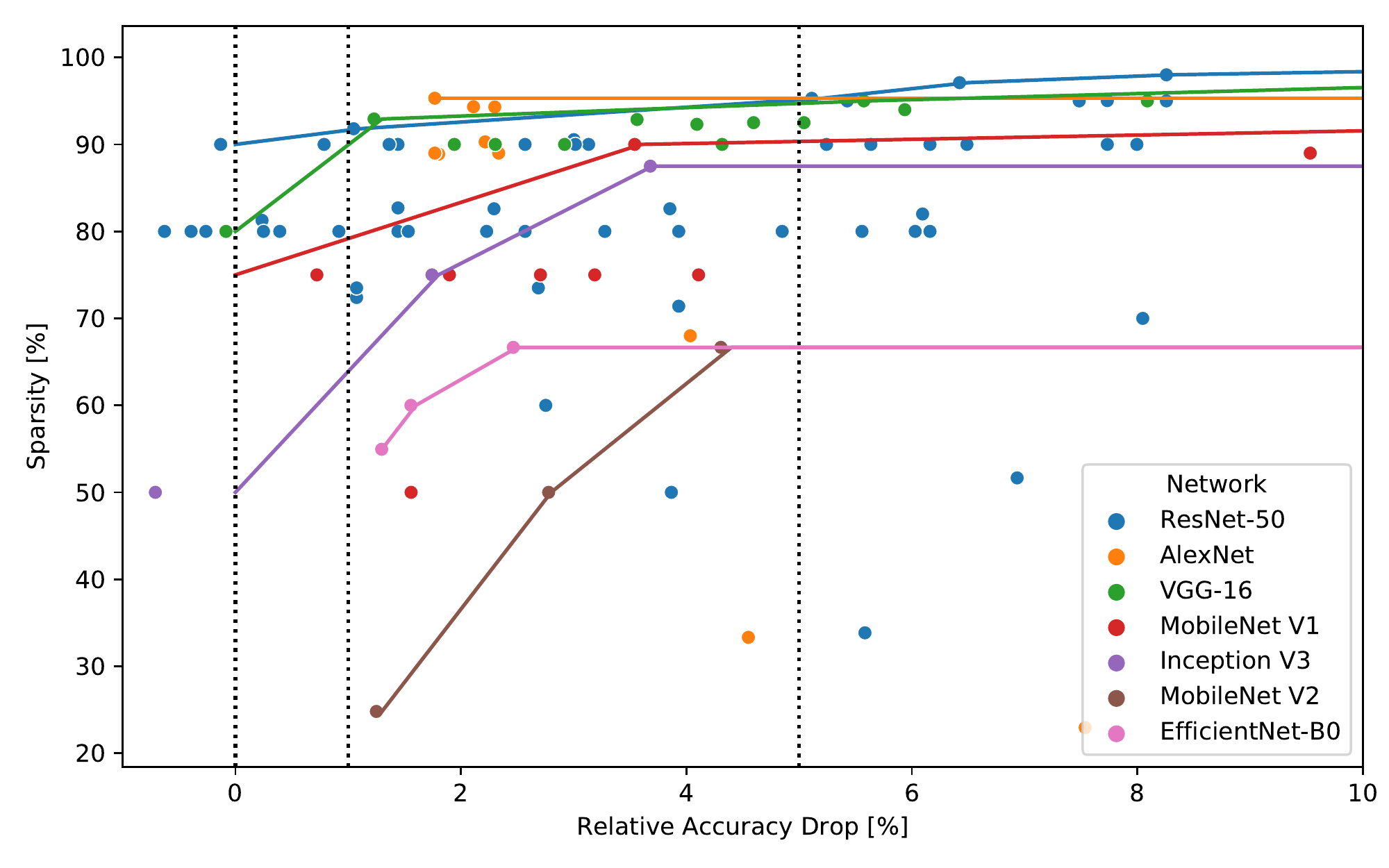}
		\caption{Grouped by network}
		\label{fig:paramslack:network}	
	\end{subfigure}
	\qquad
	\begin{subfigure}[b]{.45\linewidth}
		\centering
		\includegraphics[height=1.45in]{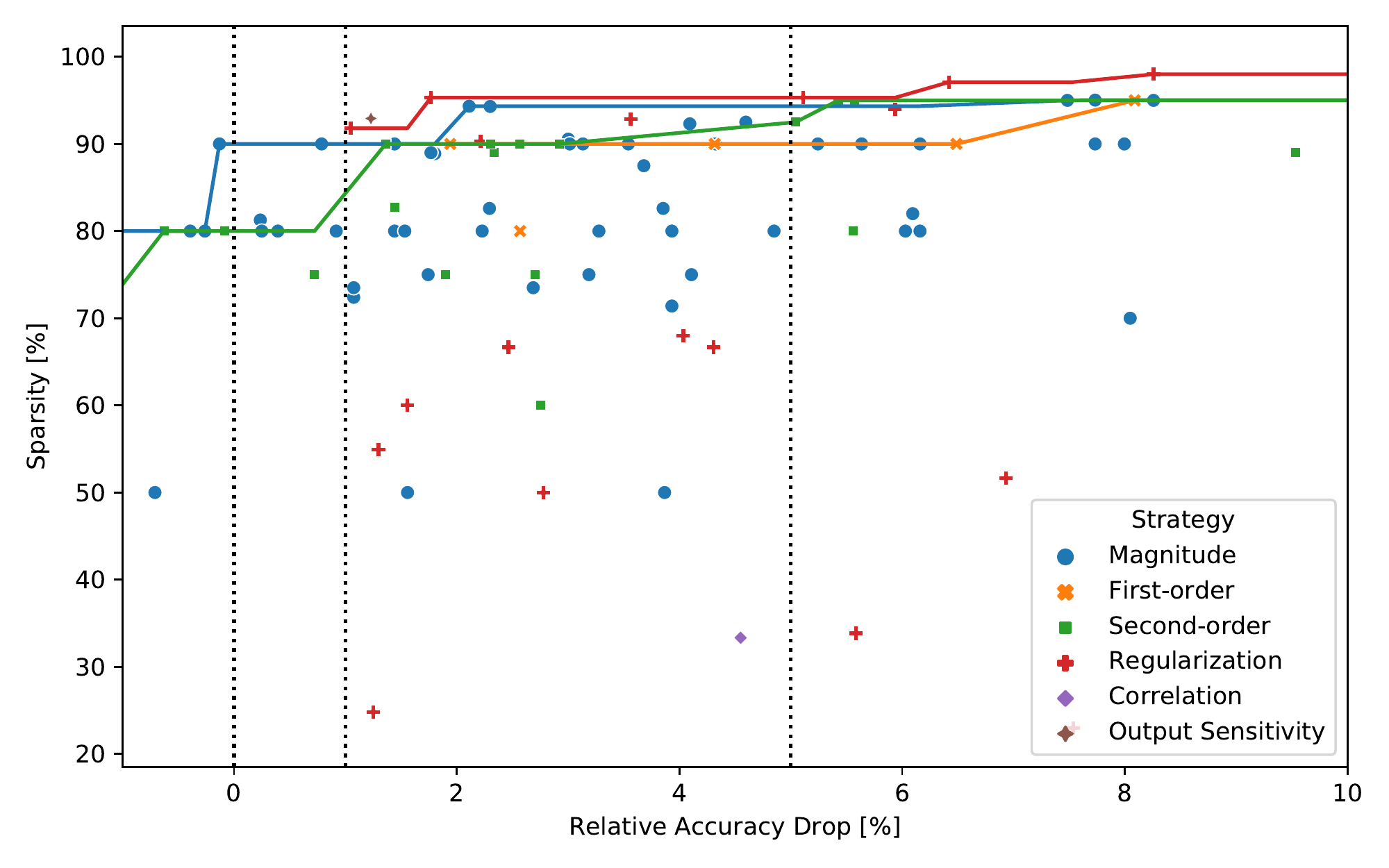}
		\caption{Grouped by strategy}
		\label{fig:paramslack:strategy}	
	\end{subfigure}
	\caption{Relative validation ImageNet accuracy loss for different pruning densities, strategies, and neural networks. Solid lines represent best-performing networks, whereas dotted lines represent accuracy thresholds (e.g., 1\% relative accuracy reduction is the maximum allowed by MLPerf ImageNet rules~\cite{2019-mattson}).
	Negative accuracy drop means improvement in generalization.}
	\label{fig:paramslack}
\end{figure}

\paragraph{Parameter Slack} Figure~\ref{fig:paramslack} shows a relative view of the same data. It shows what sparsity level is achievable if we allow a fixed decrease in accuracy, relative to the dense baseline. 
Since the sparsity is relative to the original network (and its parameter efficiency), it is hard to compare different networks in this figure; instead, we recommend to consider the curve of each network in isolation with the vertical dotted lines at markers of 0\%, 1\%, and 5\% accuracy loss budget. (This metric is inspired in part by the MLPerf ImageNet rules~\cite{2019-mattson}.) The results suggest that architectures such as AlexNet or VGG-16 have significantly higher parameter slack than, e.g., MobileNet or Inception. 
Fig.~\ref{fig:paramslack:network} shows the data grouped by network type. It allows to reason about ``parameter slack'', i.e., the steeper the curve, the higher the percentage of parameters which can be removed while preserving some percentage of the baseline accuracy. Fig.~\ref{fig:paramslack:strategy} shows the same data but grouped by element removal scheme and thus allows a comparison of different schemes. 

\parad{datatypes and binarized neural networks}
Apart from parameters, model information content (and thus parameter efficiency) is also encoded in the data type of the weights themselves.  
This is explicitly clear in binarized networks that have only values $\in \{-1,1\}$. In these cases, the additional zero value adds another piece of information, similar to ternary networks~\cite{2020-li-bits}. Networks with larger weight data types also benefit from the sparsity, however, it remains unclear whether the overhead of storing the non-zero structure (see Section~\ref{sec:perf}) is worth the gain in parameter efficiency.

\paragraph{Sparsification versus manual design}
\parad{sparsification to replace intelligent network design}
An interesting observation is that sparsification of older architectures often does not achieve the same gains as architectures developed later. 
Yet, many breakthrough results in the area of efficient convolutional neural networks can be seen as manually defined sparsifiers, such as bottleneck layers or depthwise separable convolutions~\cite{howard2017mobilenets,iandola2016squeezenet}. The resulting optimized networks are often harder to sparsify. In some sense, this manual design of inductive biases into the network is similar to feature engineering that deep neural networks replaced to begin with. 
Newer works on transformers suggest the more automated way of ``train big and then prune''~\cite{2020-li}. Here, we rely on the learning process to automatically discover good network designs. 
It is to be shown whether such automated methods can compete with hand-crafted biases for modern networks such as transformers.

We close the discussion on parameter efficiency with an observation: 
\emph{Interestingly, the fact that most of the (very different) methods presented in the literature reach similar results in terms of accuracy at a given sparsity (within relative 1\%) suggests that there are inherent compression thresholds which may be hard to overcome.}

\subsection{Generalization and biases}

\parad{Set up.}
It is a surprising fact that neural networks can be heavily pruned without impacting their overall accuracy.
Yet this raises a question: \emph{Is top-level accuracy sufficient to capture the effects of pruning when the neural network representation has changed so dramatically?}
\parad{Describe existing work.}
In recent work, \citet{2019-hooker} show that using the unstructured iterative magnitude pruning of \citet{2017-zhu} on CNNs for image classification results in a large degradation in accuracy for a small number of classes in tasks such as ImageNet, compared to the model's overall decrease.
These classes were typically less well represented in the training data.
Interestingly, they also find that, compared with pruning, quantization results in a much smaller impact to different classes.
Further, they find that pruned models are significantly more brittle under distribution shifts, such as corrupted images in ImageNet-C~\cite{2019-hendrycks-imagenetc} or naturally adversarial images in ImageNet-A~\cite{2019-hendrycks-imageneta}.

\citet{2020-hooker} build on these results and show that the increased errors on certain classes caused by pruning can amplify existing algorithmic biases.
On CelebA~\cite{2015-liu-celeba}, a dataset of celebrity faces with significant correlations between demographic groups, pruning increases errors on underrepresented subgroups.
For example, pruning a model trained to identify people with blond hair to 95\% sparsity increased the average false-positive rate for men by 49.54\%, but by only 6.32\% for others.

\parad{Future research directions.}
The biases and brittleness introduced by pruning may limit the utility of pruned models, especially in situations that often deal with protected attributes and are sensitive to fairness, such as facial recognition or healthcare.
This is unfortunate, since these domains typically deploy models in resource-constrained environments where pruning is particularly valuable.
Therefore, it is important to study the finer-grained impacts of pruning, rather than just the overall accuracy.
Identifying the impact of pruning methods beyond iterative magnitude pruning, and developing more robust pruning methods, are critical open problems.

\subsection{Best practices}

\parad{easy to get started but gets hard in the upper end}
We now focus more on the practical aspects of pruning and conclude the discussion with a set of recommendations we identified based on the body of literature in the field. 
We first note that a flurry of simple approaches enables reaching moderate sparsity levels (e.g., 50--90\%) at the same or even increased accuracy. It seems that any non-silly scheme achieves some sparsification and that there is an inherent robustness in the networks themselves. 
However, reaching higher sparsity levels (e.g., $>$95\%) requires more elaborate pruning techniques where we may be reaching the limit of gradient-based optimization techniques for learning. 
We now provide best practices in five categories that we  recommend everyone to follow when performing pruning in practice.

\paragraph{1. Pruning strategy} 
In general, highest sparsity is achieved using regularization methods in combination with iterative pruning and growth schedules. These methods have high computational costs, sometimes causing a five-fold increase in training overheads, e.g.,~\citet{2020-savarese}. 
Regularization methods are relatively hard to control and require numerous hyperparameters. The simplest training method, magnitude pruning, is easiest to control for target sparsity and accuracy in many practical settings. 
In most training methods, it is important for the structure search to enable weights to regrow, especially in phase of early structure adaptation at the beginning of training. 

\paragraph{2. Retraining/fine-tuning} 
If the focus of sparsity is to improve inference, then retraining/fine-tuning is an essential part of a sparsification schedule. 
Gradually pruned sparsification schedules perform best and it is most efficient to start each iteration from the most trained/last set of weights. 

\paragraph{3. Structure}
Structured pruning seems to provide a great tradeoff between accuracy and performance on today's architectures. 
This is partly due to the fact that hardware and frameworks are tuned for dense blocked computations. 
Furthermore, structural pruning can form a strong bias towards powerful mechanisms like locally connected layers that, together with weight sharing, yield convolutional layers. 

\paragraph{4. Distribution}
The sparsity distribution across layers/operators needs to be considered carefully. 
For this, one could hand-tune the sparsity levels for each operator type and position in the network. For example, dense layers can often be pruned more than convolutional layers and the first layer in a convolutional network can hardly be pruned. 
A simpler scheme may use a global sparsity and a learned allocation strategy. 

\paragraph{5. Combined ephemeral and model sparsity} 
Any sparse deep neural network should combine both ephemeral and model sparsity. For example, dropout often functions as a ``pre-regularizer'' and can benefit generalization greatly if enough data is available. 
Furthermore, ephemeral and model sparsity lead to a multiplicative benefit in terms of needed arithmetic operations.

\section{Challenges and Open Questions}\label{sec:challenges}

We now outline ten central challenges and open questions in the field to inspire future research. 

\begin{enumerate}  
	\item \textbf{Sparse training.} Can we use sparsity to train gigantic models whose dense version would not fit into the hardware budget? How do we sparsely train models without accuracy loss?
	\item \textbf{Structured vs. unstructured.} How does a structural bias influence the accuracy performance and model size tradeoff?
	\item \textbf{Hardware co-design.} How do we co-design hardware architectures and pruned models? What is the tradeoff between cost, accuracy, and structured sparsity?
	\item \textbf{Multi-objective pruning.} What is the best way to prune for multiple objectives simultaneously, e.g., lowest energy consumption for a certain memory size?
	\item \textbf{Architecture design.} Should one use neural architecture search (NAS) for finding efficient networks or can pruning replace NAS?
	\item \textbf{Theory of sparse learning.} What is the relationship between sparsity, learning dynamics, and generalization? 
	\item \textbf{Sparse representations.} What is the representational power of sparse neural networks? Could parameter efficiency be defined rigorously?
	\item \textbf{Method generalization.} Which of the pruning methods for MLPs or CNNs generalize to transformers or other neural architectures? 
	\item \textbf{Data-free sparsity.} Can we design one-shot and data-free methods that rival the accuracy of data-dependent methods?
	\item \textbf{Fairness and bias.} How do we design more robust sparse models and sparsification approaches? How do we prevent adversarial attacks on sparsified models?
\end{enumerate}

We do not explicitly list brain-related research challenges because our work focuses primarily on the engineering aspects of sparsity for which biological analogies are certainly a major inspiration but act mainly as a means to an end.

\section{Conclusions and Outlook}\label{sec:conclusions}

We show that sparsity can already lead to a theoretical 10--100x improvement in efficiency.
Furthermore, larger networks appear to provide more opportunity for pruning~\cite{2020-sanh, 2019-gale}  so the compression trend is likely to continue as architectures get larger. Specifically, training extremely large models with sparse methods will provide many opportunities. 
Our detailed analysis of data science and engineering aspects enables a targeted hardware-software co-design for next-generation deep learning architectures that exploit the potentially huge speedups.

We also expect that there remains potential in the data science aspects of sparsity, especially in the areas of very high sparsity (>99\%) as well as sparse training of large models in very high-dimensional spaces. Both could lead to significant breakthroughs in future deep learning systems.

\subsection*{Acknowledgments}

We thank Doug Burger, Steve Scott, Marco Heddes, and the respective teams at Microsoft for inspiring discussions on the topic. We thank Angelika Steger for uplifting debates about the connections to biological brains and Sidak Pal Singh for his support regarding experimental results. 

\bibliographystyle{ACM-Reference-Format}
\bibliography{sparsity} 

\end{document}